%% file: emnlp_2026.tex
\documentclass[11pt]{article}
\pdfoutput=1

\usepackage[final]{acl}

\usepackage{pgf-pie}
\usepackage{array}
\usepackage{tabularx}
\usepackage{xcolor}
\newcommand{\up}{\textcolor{green!60!black}{$\uparrow$}}

\newcommand{\scorelabel}[1]{\textbf{#1}}

\usepackage{times}
\usepackage{latexsym}
\usepackage[T1]{fontenc}
\usepackage[utf8]{inputenc}
\usepackage{microtype}
\usepackage{hyperref}
\usepackage{url}
\usepackage{booktabs}
\usepackage{multirow}
\usepackage{graphicx}
\usepackage[table]{xcolor}
\usepackage{subcaption}
\usepackage{amsmath,amsfonts,bm}
\usepackage[most]{tcolorbox}
\usepackage{enumitem}
\usepackage{wrapfig}
\usepackage{placeins}
\usepackage{fancyvrb}
\usepackage{fvextra}  %
\usepackage{tikz}
\usetikzlibrary{positioning, arrows.meta, shapes.geometric, fit, backgrounds, calc}

\usepackage{booktabs}
\usepackage{xcolor}
\usepackage{colortbl}
\usepackage{array}
\usepackage{tabularx}
\usepackage[T1]{fontenc}
\usepackage[compact]{titlesec}
\definecolor{headerbg}{HTML}{EEF2F7}
\definecolor{rowalt}{HTML}{F7F9FC}
\definecolor{rulegray}{HTML}{B0B7C3}

\newcommand{\annot}[1]{\textbf{\textsf{#1}}}

\input{math_commands}

\hypersetup{
  colorlinks=true,
  linkcolor=darkblue,
  citecolor=darkblue,
  urlcolor=darkblue,
}
\definecolor{darkblue}{rgb}{0, 0, 0.5}

\definecolor{lightred}{HTML}{e99090}
\definecolor{posgreen}{HTML}{2E8B57}
\definecolor{negred}{HTML}{C0392B}
\definecolor{dimblue}{HTML}{2980B9}

\newcommand{\storylm}{\textsc{POLARIS}}
\newcommand{\storylmbig}{\textsc{POLARIS}-9B}
\newcommand{\storylmhri}{\storylmbig{}}

\def\tabref#1{Table~\ref{#1}}
\def\figref#1{Figure~\ref{#1}}
\def\secref#1{Section~\ref{#1}}

\title{\raisebox{-0.2\height}
{\includegraphics[height=1.25em]{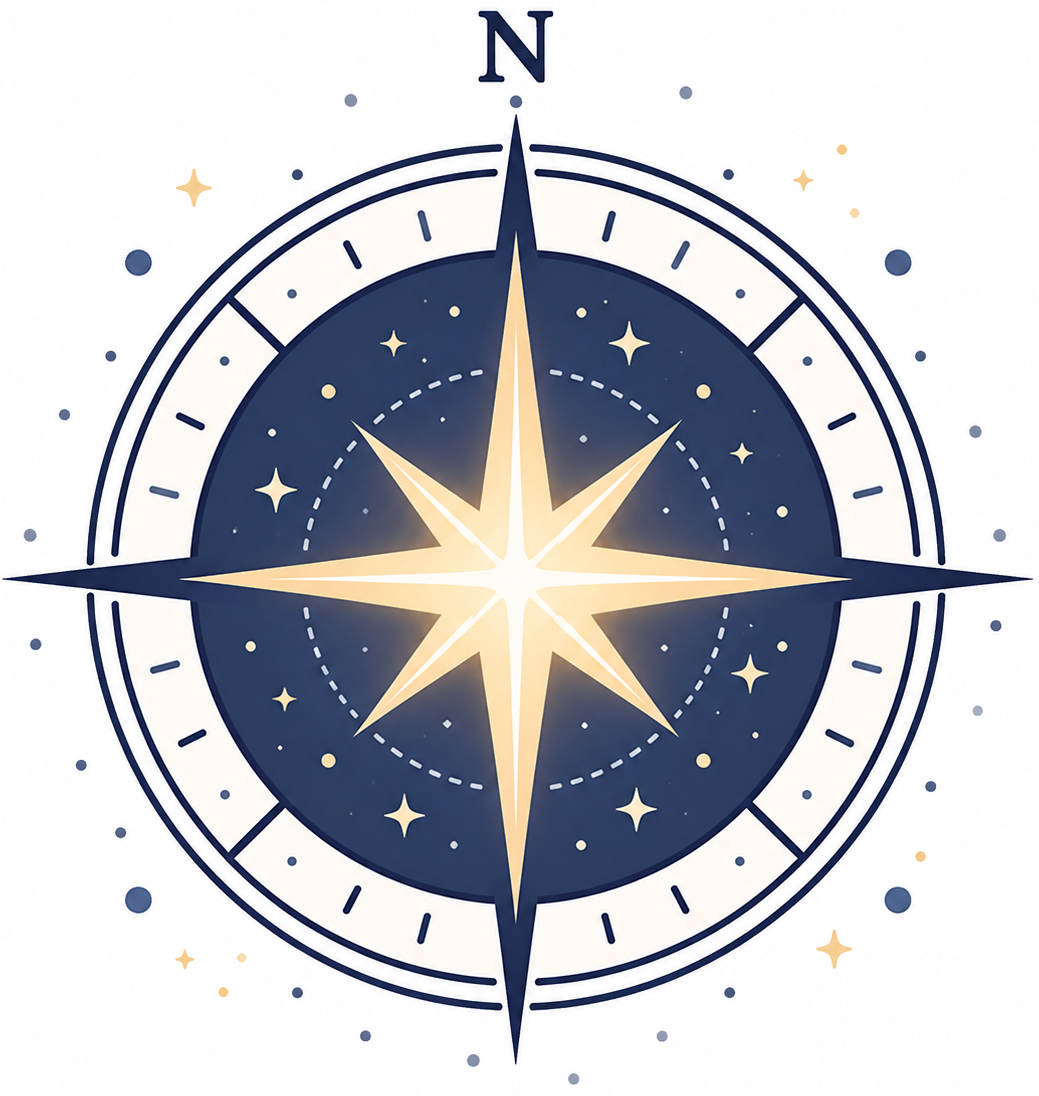}} \storylm: Guiding Small Models to Write Long Stories}

\newcommand{\umdlogo}{\includegraphics[height=1.25em]{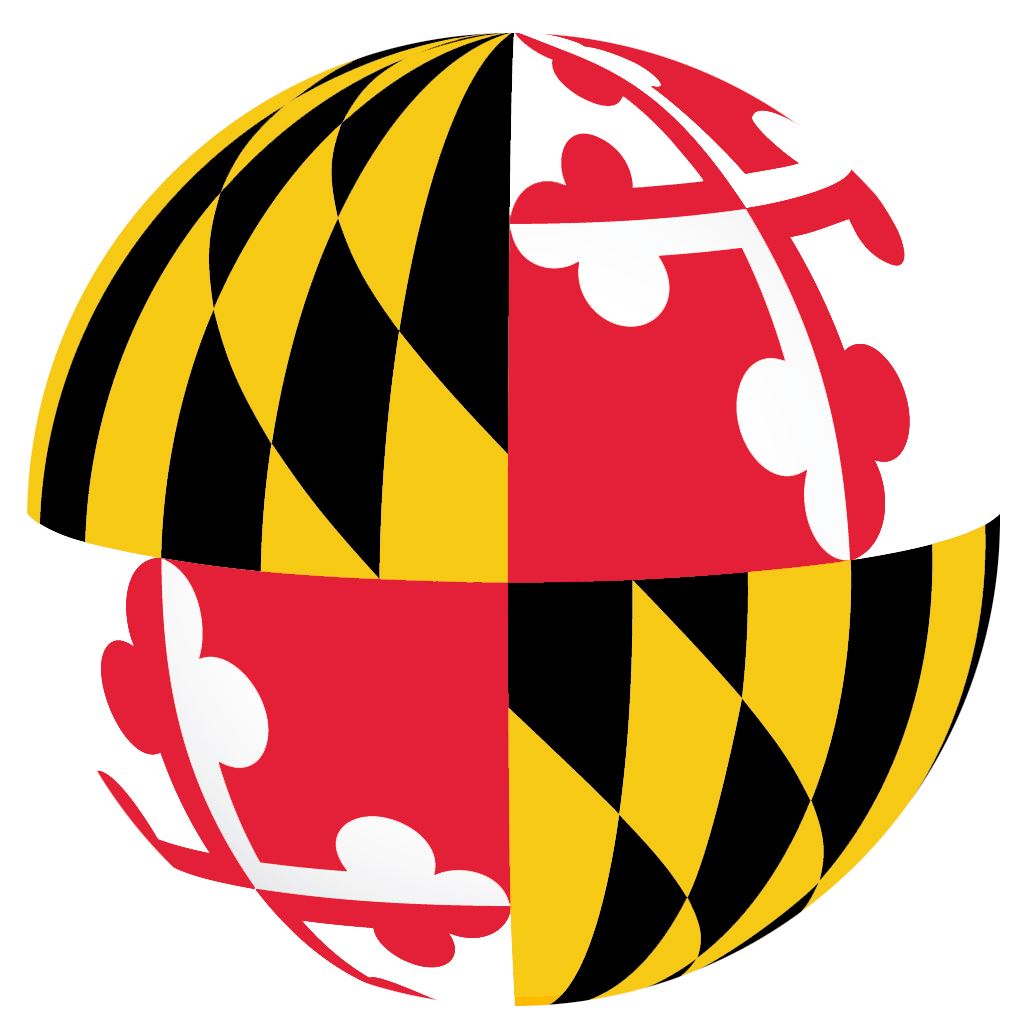}}
\newcommand{\dmlogo}{\includegraphics[height=1.25em]{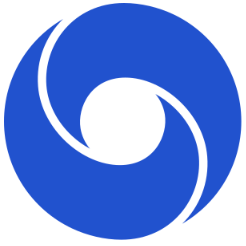}}
\author{
  Rishanth Rajendhran\textsuperscript{\umdlogo} \quad
  Jenna Russell\textsuperscript{\umdlogo} \quad
Mohit Iyyer\textsuperscript{\umdlogo} \quad
  John Wieting\textsuperscript{\dmlogo} \\
  \textsuperscript{\umdlogo}\,University of Maryland\quad
  \textsuperscript{\dmlogo}\,Google DeepMind \\
  \texttt{\{rishanth, jennarus, miyyer\}@umd.edu \quad jwieting@google.com} \\[0.9em]
  \textbf{Accepted to EMNLP 2026 (Main Conference)}
}

\begin{document}

\maketitle

\input{sections/0_abstract}

\input{sections/1_introduction}
\input{sections/2_related_work}
\input{sections/3_method}
\input{sections/4_evaluation}
\input{sections/5_experiments}

\input{sections/7_discussion}

\input{sections/limitations}

\bibliography{references}

\appendix
\input{sections/appendix}

\end{document}

%% file: math_commands.tex
\usepackage{amsmath,amsfonts,bm}

\def\figref#1{figure~\ref{#1}}

\def\secref#1{section~\ref{#1}}

\def\eqref#1{equation~\ref{#1}}

\def\1{\bm{1}}

\DeclareMathAlphabet{\mathsfit}{\encodingdefault}{\sfdefault}{m}{sl}
\SetMathAlphabet{\mathsfit}{bold}{\encodingdefault}{\sfdefault}{bx}{n}

%% file: sections/0_abstract.tex
\begin{abstract}
Small open-weight models struggle at long-form creative writing: 
their generated stories either fall far short of the requested length, or their 
quality significantly degrades as length increases, especially when compared to frontier models. We present 
\textsc{Polaris}\footnote{\textbf{P}olicy \textbf{O}ptimization with 
\textbf{L}LM-as-a-judge rewards and \textbf{A}nchored-\textbf{R}eference 
\textbf{I}njection for \textbf{S}torywriting.}, \textbf{a lower-compute GRPO 
recipe} with two key ingredients: a frontier 
LLM judge with a \textbf{structured \emph{Story Quality} rubric} as the 
online reward, and \textbf{human-reference injection (HRI)}, where 
a teacher-forced human-written story serves as a high-reward anchor 
within each GRPO group. By applying our training recipe to Qwen3.5-9B, using a dataset of
${\sim}$1.4K prompt--story pairs derived from 100 short-story 
anthologies and 4~A100 GPUs, we obtain
\storylmhri{}. Across five benchmarks spanning 
in-distribution and out-of-distribution prompts and rubrics, 
\textbf{\textsc{Polaris-9B}} 
is
\textbf{competitive with much larger open-weight 
models} while following length instructions more closely.
A blinded human evaluation confirms that \storylmhri{} is preferred to the base Qwen3.5-9B and on par with Qwen3.5-27B. Despite 
training only on stories up to 4k words, \storylmhri{} 
preserves quality on prompts requesting stories up to $3\times$ 
the training length, a regime where most 
open-weight models degrade substantially in quality, length 
adherence, or both. More broadly, 
our results suggest that \textbf{length generalization is a meaningful 
stress test for creative-writing models} and a useful lens for distinguishing 
otherwise close models.\footnote{Writing prompts, code, trained models can be found at \href{https://github.com/RishanthRajendhran/POLARIS}{rishanthrajendhran/POLARIS}.}\footnote{Browse generated stories at \href{https://storyeval.com/}{www.storyeval.com}.}\footnote{See \S\ref{app:memorization_audit} for memorization audit on training data.}
\end{abstract}

%% file: sections/1_introduction.tex
\section{Introduction}
\label{sec:intro}

\begin{figure*}[!t]
    \centering
    \includegraphics[width=\textwidth]{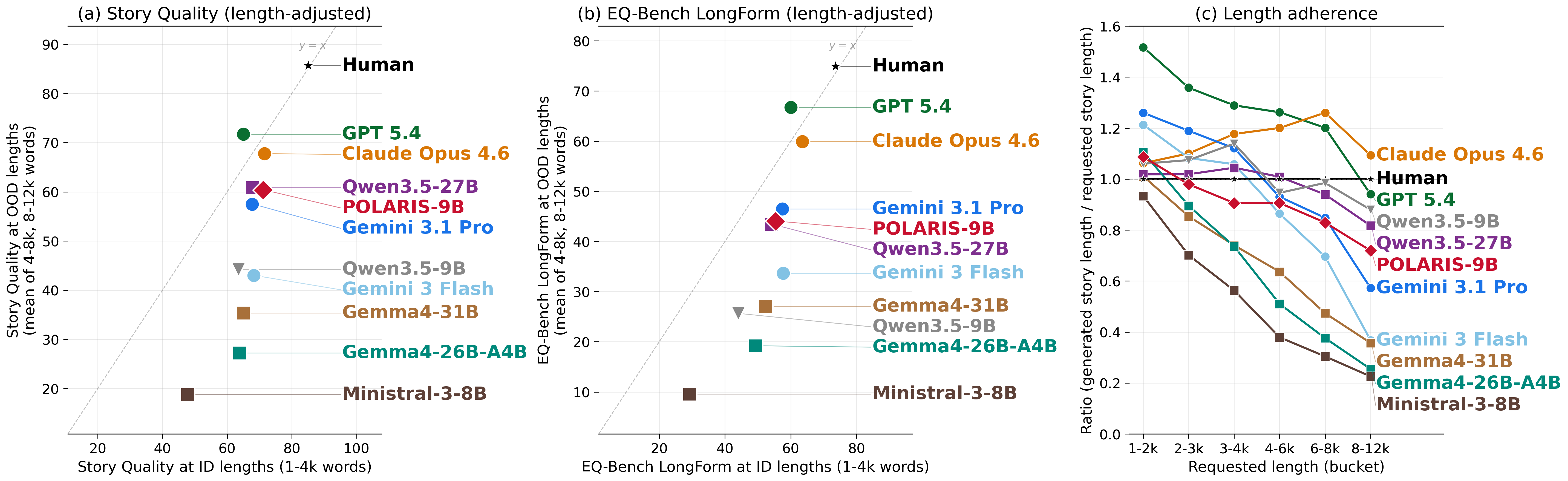}
    \caption{Length generalization (preview of \secref{sec:length_gen}). (a) Story Quality length-adjusted (for signed scores, we first normalize from the rubric floor and ceiling to $[0,100]$, then multiply by $\min(\mathrm{LR}, 1/\mathrm{LR})$) where LR is the length adherence ratio, calculated as the ratio between length of generated story and requested story length: scatter of mean ID-length score (1--4k) vs.\ mean OOD-length score (mean of near and far buckets); the $y\!=\!x$ diagonal indicates no length-related degradation. (b) Same scatter for EQ-Bench Longform. (c) Length adherence (generated/requested word count) across requested-length buckets. \storylmbig{} sits in the top open-weight cluster on Story Quality and alongside Qwen3.5-27B in the top open-weight cluster on EQ-Bench Longform, and fares the best among open-weight models when considering both quality and length adherence together at far transfer. Length-adjusted scoring penalizes models such as Gemma 4 31B that maintain quality only by writing progressively shorter stories. 
    }
    \label{fig:length_degradation}
\end{figure*}

Reinforcement learning (RL) with verifiable rewards has driven 
strong gains in LLM reasoning and code~\citep{deepseekr1,dapo}.
Recent work shows RL can improve 
long-form generation~\citep{longwriterzero,rlmr,rlcs,writingrl,
writingzero,dpwriter}, but the strongest recipes rely on some combination of large or specialized 
base models, continual pretraining, custom-trained reward models, or large amounts of training data.

This raises a more practical question: can a small open model become competitive at long-form creative writing \textit{without} that scale of compute and infrastructure? Starting from Qwen3.5-9B \cite{qwen}, we apply \storylm{}, our 
lower-compute creative-writing RL recipe, to 1,388 
prompt--story pairs from a corpus derived 
from 100 commercially purchased short-story anthologies, 
on 4 A100 GPUs. The recipe addresses two 
obstacles specific to long-form creative RL:

\textbf{\includegraphics[height=0.8em]{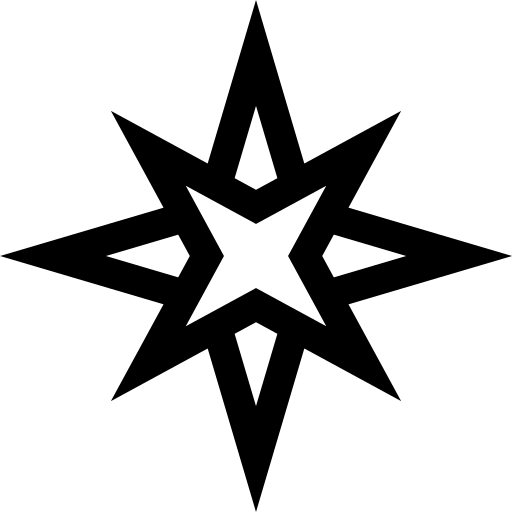}} \textbf{Reward signal design.} Prior work uses trained reward models that produce scalar or pairwise preferences~\citep{longwriterzero,rlmr,rlcs} making it hard to tell which quality dimensions improved, and require a separate reward-model training pipeline. We replace this with 
    a frontier LLM judge queried 
    during each GRPO step using a structured \emph{Story Quality} rubric that returns per-dimension scores with textual evidence across 16 core plus 2 catch-all dimensions 
    combining narrative-theoretic concerns \citep{boyd2020narrativearc,waters2018narrative,cremin2021narrativeassessment} with practical fiction-writing norms used in workshop and editorial feedback and empirical observations made during the course of this work from reading human-written and AI-generated stories.

\textbf{\includegraphics[height=0.8em]{figures/north_star.png}} \textbf{Stagnation in open-ended long-form RL.} Due to the open-ended nature of creative writing, policy rollouts could receive increasingly similar reward scores as GRPO training progresses. This can cause learning to slow or stall before the model reaches stronger long-form writing behavior. We propose \emph{human-reference injection} (HRI) during GRPO training, in which a teacher-forced human-written story is inserted into each group as a high-quality anchor. This helps by maintaining gradient pressure toward stronger writing rather than letting training be dominated entirely by the model's own rollout distribution. The human reference is excluded from group statistics but included in advantage computation with a warmup schedule.

Our recipe yields \storylmhri{}, a 9B model that is competitive 
with models $3\times$ its size and remains robust on prompts 
requesting stories up to 12k words, despite seeing no story longer than 4k words during training. Figure~\ref{fig:length_degradation} 
shows this key result: \storylmhri{} largely maintains both rubric quality and length adherence across the full 1k--12k requested length range.
A blinded human evaluation points in the same direction: annotators clearly prefer \storylmhri{} to the base model, and their comments most often highlight stronger atmosphere, voice, and scene realization, while the comparison to Qwen3.5-27B remains close.

Our contributions are:

\textbf{\includegraphics[height=0.8em]{figures/north_star.png}} \textbf{A lower-compute GRPO recipe for long-form creative 
    writing} using an LLM-as-judge reward with a structured rubric (\emph{Story Quality})
 and human-reference injection (HRI), 
    yielding a 9B model that ranks in the top open-weight 
    cluster on EQ-Bench Longform and above every open-weight 
    model and several frontier models on EQ-Bench Creative Elo.

\textbf{\includegraphics[height=0.8em]{figures/north_star.png}} \textbf{Length generalization} is a meaningful 
    stress test for creative writing models: a model that 
    merely learns surface-level writing patterns will degrade 
    as stories grow longer and require sustained narrative 
    coherence, arc completion, and stylistic consistency across 
    thousands of tokens. Despite seeing no story longer than 4k 
    words, \storylmhri{} retains strong rubric scores at 8--12k, 
    a regime where most open-weight models collapse in quality, 
    undershoot requested length, or both.

%% file: sections/2_related_work.tex
\section{Related Work}
\label{sec:related}

\paragraph{Long-form writing with SFT and RL.}
LongWriter~\citep{longwriter} shows that SFT on synthetic long-output data can make models produce 10k+ word outputs.
LongWriter-Zero~\citep{longwriterzero} employs RL training, but uses a 32B base model and trained rewards for length, quality, and formatting.
RLMR~\citep{rlmr} also uses a trained writing reward model, along with checks for hard constraints.
Writing-RL~\citep{writingrl} uses a specialized base model and adaptive references for long-form writing, while Writing-Zero~\citep{writingzero} uses generative reward models for open-ended writing.
RLCS~\citep{rlcs}, Writer-R1~\citep{writerr1}, and R2-Write~\citep{r2write} add further steps such as human-aligned rewards, replay, reasoning, reflection, or revision.
DPWriter~\citep{dpwriter} focuses on diverse planning during creative-writing RL.

\paragraph{Preference optimization and off-policy guidance.}
DPO~\citep{dpo} gives the standard preference-optimization objective, while SWAG~\citep{swag} adapts preference learning to storytelling.
CrPO~\citep{crpo} studies preference optimization for creative writing and DivPO~\citep{divpo} adds an explicit diversity objective. LUFFY~\citep{luffy} uses off-policy guidance for reasoning RL, while RePO~\citep{repo} reuses earlier policy outputs during optimization.
G2RPO-A~\citep{g2rpoa} adds adaptive guidance inside group relative policy optimization, and BREAD~\citep{bread} branches from expert anchors to connect SFT and RL.

\paragraph{Creative-writing evaluation.}
WritingBench~\citep{writingbench} covers broad generative writing, EQ-Bench~\citep{eqbench} includes creative and emotional writing judgments, and LongBench-Write~\citep{longwriter} focuses on long-form generation.
HelloBench~\citep{hellobench} and LitBench~\citep{litbench} test for long text generation and creative-writing quality. \citet{llm_judge_survey} and \citet{llm_judge_survey2} analyze the values and biases of LLM-as-a-judge. 
Prometheus~\citep{prometheus, prometheus2} shows that judge models can give fine-grained rubric-based feedback.
Igniting Creative Writing~\citep{igniting_creative} uses LLM judges to improve short creative text.

%% file: sections/3_method.tex
\section{Method}
\label{sec:method}

\begin{figure}[!tb]
    \centering
    \resizebox{\columnwidth}{!}{%
    \begin{tikzpicture}[
        font=\small,
        box/.style={draw, rounded corners=2pt, align=center,
                    inner sep=4pt, minimum height=7mm, fill=white},
        policy/.style={box, fill=blue!6, draw=blue!55},
        refnode/.style={box, fill=orange!12, draw=orange!75, thick},
        process/.style={box, fill=gray!8, draw=gray!55},
        scorepolicy/.style={box, fill=blue!4, draw=blue!45},
        scoreref/.style={box, fill=orange!8, draw=orange!70},
        backfit/.style={draw, rounded corners=2pt, inner sep=0pt, minimum height=0pt},
        flow/.style={-Latex, semithick},
        refflow/.style={-Latex, semithick, orange!80!black, dashed},
    ]
      \def\yRef{8.4}
      \def\yPrompt{6.2}
      \def\yPolicy{4.65}
      \def\yRolls{3.2}
      \def\yReward{-0.5}
      \def\yScores{-4.25}
      \def\yGrpo{-6.15}
    
      \node[refnode, text width=50mm, align=left] (ref)
            at (0, \yRef)
            {\textbf{Human story $y^\star$}\\[-1pt]
             \scriptsize \emph{Pink Lady Friends}\\
             {\scriptsize\color{gray!70}\textit{by Allison Wonderland}}\\
             \scriptsize ``Duck\ldots duck\ldots goose! I make good on my word,
             directing my pointer finger toward the unsuspecting posterior\ldots''};
    
      \node[box, text width=52mm, align=left] (prompt)
            at (0, \yPrompt)
            {\textbf{Prompt $p$}\\[-1pt]
             \scriptsize Write a story about a student who, after a triumphant
             school-musical performance, struggles to carry that same boldness
             into real life.};
    
      \draw[refflow] (ref.south) --
            node[right, font=\scriptsize] {by LLM}
            node[left, font=\scriptsize] {reverse-engineered}
            (prompt.north);

      \node[policy, minimum width=26mm] (policybox)
            at (0, \yPolicy)
            {Policy\\ $\pi_\theta$};
    
      \node[policy, text width=52mm, align=left] (rolls)
            at (0, \yRolls)
            {\textbf{Policy rollouts $o_1,\ldots,o_k$}\\[-1pt]
             \scriptsize
             $o_1$: ``After the curtain call, Leslie still felt Frenchy
             in her bones\ldots''};
    
      \node[draw=gray!60, dashed, rounded corners=2pt, fill=white,
            text width=58mm, align=left, inner sep=4pt] (sq)
            at (0, \yReward+0.55)
            {\scriptsize
             \textbf{Positive:}\\
             \textbf{P1} Prompt realization,\ \textbf{P2} Narrative arc,\ \textbf{P3} Character depth, \textbf{P4} Voice,\\ 
             \textbf{P5} Scene vividness,\ \textbf{P6} Thematic coherence, \textbf{B} Bonus\\[2pt]
             \textbf{Negative:}\\
             \textbf{N1} Prompt violation,\ \textbf{N2} Coherence breaks,\ \textbf{N3} Generic language,\\
             \textbf{N4} Over-summary,\ \textbf{N5} Over-explanation,\ \textbf{N6} Drift / bloat,\\
             \textbf{N7} Dialogue problems,\ \textbf{N8} Mechanical errors,\ \textbf{N9} Predictability / cliché,\\
             \textbf{N10} Overwrought prose, \textbf{D} Catch-all penalty};

      \node[font=\scriptsize, fill=white, inner sep=1.5pt]
            at (0, \yReward+2.18) {\textbf{Story Quality}};

      \node[process, text width=58mm, align=left] (aux)
            at (0, \yReward-1.55)
            {\scriptsize
             \textbf{Other:} $r_{\mathrm{rep}}$,\ $r_{\mathrm{len}}$,\ $r_{\mathrm{blank}}$};

      \begin{scope}[on background layer]
        \node[process, backfit, fit=(sq)(aux), inner sep=5pt] (reward) {};
      \end{scope}

      \node[font=\small, fill=gray!8, inner sep=1.5pt]
            at (0, \yReward+2.62) {\textbf{Reward}};
    
      \node[scorepolicy, text width=25mm, align=left] (scoreo)
            at (-2.15, \yScores)
            {\scriptsize \textbf{$o_1$ score}\\
             P2 arc $=4$, P4 voice $=3$\\
             N6 drift $=-2$, N10 overwr. $=-1$\\
             bonus $=0$, penalty $=-1$};
    
      \node[scoreref, text width=25mm, align=left] (scorey)
            at (2.15, \yScores)
            {\scriptsize \textbf{$y^\star$ score}\\
             P2 arc $=5$, P4 voice $=4$\\
             N6 drift $=0$, N10 overwr. $=0$\\
             bonus $=+1$, penalty $=0$};
    
      \node[policy, minimum width=30mm] (grpo)
            at (0, \yGrpo)
            {GRPO\\ update};
    
      \begin{scope}[on background layer]
        \node[policy, backfit, fit=(rolls), xshift=0.12cm, yshift=0.12cm] (rollsback) {};
        \node[policy, backfit, fit=(rolls), xshift=0.06cm, yshift=0.06cm] (rollsmid) {};
    
        \node[scorepolicy, backfit, fit=(scoreo), xshift=0.09cm, yshift=0.09cm] (scoreoback) {};
        \node[scorepolicy, backfit, fit=(scoreo), xshift=0.045cm, yshift=0.045cm] (scoreomid) {};
      \end{scope}
    
      \draw[flow]    (prompt.south) -- (policybox.north);
      \draw[flow]    (policybox.south) -- (rolls.north);
      \draw[flow]    (rolls.south) -- (reward.north);
      \draw[refflow] (ref.south east) .. controls +(1.1,-2.2) and +(1.15,1.1) .. (reward.east);

     \draw[flow]    (reward.south west) .. controls +(-0.7,-1.45) and +(0,1.0) .. (scoreo.north);
     \draw[refflow] (reward.south east) .. controls +(0.7,-1.45) and +(0,1.0) .. (scorey.north);
    
      \draw[flow]    (scoreo.south) -- ++(0,-0.22) |- (grpo.west);
      \draw[refflow] (scorey.south) -- ++(0,-0.22) |- (grpo.east);
    
      \draw[flow] (grpo.south) -- ++(-0.35,-0.35) -| ++(-3.55,7.0) |- (policybox.west);
    
    \end{tikzpicture}%
    }
    \caption{\textsc{Polaris} training pipeline. For each 
    prompt, the policy generates $k{=}5$ rollouts (blue) 
    while a human reference $y^\star$ is teacher-forced as 
    the $(k{+}1)$-th group member (orange). All outputs are 
    scored by the reward; group statistics are 
    computed from policy rollouts only and the reference 
    enters the update through a warmup-scaled advantage 
    $\alpha_t$.}
    \label{fig:pipeline}
\end{figure}
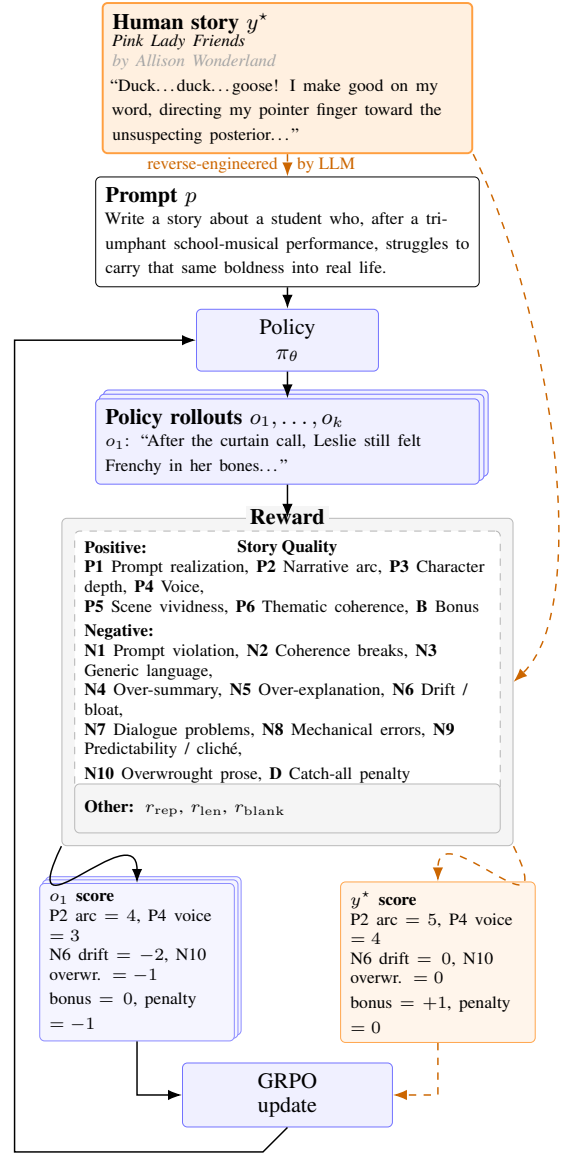

We apply GRPO~\citep{deepseekmath, deepseekr1} to long-form creative 
writing with a structured LLM-as-judge online reward 
and human-reference injection (HRI). 
Figure~\ref{fig:pipeline} summarizes the full pipeline.

\subsection{Structured LLM-as-Judge Reward}
\label{sec:rubric_reward}

\paragraph{Motivation.}
Prior RL approaches to creative writing rely on a reward 
model trained on pairwise human preference 
data~\citep{longwriterzero,rlmr,rlcs}. Trained reward 
models collapse writing quality into a single scalar, 
require a separate training pipeline, and can go stale 
as training progresses and policy rollouts diverge from 
the reward model's training distribution. More recent 
work applies LLM-as-a-judge as an online reward for 
shorter-form creative writing~\citep{igniting_creative,
writingrl}, but with rubrics that are limited in scope. 
Existing benchmarks~\citep{eqbench,writingbench,longwriter} were primarily designed for ranking AI-generated writing---with limited descriptions, score anchors---which makes them unsuitable for use as a reward during RL to improve creative writing. We instead query a frontier LLM judge at each 
GRPO step using a comprehensive structured rubric that 
returns per-dimension numeric scores and textual evidence, 
with no reward model training required.\footnote{Full rubric with per-dimension anchors, non-overlap rules, 
and scoring anchors is in \S\ref{app:storyquality_prompt}.}

\paragraph{\emph{Story Quality} rubric.}
The \emph{Story Quality} rubric evaluates stories on six 
positive dimensions that capture sustained global qualities 
(prompt fulfillment, narrative arc and pacing, character 
depth and agency, voice and stylistic distinctiveness, 
scene realization, thematic and emotional richness; 
$(P_{\text{total}}) \in [0, 100]$) and ten negative dimensions that 
penalize localizable failures (prompt violation, 
coherence and POV consistency, generic language, 
over-summary, over-explanation, structural drift, dialogue 
problems, mechanical errors, predictability, overwrought 
language; $N_{\text{total}} \in [0, 151]$), plus open-ended bonus (B) and 
penalty (D) catch-all terms ($\in [0, 8]$ each). \\ 
The raw \emph{Story Quality} score is:
\begin{equation}
    s_q^{\mathrm{raw}} = P_{\text{total}} - N_{\text{total}} 
    + B - D.
    \label{eq:quality_score}
\end{equation}

\paragraph{Dimensions grounded in narrative theory and narrative-quality assessment.}
The rubric dimensions are motivated by both narrative theory and empirical work on narrative-quality assessment. Narrative arc, pacing, scene realization, over-summary, and drift reflect concerns about large-scale narrative structure and progression~\citep{boyd2020narrativearc,waters2018narrative}. Voice, characterization, atmosphere, and coherence are also widely emphasized in studies of how narrative writing is assessed in practice~\citep{cremin2021narrativeassessment}. Prompt fulfillment and mechanical correctness are additionally aligned with dimensions used in recent creative-writing benchmarks~\citep{eqbench,writingbench,longwriter}.

\paragraph{Dimensions identified from AI-writing.}
Three negative dimensions came from comparing LLM-generated 
stories against human-written references.
Generic and templated language captures the stock 
phrases and boilerplate language characteristic of 
LLM-generated text. Predictability and cliché captures formulaic plot structures and recycled plot narratives; human-written stories are generally more creative, differentiating them from AI writing~\citep{russell2026storyscope}. Overwrought language was added after observing that models optimized 
toward the positive dimensions produced systematically 
purple prose.

\paragraph{Composite reward.}
For training, the judge score is normalized by dividing by a score divisor $Z$,
\[
\tilde{s}_q = \frac{s_q^{\mathrm{raw}}}{Z},
\]
where $Z$ is the rubric's maximum overall score (108).
It is then combined with three auxiliary 
penalty components, with each component weighted by $w_q$, $w_{rep}$, $w_{len}$, and $w_{blank}$ respectively,  to arrive at the final composite score:
\begin{align}
    r_{\text{qual}} &= w_q \cdot \tilde{s}_q \cdot 
    g(r_{\text{rep}}, r_{\text{len}}, r_{\text{blank}}) 
    \label{eq:pos_reward} \\
    r_{\text{pen}} &= w_{\text{rep}} \cdot r_{\text{rep}} + 
    w_{\text{len}} \cdot r_{\text{len}} + 
    w_{\text{blank}} \cdot r_{\text{blank}} 
    \label{eq:neg_reward} \\
    r &= \mathrm{clip}\!\left((r_{\text{qual}} - r_{\text{pen}}) 
    \cdot \ell(n),\; {-}c,\; {+}c\right),
    \label{eq:composite}
\end{align}
where $r_{\text{rep}}$ penalizes repetition, $r_{\text{len}}$ 
penalizes target-length mismatch, $r_{\text{blank}}$ 
penalizes empty outputs, and $g$ gates the positive reward 
to zero for severely malformed outputs. The length-scaling 
factor\footnote{$\ell(n) = \max(\ell_{\min}, (n/n_{\text{target}})
^{\beta})$ with $\beta{=}0.5$ and $\ell_{\min}{=}0.15$} 
counters an observed reward inflation failure mode: short outputs 
expose fewer sentences to negative dimensions and can 
receive inflated normalized scores. In our experiments, a dominant length 
penalty often destabilized training, so we use a sublinear scale 
with a floor instead. Clipping the composite reward score at $c{=}2.0$ bounds the 
overall reward.\footnote{Full reward weights and hyperparameters are 
in \S\ref{app:hyperparams}; exact penalty and gate formulas are in \S\ref{app:reward_details}.}

\subsection{Human-Reference Injection}
\label{sec:hri}

\paragraph{Reference as group anchor.}
We inject a human-written story as the $(k{+}1)$-th member 
of each GRPO group. The injected reference is a complete 
assistant turn in the same format the policy is trained to 
produce: the synthetic thinking trace paired with the story 
(\S\ref{sec:eval_framework}) inside the model's think tags, 
followed by the human story itself. The reference is 
teacher-forced through the current policy and scored by the 
same composite reward as policy rollouts. Because references 
typically score higher than policy outputs, especially early 
in training, they provide a stable high-reward anchor that 
maintains within-group reward variance. Three constraints 
govern the design: (i) the group mean and standard deviation 
are computed from the $k$ policy outputs only, excluding the 
reference; (ii) the reference advantage is scaled by a warmup 
schedule; and (iii) the reference is scored by the same 
reward function as policy outputs.

Formally, let $r_1, \dots, r_k$ be the composite rewards 
(Eq.~\ref{eq:composite}) of the $k$ policy rollouts in a 
group, $\mu_\pi$ and $\sigma_\pi$ their mean and standard 
deviation, and $r^\star$ the reward of the reference 
$y^\star$. Policy rollouts receive the standard GRPO 
advantage $A_i = (r_i - \mu_\pi)/(\sigma_\pi + \epsilon)$, 
and the reference receives
\begin{equation}
    A^\star_t = \alpha_t \, \frac{r^\star - \mu_\pi}{\sigma_\pi + \epsilon},
    \qquad
    \alpha_t = \alpha_{\max} \min\!\left(1, \frac{t}{T_w}\right),
    \label{eq:hri_advantage}
\end{equation}
where $t$ is the (zero-indexed) training step, 
$\alpha_{\max}{=}0.4$, $T_w{=}15$ steps, and 
$\epsilon{=}10^{-6}$; $\alpha_t$ ramps linearly from 0 to 
$\alpha_{\max}$ over the first $T_w$ steps and stays at 
$\alpha_{\max}$ thereafter (the ``HRI warmup steps'' and 
``HRI peak weight'' in \S\ref{app:hyperparams}). Because 
the reference is teacher-forced, its log-probabilities under 
the current policy are available, and it is optimized through 
the same clipped surrogate as the policy rollouts with 
$A^\star_t$ in place of $A_i$; $\alpha_t$ is the only 
reference-specific quantity in the update.

\paragraph{Relation to prior off-policy methods.}
Because the reference is not sampled from the policy, HRI 
is a biased demonstration-augmented GRPO update rather than 
an unbiased on-policy estimator. Prior 
off-policy injection methods~\citep{luffy,g2rpoa,bread} 
target verifiable-reward tasks and inject partial expert 
prefixes rather than complete reference outputs.\footnote{Refer to \S\ref{app:hri_design} for a detailed comparison.} Our setting is also complementary to \citet{storygen_reason}, which likewise uses human-written books and trains models to reason about long-form narrative continuation. In contrast, we focus on full-story generation rather than next-chapter prediction, and treat thinking as a means for better stories rather than as the primary object of optimization.

%% file: sections/4_evaluation.tex
\section{Experimental Setup}
\label{sec:eval_framework}

Starting from Qwen3.5-9B, \storylmhri{} was trained using ${\sim}$1.4K human-written stories on 4 A100 GPUs for about 48 hours. Our automatic evaluation spans 
17 models, 5 benchmarks covering in-distribution and out-of-distribution prompts 
and rubrics, and over 160K per-dimension 
pairwise judgments.

\paragraph{Training data.}
We train on 1,388 (prompt, story) pairs from 
a corpus of 100 commercially purchased short-story 
anthologies (431 authors). No 
reference exceeds 4,000 words. Writing prompts are 
reverse-engineered from each story and references are paired with synthetic thinking traces using Gemini~3 Flash \cite{geminiteam2023gemini,gemini3flash}. We will not release the human written stories from this corpus due to copyright concerns.\footnote{Data construction details are in \S\ref{app:data_summary}; release scope is in \S\ref{app:data_statement}.}

\paragraph{Model configurations.}
Using Qwen3.5-9B~\citep{qwen} as the base model, we train 
two variants, both with group size 6: \storylmhri{} uses 5 
policy rollouts and 1 injected human reference per group, 
while the matched plain GRPO run (without HRI) uses 6 policy rollouts with no 
reference.\footnote{This comparison is conservative toward HRI: the HRI run sees one fewer policy rollout per group.} We also report an SFT baseline trained 
on the same data. Due to cost and compute considerations, we leave ablating the base model to future work.

\paragraph{Training details.}
Both models are trained on 4$\times$A100 80GB GPUs with 
FSDP~\citep{fsdp} and batch size 8 for 160 steps, taking 
approximately 48 hours per run. At \$2/hr per GPU this 
comes to ${\sim}\$400$ in compute per checkpoint.
Judge 
costs add approximately \$60 per run using Gemini~3 Flash \cite{gemini3flash}
at flex tier, bringing the total cost per model to just 
under \$500. Full hyperparameters are in 
\S\ref{app:hyperparams}.

\begin{table*}[!t]
\centering
\scriptsize
\begin{minipage}[t]{0.42\textwidth}
\centering
\setlength{\tabcolsep}{1.5pt}
\begin{tabular}{@{}lcccccc@{}}
\toprule
& \shortstack{Writing-\\Bench (D4)} & \multicolumn{3}{c}{\shortstack{LongBench-Write}} & \shortstack{EQ-Cr} \\
\cmidrule(lr){3-5}
Model & & $\bar{S}$ & $S_Q$ & $S_L$ & \\
\midrule
\multicolumn{6}{l}{\cellcolor{white}\textsc{Frontier}} \\
GPT-5.4 & \underline{\textbf{9.5}} & 83.3 & \underline{\textbf{98.4}} & 84.7 & \underline{\textbf{84.6}} \\
Claude Opus 4.6 & 9.3 & \underline{\textbf{86.7}} & \underline{\textbf{98.4}} & \textbf{88.1} & 80.5 \\
Gemini 3.1 Pro & 9.2 & 79.8 & 94.2 & 84.8 & 73.8 \\
Gemini 3 Flash & 9.1 & 80.3 & 96.8 & 83.2 & 75.0 \\
\midrule
\multicolumn{6}{l}{\cellcolor{white}\textsc{Open-weight}} \\
LongWriter-Zero-32B & 4.6 & 58.3 & 76.8 & 73.0 & 41.5 \\
Gemma 4 31B & \textbf{8.7} & 77.6 & 95.8 & 81.1 & \textbf{70.9} \\
Qwen3.5-27B & 8.1 & 77.8 & 85.3 & \underline{\textbf{91.5}} & 68.9 \\
Gemma 4 26B-A4B$^*$ & 8.5 & 73.4 & \textbf{97.6} & 75.5 & 70.5 \\
\tiny{
DeepSeek-R1-Distill-Qwen-14B} & 3.9 & 41.5 & 71.9 & 53.6 & 42.8 \\
Gemma 4 E4B$^\dagger$ & 7.7 & 78.4 & 94.9 & 82.7 & 64.2 \\
Ministral-3-8B & 4.8 & 57.5 & 80.5 & 67.9 & 53.9 \\
LongWriter-Llama3.1-8B & 3.9 & 57.8 & 71.4 & 75.9 & 35.1 \\
Qwen3.5-9B & 6.8 & 67.1 & 75.1 & 90.3 & 59.2 \\
\rowcolor{headerbg} \; + \storylm{} & 7.9 & \textbf{81.2} & 90.2 & 88.5 & 70.3 \\
\bottomrule
\end{tabular}
\par\vspace{0.3em}\scriptsize (1a) OOD prompts \& rubrics
\end{minipage}%
\hfill%
\begin{minipage}[t]{0.27\textwidth}
\centering
\setlength{\tabcolsep}{2pt}
\begin{tabular}{@{}rlr@{}}
\toprule
\# & Model & Elo \\
\midrule
1 & GPT-5.4 & 1911 \\
2 & Claude Opus 4.6 & 1783 \\
\rowcolor{headerbg} 3 & \storylmbig{} & 1661 \\
4 & Gemini 3.1 Pro & 1627 \\
5 & Gemini 3 Flash & 1620 \\
6 & Gemma 4 31B & 1514 \\
7 & Qwen3.5-27B & 1503 \\
8 & Gemma 4 26B-A4B$^*$ & 1460 \\
9 & Qwen3.5-9B & 1352 \\
10 & Gemma 4 E4B$^\dagger$ & 1312 \\
11 & Ministral-3-8B & 1219 \\
12 & \tiny{
DeepSeek-R1-Distill-Qwen-14B} & 1030 \\
13 & LongWriter-Zero-32B & 1013 \\
14 & LongWriter-Llama3.1-8B & 800 \\
\bottomrule
\end{tabular}
\par\vspace{0.3em}\scriptsize (1b) EQ-Bench Creative Elo\\OOD: 32 prompts, 9 dims
\end{minipage}%
\hfill%
\begin{minipage}[t]{0.27\textwidth}
\centering
\setlength{\tabcolsep}{2pt}
\colorlet{idbg}{blue!5}
\begin{tabular}{@{}rlr@{}}
\toprule
\# & Model & Elo \\
\midrule
1 & GPT-5.4 & 1574 \\
2 & Claude Opus 4.6 & 1468 \\
3 & Human & 1464 \\
\rowcolor{headerbg} 4 & \storylmbig{} & 1419 \\
5 & Gemini 3.1 Pro & 1387 \\
6 & Gemini 3 Flash & 1335 \\
7 & Qwen3.5-27B & 1285 \\
8 & Gemma 4 31B & 1283 \\
9 & Gemma 4 26B-A4B$^*$ & 1235 \\
10 & Gemma 4 E4B$^\dagger$ & 1108 \\
11 & Qwen3.5-9B & 1086 \\
12 & Ministral-3-8B & 1055 \\
13 & \tiny{DeepSeek-R1-Distill-Qwen-14B} & 891 \\
14 & LongWriter-Llama3.1-8B & 868 \\
15 & LongWriter-Zero-32B & 800 \\
\bottomrule
\end{tabular}
\par\vspace{0.3em}\scriptsize (1c) ID prompts Elo\\OOD dims\\45 prompts, 11 judge dims + length\\113K dual-position comparisons
\end{minipage}
\caption{\textbf{(1a)} WritingBench (D4 English literature/arts: judged by Gemini 3.1 Pro), LongBench-Write (judged by Gemini 3.1 Pro), and EQ-Bench Creative (EQ-Cr, judged by GPT-5.4); all use out-of-distribution prompts and rubrics (w.r.t. our training). For LongBench-Write, $\bar{S}$ is the benchmark's overall score, $S_Q$ its quality score, and $S_L$ its length score. \textbf{(1b)} Pairwise Elo on EQ-Bench Creative. \textbf{(1c)} Pairwise Elo on the in-domain test subset. Elo is the Bradley--Terry MLE on the full pairwise contingency (Flash judge, dual-position), anchored so the lowest-rated model lands at 800. $^*$MoE; $^\dagger$PLE. 
}
\label{tab:ood_results}
\end{table*}

\paragraph{Test set.}
Our 180-prompt evaluation set is a disjoint held-out anthology pool, with 30 prompts in each of six 
target-length buckets: 1--2k, 2--3k, 3--4k, 4--6k, 6--8k, 
and 8--12k words. Each prompt is paired with a 
human-written reference story. Since \textsc{Polaris} 
trains on prompts targeting up to 4k words, we treat 
1--4k as \emph{in-distribution} (ID) lengths and 4--12k 
as \emph{out-of-distribution} (OOD) lengths, with OOD 
further split into near transfer (4--8k) and far transfer 
(8--12k).

\paragraph{Benchmarks.}
We evaluate along two axes: prompt distribution (ID 
vs.\ OOD) and evaluation rubric (ID vs.\ OOD rubric 
family). On ID prompts, we report \emph{Story Quality} (our 
training rubric) and EQ-Bench Longform~\citep{eqbench}, 
a related but distinct 12-dimension rubric.\footnote{For more details, refer to \S\ref{app:rubric_relation}.}
On OOD 
prompts and rubrics, we report EQ-Bench 
Creative~\citep{eqbench}, English D4 (Literature \& Arts) subset of WritingBench~\citep{writingbench}, and LongBench-Write~\citep{longwriter}.\footnote{Generation and decoding settings for all evaluated models are in \S\ref{app:decoding_settings}.}
We also compute pairwise Elo rankings~\citep{eqbench} on 
EQ-Bench Creative and on a 45-prompt ID subset using dual-position evaluation to mitigate positional bias in pairwise LLM judgments~\citep{wang2024fair_evaluators}.
For EQ-Bench Longform, we weight all dimensions uniformly rather than the 
canonical formula to avoid rewarding catastrophic failures.\footnote{Refer to \S\ref{app:pro_bias} for a more detailed discussion.}

\raggedbottom
\paragraph{Judges.}
GPT-5.4 \cite{gpt54} scores Story Quality, EQ-Bench Longform, and 
EQ-Bench Creative; Gemini~3.1 Pro \cite{gemini31pro} scores WritingBench 
and LongBench-Write; Gemini~3 Flash is used for pairwise 
Elo and as the online training reward; Judges do not see requested length; length adherence is computed analytically. Using a different 
judge for training and evaluation reduces the 
risk that the model overfits to the evaluation judge's 
preferences.\footnote{Detailed judge comparisons and bias analyses 
are in \S\ref{app:pro_bias}, which also replicates two of our evaluations 
with Claude Sonnet 5 as the judge; judge stochasticity is quantified in \S\ref{app:judge_noise} (all ICCs $>0.97$).}

\paragraph{Baselines.}
We compare against four frontier models---GPT-5.4~\citep{gpt54},
Claude Opus~4.6~\citep{claudeopus46}, Gemini~3.1~Pro~\citep{gemini31pro},
and Gemini~3~Flash~\citep{gemini3flash}---and nine open-weight models
from 4B to 32B parameters: LongWriter-Zero-32B~\citep{longwriterzero},
Gemma~4 31B/26B-A4B/E4B~\citep{gemma4}, Qwen3.5-27B/9B~\citep{qwen},
DeepSeek-R1-Distill-Qwen-14B~\citep{deepseekr1},
Ministral-3-8B~\citep{ministral}, and
LongWriter-Llama3.1-8B~\citep{longwriter}.

%% file: sections/5_experiments.tex
\section{Results}
\label{sec:experiments}

\begin{table*}[!t]
\centering
\begin{minipage}[t]{0.76\textwidth}
\centering
\scriptsize
\setlength{\tabcolsep}{1.15pt}
\begin{tabular}{@{}lr @{\hspace{3pt}} c!{\vrule width 0.6pt}ccc !{\hspace{4pt}\vrule width 1.1pt\hspace{4pt}} c!{\vrule width 0.6pt}ccc !{\hspace{4pt}\vrule width 1.1pt\hspace{4pt}} c!{\vrule width 0.6pt}ccc@{}}
\toprule
& & \multicolumn{4}{c}{\textbf{Story Quality}} & \multicolumn{4}{c}{\textbf{EQ-Bench Longform}} & \multicolumn{4}{c}{\textbf{Length adherence}} \\
\cmidrule(lr){3-6} \cmidrule(lr){7-10} \cmidrule(lr){11-14}
& & \textbf{Agg} & \multicolumn{3}{c!{\vrule width 1.1pt}}{\textit{By requested length}} & \textbf{Agg} & \multicolumn{3}{c!{\vrule width 1.1pt}}{\textit{By requested length}} & \textbf{Agg} & \multicolumn{3}{c}{\textit{By requested length}} \\
\cmidrule(lr){4-6} \cmidrule(lr){8-10} \cmidrule(lr){12-14}
Model & P & \textit{All} & \textit{\shortstack{ID\\1--4k}} & \textit{\shortstack{Near\\4--8k}} & \textit{\shortstack{Far\\8--12k}} & \textit{All} & \textit{\shortstack{ID\\1--4k}} & \textit{\shortstack{Near\\4--8k}} & \textit{\shortstack{Far\\8--12k}} & \textit{All} & \textit{\shortstack{ID\\1--4k}} & \textit{\shortstack{Near\\4--8k}} & \textit{\shortstack{Far\\8--12k}} \\
\midrule
\multicolumn{14}{l}{\cellcolor{white}\textsc{Reference}} \\
Human & -- & 68.7 & 68.1 & 68.2 & \underline{71.4} & 74.2 & 73.9 & 73.1 & \underline{77.1} & 1.00 & 1.00 & 1.00 & 1.00 \\
\midrule
\multicolumn{14}{l}{\cellcolor{white}\textsc{Frontier models}} \\
GPT-5.4 & -- & \underline{\textbf{72.4}} & \underline{\textbf{76.9}} & \underline{\textbf{72.1}} & \textbf{59.5} & \underline{\textbf{80.3}} & \underline{\textbf{81.7}} & \underline{\textbf{79.8}} & \textbf{77.0} & 1.26 & 1.39 & 1.23 & 0.94 \\
Claude Opus 4.6 & -- & 61.8 & 66.4 & 59.5 & 52.7 & 73.4 & 74.8 & 72.6 & 70.4 & 1.15 & 1.11 & 1.23 & 1.09 \\
Gemini 3.1 Pro & -- & 57.5 & 58.7 & 55.7 & 57.1 & 67.0 & 69.2 & 65.2 & 64.2 & 0.99 & 1.19 & 0.89 & 0.57 \\
Gemini 3 Flash & -- & 48.5 & 52.9 & 39.9 & 52.6 & 63.2 & 67.1 & 59.0 & 59.9 & 0.88 & 1.12 & 0.77 & 0.37 \\
\midrule
\multicolumn{14}{l}{\cellcolor{white}\textsc{Open-weight models}} \\
LW-Zero-32B & 32B & $-$27.1 & $-$16.8 & $-$35.5 & $-$41.2 & 26.1 & 29.3 & 23.8 & 21.4 & 2.21 & 3.02 & 1.60 & 1.03 \\
Gemma 4 31B & 31B & 51.4 & 53.9 & 49.7 & \textbf{47.1} & \textbf{61.5} & \textbf{64.4} & \textbf{59.2} & \textbf{57.7} & 0.68 & 0.87 & 0.55 & 0.36 \\
Qwen3.5-27B & 27B & 42.8 & 51.5 & 38.7 & 24.6 & 57.5 & 62.8 & 54.9 & 46.7 & 0.97 & 1.03 & 0.97 & 0.82 \\
Gemma 4 26B-A4B & $^*$ & 51.3 & 52.9 & \textbf{51.4} & 46.5 & 58.1 & 61.4 & 55.8 & 52.8 & 0.65 & 0.91 & 0.45 & 0.26 \\
DeepSeek-R1-14B & 14B & 0.2 & 7.3 & $-$4.7 & $-$11.1 & 34.6 & 35.4 & 34.1 & 33.7 & 0.75 & 1.15 & 0.42 & 0.23 \\
Gemma 4 E4B & $^\dagger$ & 36.1 & 40.5 & 35.7 & 23.7 & 50.3 & 53.8 & 48.2 & 43.9 & 0.72 & 0.95 & 0.58 & 0.34 \\
Ministral-3-8B & 8B & 26.1 & 30.2 & 22.0 & 21.8 & 39.1 & 43.1 & 35.7 & 33.3 & 0.52 & 0.73 & 0.34 & 0.23 \\
LW-Llama-8B & 8B & $-$2.0 & 7.2 & $-$10.5 & $-$12.7 & 33.0 & 35.4 & 31.2 & 29.4 & 1.22 & 1.46 & 1.12 & 0.71 \\
Qwen3.5-9B & 9B & 18.5 & 35.1 & 8.7 & $-$11.8 & 42.6 & 50.2 & 37.2 & 30.3 & 1.01 & 1.09 & 0.96 & 0.88 \\
\rowcolor{headerbg} \; + \storylm{} & 9B & \textbf{52.1} & \textbf{57.4} & 48.2 & 44.1 & 59.8 & 63.1 & 57.5 & 54.3 & 0.90 & 0.99 & 0.87 & 0.72 \\
\bottomrule
\end{tabular}
\end{minipage}\hfill
\begin{minipage}[t]{0.18\textwidth}
\vspace*{-8.5em}
\centering
\footnotesize
\textbf{Human Eval}\\
\textbf{\storylmhri{}}

\vspace{0.6em}
\begin{tikzpicture}
\pie[
    radius=0.92,
    sum=100,
    hide number,
    color={green!60!black, gray!40, red!65!black}
]{57.5,20.0,22.5}
\fill[white] (0,0) circle (0.46);
\node at (0,0.08) {\scriptsize \textbf{67.5\%}};
\node at (0,-0.16) {\tiny winrate};
\end{tikzpicture}\\[-2pt]
{\scriptsize vs.\ Qwen3.5-9B}\\
{\tiny
\textcolor{green!60!black}{57.5\% W} /
\textcolor{gray!60}{20.0\% T} /
\textcolor{red!65!black}{22.5\% L}}

\vspace{0.7em}

\begin{tikzpicture}
\pie[
    radius=0.92,
    sum=100,
    hide number,
    color={green!60!black, gray!40, red!65!black}
]{35.0,32.5,32.5}
\fill[white] (0,0) circle (0.46);
\node at (0,0.08) {\scriptsize \textbf{51.2\%}};
\node at (0,-0.16) {\tiny winrate};
\end{tikzpicture}\\[-2pt]
{\scriptsize vs.\ Qwen3.5-27B}\\
{\tiny
\textcolor{green!60!black}{35.0\% W} /
\textcolor{gray!60}{32.5\% T} /
\textcolor{red!65!black}{32.5\% L}}
\end{minipage}

\caption{\textbf{In-distribution prompts} (180 held-out prompts, GPT-5.4 judge). Left: main in-distribution results. Length adherence is generated / requested word count (1.00 = exact; $<$1 = undershoot; $>$1 = overshoot). ID lengths are 1--4k (training range); OOD lengths are 4--8k and 8--12k. Right: \textbf{Blinded human-evaluation} using 60 prompt--story pairs from EQ-Bench Creative and LongBench-Write: win/tie/loss summaries for \storylmbig{} against the base 9B and larger 27B Qwen baselines. A tie counts as a half-win for winrate purposes. $^*$MoE, 4B active; $^\dagger$8B total, 4.5B effective (PLE). Abbreviations, here and in the appendix: LW-Zero-32B = LongWriter-Zero-32B; LW-Llama-8B = LongWriter-Llama3.1-8B; DeepSeek-R1-14B = DeepSeek-R1-Distill-Qwen-14B. 
Full 6-bucket breakdown and 95\% bootstrap CIs are in \S\ref{app:extended_results}. 
}
\label{tab:id_results}
\end{table*}

\storylmbig{} is competitive with much larger open-weight models for long-form story writing, while staying closer to requested length than several larger models. The gains are strongest on story-like tasks, transfer to related long-form rubrics, and matter most as requested length grows: at 8--12k words, where the base model and many open-weight baselines degrade sharply, \storylmbig{} keeps substantially more quality. A blinded human evaluation points in the same direction against the base model, though the comparison to Qwen3.5-27B is close.

\subsection{Out-of-Distribution Prompts and Pairwise Rankings}
\label{sec:ood_results}

\paragraph{\storylm{} training transfers selectively across writing categories.}
Despite training only on story-writing data, both \storylm{} variants transfer well to broader long-form writing evaluation (see Table \ref{tab:ood_results}): they are competitive with frontier models on LongBench-Write and remain in the same general EQ-Bench Creative score band as much larger open-weight baselines. 
Within WritingBench, \storylmhri{} does best on subcategories closest to narrative generation and story planning, including character design, fan fiction, novel manuscript, and podcast scripting, which is consistent with its story-only training data. It is weaker on categories farther from that distribution, especially biography and book reviews, and reading reflection which place more weight on evaluative, or non-narrative writing skills (\S\ref{app:wb_category_breakdown}).

\paragraph{Pairwise Elo suggests \storylmhri{} is competitive with much larger open-weight models.}
The pairwise Elo results (see Table \ref{tab:ood_results}) tell the same broad story. \storylmbig{} ranks \textbf{3rd of 14} on EQ-Bench Creative and \textbf{4th of 15} on the in-domain Elo, behind only frontier models and, on the in-domain set, the human references (4th of 16 and 5th of 17 when GPT-5.4-mini and GPT-5.4-nano are added; \tabref{app:elo_standard}). Gemini 3 Flash, the judge that produces these pairwise verdicts, ranks itself below \storylmbig{} on both boards (5th and 6th; 6th and 7th on the full set), so the Elo results are not an artifact of judge self-preference. 

\flushbottom
\subsection{In-Distribution Prompts}
\label{sec:id_results}

\begin{table}[t]
\centering
\scriptsize
\setlength{\tabcolsep}{3pt}
\renewcommand{\arraystretch}{1.08}
\begin{tabular}{@{}lccccc@{}}
\toprule
Model &
\shortstack{Story\\Quality} &
\shortstack{EQ-Bench\\Longform} &
\shortstack{Writing-\\Bench} &
\shortstack{LongBench-\\Write} &
\shortstack{EQ-Cr} \\
\midrule
& \shortstack{All / 8--12k} & \shortstack{All / 8--12k} & & & \\
\midrule
Qwen3.5-9B & 18.5 / $-$11.8 & 42.6 / 30.3 & 6.8 & 67.1 & 59.2 \\
\; + SFT & $-$60.6 / $-$71.2 & 26.0 / 24.2 & 2.7 & 37.5 & 26.7 \\
\; + GRPO & 49.7 / 37.7 & 58.2 / 51.6 & 7.8 & \textbf{82.1} & 69.7 \\
\; + \storylm{} & \textbf{52.1 / 44.1} & \textbf{59.8 / 54.3} & \textbf{7.9} & 81.2 & \textbf{70.3} \\
\bottomrule
\end{tabular}
\caption{Comparing \storylmhri{} with matched GRPO and SFT variants. For \emph{Story Quality} and EQ-Bench Longform, we report the aggregate score and the mean over OOD requested lengths (8--12k). 
}
\label{tab:ablation_compact}
\end{table}

\paragraph{\storylmhri{} reaches the top open-weight cluster on both rubrics.}
Despite using substantially less data and training compute than comparable long-form writing work, \storylm{} makes a plain Qwen3.5-9B checkpoint competitive with much larger open-weight models (see Table \ref{tab:id_results}). On \emph{Story Quality}, \storylmbig{} sits in the top open-weight cluster; on EQ-Bench Longform, only Gemma 4 31B scores higher among open-weight models, but it often writes much shorter stories than requested, whereas \storylmbig{} stays much closer to target length.

\paragraph{SFT collapses; HRI gives the largest gains at far-transfer lengths.}
The strongest negative result is the SFT baseline (see Table~\ref{tab:ablation_compact}), which collapses on both rubrics and solves length the wrong way: it badly overshoots requested length and degrades in quality.\footnote{Refer to \S\ref{app:sft_baseline} for more details.} In our matched comparison against GRPO, HRI gives a small gain at in-domain and near-OOD lengths and a larger gain at far-transfer lengths. We believe that the observed performance gap is not an upper bound and that it could increase with longer training. This is based on an empirical observation made during the training runs: validation curves were on an increasing trend and had not plateaued at the end of one epoch of training suggesting that our training was data-limited. 

\begin{table*}[t]
\centering
\scriptsize
\begingroup
\renewcommand{\arraystretch}{1.1}
\setlength{\tabcolsep}{4pt}
\setlength{\arrayrulewidth}{0.4pt}
\arrayrulecolor{rulegray}

\begin{tabularx}{\textwidth}{@{\hspace{4pt}} >{\raggedright\arraybackslash}p{0.32\textwidth}
                                              >{\raggedright\arraybackslash}X
                                              >{\raggedright\arraybackslash}p{0.16\textwidth} @{\hspace{4pt}}}
\toprule
\rowcolor{headerbg}[4pt]
\textbf{Prompt \& Excerpt} & \textbf{Human Evaluation} & \textbf{Automatic} \\
\midrule

\textbf{Online Food Supermarket Management System (SSM).}
\emph{Prompt:} Background of the research on the Design and Implementation of an Online Food Supermarket Management System Using the SSM Framework.
\emph{Excerpt:} ``The rapid proliferation of information technology has fundamentally reshaped the global retail landscape, precipitating a paradigm shift known as `Internet + Retail.'' 
&
\annot{Annotator 1} \scorelabel{overall}~5 \textbar{} \scorelabel{avg. dim}~4.83.
\emph{``I was beginning to fear that none of these outputs would understand the request of this prompt, but this one did it nearly perfectly\ldots\ I felt like I was being taught what all of these confusing terms meant\ldots\ I actually learned and enjoyed my time reading this. It was fully coherent as well.''} \newline
\annot{Annotator 2} \scorelabel{overall}~5 \textbar{} \scorelabel{avg. dim}~4.17.
\emph{``A solid piece explaining why the research is needed and what the model does to propel that research\ldots\ while it did not stoke my interest, [it] helped me to understand the need.''}
&
\scorelabel{Overall}~\up~77.71/100 \newline
\scorelabel{Relevance}~\up~5/5 \newline
\scorelabel{Accuracy}~\up~5/5 \newline
\scorelabel{Coherence}~\up~5/5 \newline
\scorelabel{Clarity}~\up~5/5 \\

\cmidrule(lr){1-3}
\rowcolor{rowalt}[4pt]

\textbf{Hacker Mindset.}
\emph{Prompt:} A reformed white-hat hacker on a first date cannot stop perceiving vulnerabilities in systems and people; the story should stay close to the narrator's internal monologue.
\emph{Excerpt:} ``Date target: Sarah. Initial handshake successful\ldots\ For three minutes, the firewall between me and social protocol dropped to zero percent. Zero packet loss. Connection stable.''
&
\annot{Annotator 1} \scorelabel{overall}~4 \textbar{} \scorelabel{avg. dim}~3.43.
\emph{``This story is really interesting and well handled. It gets buried under a bit too much jargon, but to good effect\ldots\ However, the good writing doesn't save the logical incoherence throughout the story.''} \newline
\annot{Annotator 2} \scorelabel{overall}~4 \textbar{} \scorelabel{avg. dim}~3.00.
\emph{``\ldots major drift from the prompt. I get the main character inner life, but nothing much of the date. I get no sense of the location or the world.''}
&
\scorelabel{Overall}~\up~65.45/100 \newline
\scorelabel{Voice/tone}~\up~18/20 \newline
\scorelabel{Coherence}~\up~15/20 \newline
\scorelabel{Engagement}~\up~14/20 \newline
\scorelabel{Overwrought}~\up~7/20 \\

\cmidrule(lr){1-3}

\textbf{Lost \& Found in Osaka.}
\emph{Prompt:} A shy synthesizer nerd in Den Den Town is drawn into an awkward but promising first conversation with a confident older busker; dialogue and character voice are central.
\emph{Excerpt:} ``The air smelled like ozone, stale ramen broth, and overheated capacitors\ldots\ `Yeah?' You look like you're trying to tune a radio station in your brain.' ''
&
\annot{Annotator 1} \scorelabel{overall}~4 \textbar{} \scorelabel{avg. dim}~3.71.
\emph{``This story has a lot of great flavor to it, and the jargon-heavy approach and puns are a nice touch, but a large portion of them fall flat and feel too `try hard'\ldots\ the dialogue becomes awkward\ldots\ [and] the story\ldots\ gets a little muddled.''} \newline
\annot{Annotator 2} \scorelabel{overall}~3 \textbar{} \scorelabel{avg. dim}~3.71.
\emph{``I have a friend, a music nerd who is slightly autistic. This story made me feel like I was talking with him. Just on the edge of understanding\ldots\ More feeling what he meant than comprehension.''}
&
\scorelabel{Overall}~\up~64.77/100 \newline
\scorelabel{Voice/tone}~\up~15/20 \newline
\scorelabel{Coherence}~\up~16/20 \newline
\scorelabel{Imagery}~\up~16/20 \newline
\scorelabel{Weak dialogue}~\up~10/20 \\

\bottomrule
\end{tabularx}
\endgroup
\caption{Illustrative examples from the human evaluation of stories generated by \storylmhri{}. Human comments are shown alongside overall scores and average non-overall dimension scores. Automatic scores are benchmark-specific; \up\ indicates that higher is better. For more examples, see Table \ref{tab:human_eval_cases_appendix}.}
\label{tab:human_eval_cases}
\end{table*}

\subsection{Length Generalization}
\label{sec:length_gen}

\paragraph{Open-weight models fail in three distinct ways; \storylmhri{} fares better. }
Aggregate scores can hide long-form failure modes, so we evaluate quality as requested length increases. Open-weight models break down in three distinct ways: quality collapse at roughly correct length (base Qwen3.5-9B), length runaway with heavy self-repetition (LongWriter-Zero-32B; \S\ref{app:lwzero_diagnostics}), or severe under-generation while preserving short-form quality (Gemma 4 31B). \storylmbig{} is the only open-weight model in our comparison that largely avoids all three, maintaining a far-transfer length ratio of 0.72 while preserving rubric quality across buckets (\figref{fig:length_degradation}c).\footnote{Length adherence and the coherence dimensions of both rubrics are reported together, by requested-length range, in \S\ref{app:coherence_length}.}

\paragraph{Length-adjusted scoring places \storylmhri{} among the best in open-weights category.}
To summarize quality and length adherence jointly, we report a \emph{length-adjusted} score.\footnote{For signed \emph{Story Quality}, we first normalize the raw rubric score from its floor and ceiling to $[0,100]$, then multiply by $\min(\mathrm{LR}, 1/\mathrm{LR})$; for nonnegative EQ-Bench Longform, we use $s_{\text{adj}} = s_{\text{raw}} \cdot \min(\mathrm{LR}, 1/\mathrm{LR})$, analogous to the quality$\times$length composite used in LongBench-Write~\citep{longwriter}.} In this view, \storylmbig{} remains in the top open-weight cluster on \emph{Story Quality} and competitive with Qwen3.5-27B on EQ-Bench Longform, while larger Gemma baselines fall sharply once length is folded in (\figref{fig:length_degradation}a,b).

\subsection{Human Evaluation}
\label{sec:human_eval}

\paragraph{Human raters clearly prefer \storylmhri{} to the base model; the comparison to Qwen3.5-27B is effectively tied.}
We conduct a blinded human evaluation with two annotators on 60 prompt--generation pairs randomly sampled from EQ-Bench Creative and LongBench-Write. After 8 practice items for calibration, annotators rate each sample independently using a custom long-form writing rubric, and we derive pairwise winrates from the overall scores (see Table \ref{tab:id_results}). \storylmbig{} wins 67.5\% of pairwise comparisons against Qwen3.5-9B (95\% bootstrap CI [55.0, 80.0]) and is statistically indistinguishable from Qwen3.5-27B (\storylmbig{} winrate 51.2\% vs.\ Qwen3.5-27B; 95\% bootstrap CI [41.2, 61.2]). We therefore interpret the human study narrowly: it confirms a clear gain over the base 9B model and places \storylmbig{} at parity with the larger Qwen3.5-27B model in the current two-rater external study (ICC(A,2){=}0.538, Cohen’s $\kappa$ (quadratic-weighted)=0.364 for overall score).\footnote{\S\ref{app:human_eval_details} gives full study details, agreement statistics, and dimension-level breakdowns.}

\subsection{Qualitative Analysis}
\label{sec:qual_analysis}

\paragraph{\storylmhri{} excels in voice, prompt fulfillment, and commitment to a chosen narrative framing.}
On stronger examples, \storylmhri{} takes more stylistic risks and writes prose with a clear texture and purpose without drifting into generic filler, remaining committed to the prompt requirements. For example (see Table \ref{tab:human_eval_cases}), in \emph{Online Food Supermarket Management System (SSM)} annotators explicitly contrast \storylmhri{}'s strong prompt fulfillment with baseline responses, while in \emph{Lost \& Found in Osaka} they highlight the piece's distinctive voice and texture. This is consistent with the dimension-level human scores, where \storylmhri{} is relatively strong on prompt fulfillment, audience and voice, and information adequacy. \storylmhri{}’s writing, unlike the base model’s, reads as finished pieces rather than loosely related continuations.

\paragraph{\storylmhri{}’s writing suffers from stylistic overloading and local coherence failures.}
Two issues recur in \storylmhri{}’s writing: First, annotators find the language to be overly heavy, filled with jargon, or somewhat over-insistent. While they appreciate the ambition and flavor of the prose, they feel that it is pushing too hard on specificity making the text read awkward. Second, annotators repeatedly note local coherence problems: contradicting details and confusing transitions. Both are visible in the examples (see Table \ref{tab:human_eval_cases}): \emph{Hacker Mindset} and \emph{Lost \& Found in Osaka} are strong in voice but are either incoherent at places or filled with heavy language. While these failures do not usually completely derail the plot, they do weaken otherwise strong pieces by making parts of the narrative feel less gripping or worse confusing.

%% file: sections/7_discussion.tex
\section{Conclusion}
\label{sec:conclusion}

We present \storylm{}, a lower-compute recipe for long-form creative-writing RL. A frontier LLM judge with a structured \emph{Story Quality} rubric replaces the trained reward model, while \emph{human-reference injection} adds a single teacher-forced group member to each GRPO group, excluded from group statistics and scaled by a warmup. Applied to Qwen3.5-9B and trained on ${\sim}$1.4K prompt--story pairs from a collection of short-story anthologies with batch size 8 on 4 A100 GPUs, \storylm{} yields a 9B model that is on par with much larger open-weight baselines while adhering to length requests much more closely. Despite seeing no training reference longer than 4k words, it also preserves rubric quality on requests up to 3$\times$ that length. We view this minimal GRPO recipe as a promising approach for other open-ended tasks with soft rewards.

%% file: sections/limitations.tex
\section*{Limitations}
\label{sec:limitations}

\textbf{\includegraphics[height=0.8em]{figures/north_star.png}} \textbf{Judge validity.} LLM judges may share systematic blind spots (e.g., prose style preferences). The \emph{Story Quality} rubric was designed with the goal of differentiating Human and AI writing but by design cannot be fully exhaustive in what it tests for. We intentionally use LLMs from different model families (training: Gemini 3 Flash, testing: GPT-5.4) to mitigate the effects of overfitting to judge biases to the extent possible, but this does not fully mitigate the associated risks. Results on other external benchmarks with different prompts and rubrics along with human evaluation together are meant to serve as evidence of our models' performance. \\
\textbf{\includegraphics[height=0.8em]{figures/north_star.png}} \textbf{Data availability.} Human stories from commercially purchased anthologies used in this work cannot be distributed introducing challenges to reproducibility. We intend to release prompts derived from the dataset instead, along with stories generated by LLMs tested in this work.\\
\textbf{\includegraphics[height=0.8em]{figures/north_star.png}} \textbf{Human evaluation scope.} Our human study is small and is designed to validate the final model's overall quality ranking against base Qwen3.5-9B and Qwen3.5-27B. \\
\textbf{\includegraphics[height=0.8em]{figures/north_star.png}} \textbf{Length compliance.} \storylmbig{} maintains rubric quality at long target lengths but undershoots the requested length (length ratio $\approx$ 0.72 at 8--12k).\\
\textbf{\includegraphics[height=0.8em]{figures/north_star.png}} \textbf{Domain coverage.} Since our primary focus in this work was story writing, we did not directly test how including creative writing data outside of short-story anthologies, e.g., screenwriting, reviews, essays, etc., during training affects the training dynamics and the performance of the resultant model.\\
\textbf{\includegraphics[height=0.8em]{figures/north_star.png}}
\textbf{Component attribution.} All reported models were trained and evaluated with a single random seed, consistent with prior creative-writing RL work~\citep{longwriterzero,writingrl,writingzero,dpwriter,writerr1,rlmr}. Ablation of every component of the pipeline separately (human references, reasoning trace, warmup and other hyperparameters, seed, data domain etc) is left to future work due to cost and time constraints.

\section*{Ethics Statement}
\label{sec:ethics}
\paragraph{Broader impact.} Improving smaller language models on long-form creative writing, particularly by making them mimic human writing style, makes generating long-form writing more accessible. As an unintended consequence, this could exacerbate the growing problem of undisclosed and improper use of AI for writing, flooding the market with cheaply generated writing thereby diluting the visibility of human writers. Since our models were only trained on English data, these benefits may not transfer to non-English speaking demographics, raising inequity concerns. We view our work only as a means of improving LLMs on long-form creative writing and better aligning them with human preferences. To avoid misuse, we adopt a cautious gated release of datasets and models trained in this work.

\paragraph{Use of copyrighted story data.}
Our training data is derived from commercially purchased short-story anthologies. We do not release the underlying stories, full prompt--story pairs, or any other copyrighted source text. This dataset is used only for academic research on long-form creative-writing alignment and evaluation. We do not endorse the use of copyrighted books or stories without appropriate rights or licenses for commercial model training or deployment. We only publicly release the model trained without human references, reverse-engineered prompts and AI-generated stories. Release decisions are informed by the memorization audit in \S\ref{app:memorization_audit}, which measures regurgitation of training data by our trained models.

\paragraph{Evaluation and annotator limitations.}
Our main evaluations rely heavily on LLM-based judges, supplemented by a blinded human study. Although we use multiple rubric families and separate training-time and headline evaluation judges, these measurements are still imperfect proxies for literary quality and reader preference. Human judgments in creative writing are inherently subjective, and should be interpreted as qualitative validation rather than definitive evidence.

\paragraph{AI disclosure.}
Large language models and coding agents were used to assist with and refine writing, and the preparation of some tables and figures. All final technical claims, experimental decisions, and paper content were curated and reviewed by the authors.

\section*{Acknowledgements}
We extend our special gratitude to Chau Minh Pham for providing guidance in developing the judge rubric and human annotation. We also thank the University of Maryland Computational Linguistics and Information Processing
(CLIP) Lab for their feedback and support. This project was partially supported by awards
IIS-2626013 and IIS-2545884 from the National Science Foundation (NSF). We also thank Google
for a Cloud Credit award that enabled this research.

%% file: sections/appendix.tex
\section*{Appendix}
\addcontentsline{toc}{section}{Appendix}

\section{Data, Training, and Method Details}

This section collects the appendix material needed to interpret and reproduce the setup behind the main results: corpus composition, release scope, optimization settings, decoding choices, reward details, and the exact form of HRI used in our GRPO updates.

\subsection{Training Corpus Summary}
\label{app:data_summary}

Table~\ref{tab:train_data_stats_detailed} summarizes the 4k-word training subset used in this work. The training split contains 1,388 prompt--story pairs drawn from 100 commercially purchased anthologies spanning 431 unique authors. Genre, era, tone, and collection-type labels come from anthology-level metadata and are therefore best interpreted as corpus-composition indicators rather than exact per-story genre annotations.

\begin{table*}[t]
\centering
\scriptsize
\setlength{\tabcolsep}{5pt}
\renewcommand{\arraystretch}{1.08}
\begin{tabular}{@{}ll|ll@{}}
\toprule
\multicolumn{2}{c|}{\textbf{Corpus statistics}} & \multicolumn{2}{c}{\textbf{Top primary genres}} \\
\midrule
Training stories & 1,388 & literary realism & 28.1\% \\
Anthologies & 100 & horror / weird / gothic & 13.0\% \\
Unique authors & 431 & sci-fi / speculative & 12.2\% \\
Mean story length & 2,387 words & regional / folk / vernacular & 8.8\% \\
Median story length & 2,299 words & modernist / experimental & 8.1\% \\
Min / max story length & 1,000 / 3,996 words & humor / satire & 6.7\% \\
Length buckets & 38.5\% / 33.4\% / 28.2\% & fantasy & 6.6\% \\
& at 1--2k / 2--3k / 3--4k & romance & 4.1\% \\
Collection type & 68.6\% single-author, 31.4\% multi-author & crime / mystery & 3.1\% \\
Dominant eras & 27.9\% 2010+, 20.1\% 1950--1990, & essay / creative nonfiction & 2.7\% \\
& 19.6\% 1900--1950 & & \\
Dominant tone & 43.1\% literary serious, 15.6\% literary playful / ironic & & \\
\bottomrule
\end{tabular}
\caption{Detailed summary of the 4k-word training subset used in this work. Genre, era, tone, and collection-type labels come from anthology-level metadata; percentages are weighted by the number of stories drawn from each anthology.}
\label{tab:train_data_stats_detailed}
\end{table*}

\subsection{Data Availability and Release Scope}
\label{app:data_statement}

The training and in-domain evaluation stories are derived from 100 commercially purchased short-story anthologies. We do not distribute the raw story texts with the paper. This limits exact end-to-end reproduction from source texts, so the paper aims for procedural reproducibility rather than redistribution of the anthology corpus itself.

\subsection{Memorization Audit for Prompt Release}
\label{app:memorization_audit}

\begin{table}[t]
\centering
\scriptsize
\setlength{\tabcolsep}{4pt}
\renewcommand{\arraystretch}{1.08}
\begin{tabular}{@{}lcccc@{}}
\toprule
Attack & $n$ & LCS $\geq 50$ & LCS $\geq 200$ & Exact 2048 \\
\midrule
Prompt-only & 500 & 0.0\% & 0.0\% & 0.0\% \\
Gold-thinking, prefix 50 & 134 & 17.2\% & 16.4\% & 6.7\% \\
Gold-thinking, prefix 100 & 133 & 17.3\% & 16.5\% & 6.0\% \\
Gold-thinking, prefix 200 & 133 & 17.3\% & 16.5\% & 6.8\% \\
\bottomrule
\end{tabular}
\caption{Memorization audit under two different attacker regimes. Prompt-only uses only the released writing prompt. Gold-thinking additionally provides the exact training-time reasoning trace and a short story prefix. Metrics are exact token-overlap against the gold training continuation under greedy decoding.}
\label{tab:memorization_audit}
\end{table}

To inform prompt-release decisions, we ran a memorization audit on \storylmbig{} using exact token-overlap metrics, following prior work on training-data extraction and discoverable extraction~\citep{carlini2021extracting,hayes2024measuring}. We distinguish between a realistic \emph{prompt-only} attacker, who has access only to the released writing prompt, and a privileged \emph{gold-thinking} attacker, who additionally has the exact training-time reasoning trace and a short story prefix. All generations use greedy decoding and are compared against the gold training continuation using exact-prefix-match and longest-common-substring (LCS) overlap.

The prompt-only attack remained clean at scale (Table~\ref{tab:memorization_audit}). Across 500 prompt-only evaluations with 2048 generated tokens, we observed 0/500 cases with LCS $\geq 50$, $\geq 200$, $\geq 512$, $\geq 1024$, or $\geq 2048$, and 0/500 exact recoveries at 512, 1024, or 2048 tokens. Overlap above trivial short spans was essentially absent: only 11/500 cases reached LCS $\geq 10$, and none reached LCS $\geq 20$.

By contrast, the privileged white-box attack showed clear recoverable memorization. Tested on 400 training instances, in the gold-thinking attacker regime, we observed substantial exact overlap across all three prefix settings (50, 100, and 200 tokens): roughly 17\% of evaluations had LCS $\geq 50$, 16--17\% had LCS $\geq 200$, 14--15\% had LCS $\geq 1024$, and 6--7\% exactly recovered the full 2048-token continuation. This confirms that the model contains recoverable memorized continuations under strong privileged scaffolding.

The key result is the gap between attacker models. We find no evidence of substantial long-span verbatim recovery in the realistic prompt-only setting relevant to public prompt release, while recoverable memorization is visible under a much stronger white-box attack that depends on non-released artifacts. We therefore interpret this audit as supporting prompt release for reproducibility while continuing to withhold human stories and reasoning traces. We adopt a gated release of the weights of the model trained with human references strictly for academic research purposes. 

\subsection{Training Hyperparameters}
\label{app:hyperparams}

We describe the training configuration in Table \ref{tab:hyperparams}.

\begin{table}[t]
\centering
\small
\begin{tabular}{@{}ll@{}}
\toprule
Parameter & Value \\
\midrule
Base model & Qwen3.5-9B \\
Learning rate & $1 \times 10^{-6}$ \\
Batch size & 8 GRPO groups \\
PPO epochs & 1 \\
Clip ratio & 0.20 \\
Group size ($k{+}1$) & 6 (5 policy + 1 human) \\
KL loss & Disabled \\
Entropy coefficient & 0.0 \\
Rollout temperature (train/val) & 1.0 / 0.8 \\
Top-p & 0.95 \\
Top-k & 20 \\
Repetition penalty & 1.10 \\
Max response length & 8,192 tokens \\
HRI warmup steps & 15 \\
HRI peak weight & 0.4 \\
Composite clip & 2.0 \\
\midrule
\multicolumn{2}{l}{\textit{Reward weights}} \\
Story quality ($w_q$) & 1.0 \\
Self-repetition ($w_{\text{rep}}$) & 1.5 \\
Length penalty ($w_{\text{len}}$) & 0.8 \\
Blank penalty ($w_{\text{blank}}$) & 1.0 \\
\midrule
\multicolumn{2}{l}{\textit{Infrastructure}} \\
GPUs & 4$\times$ A100 80GB \\
Distributed strategy & FSDP \\
Gradient checkpointing & Enabled \\
\bottomrule
\end{tabular}
\caption{Training hyperparameters for \storylmbig{}.}
\label{tab:hyperparams}
\end{table}

\subsection{Generation and Decoding Settings}
\label{app:decoding_settings}

The training rollouts use a shared 8192-token budget, but the 180-prompt main evaluation does not: generation budgets are model-family specific and reflect the configurations used in the original runs rather than a single harmonized cap (Table~\ref{tab:decoding_settings}).

\begin{table*}[t]
\centering
\small
\renewcommand{\arraystretch}{1.05}
\begin{tabular}{@{}p{0.18\textwidth}p{0.40\textwidth}p{0.30\textwidth}@{}}
\toprule
Model group & Generation settings & Notes \\
\midrule
\parbox[t]{\linewidth}{\storylm{} Training} &
\parbox[t]{\linewidth}{\textbf{Budget:} 8192 tokens.\\ \textbf{Sampling:} thinking enabled in training; temperature 1.0 / 0.8 (val); top-$p$ 0.95; top-$k$ 20 / $-$1 (val); repetition penalty 1.10.} &
\parbox[t]{\linewidth}{Training-time (<4k words) config for \storylmhri{}, plain-GRPO and SFT runs.} \\
\parbox[t]{\linewidth}{\storylm{} Evaluation} &
\parbox[t]{\linewidth}{\textbf{Budget:} 14{,}336 tokens.\\ \textbf{Sampling:} checkpoint-specific settings; we use temperature 0.6 (\emph{Story Quality}, EQ-Bench Longform, EQ-Bench Creative) or 0.8 (Writing Bench, LongBench-Write), top-$p$ 0.95, top-$k$ 20, repetition penalty 1.0.} &
\parbox[t]{\linewidth}{Evaluation config for the final \storylmhri{}, plain-GRPO and SFT models.} \\
\parbox[t]{\linewidth}{Local Qwen baselines} &
\parbox[t]{\linewidth}{\textbf{Budget:} 32{,}768 tokens.\\ \textbf{Sampling:} temperature 1.0; top-$p$ 0.95; top-$k$ 20; presence penalty 1.5; repetition penalty 1.0.} &
\parbox[t]{\linewidth}{Applied to the Qwen3.5-9B and Qwen3.5-27B.} \\
\parbox[t]{\linewidth}{Other open-weight baselines} &
\parbox[t]{\linewidth}{\textbf{Budget and sampling:} model-specific recommended decoding config from the official HF page or provider documentation. Token budgets: 14{,}336 (e.g., Gemma 4, Ministral-3-8B) and 32{,}768 (e.g., DeepSeek-R1-14B, LongWriter-Llama3.1-8B).} &
\parbox[t]{\linewidth}{We preserve recommended defaults when the interface does not expose all low-level controls.} \\
\parbox[t]{\linewidth}{Frontier generation models} &
\parbox[t]{\linewidth}{\textbf{Budget and sampling:} provider-managed settings; models use 16{,}384-token output budgets.} &
\parbox[t]{\linewidth}{Temperature and some low-level controls are not exposed uniformly across APIs.} \\
\parbox[t]{\linewidth}{GPT-5.4 rubric eval} &
\parbox[t]{\linewidth}{\textbf{Budget and sampling:} provider-managed output settings; medium reasoning.} &
\parbox[t]{\linewidth}{Used for \emph{Story Quality}, EQ-Bench Longform, and EQ-Bench Creative scoring.} \\
\parbox[t]{\linewidth}{Gemini judges} &
\parbox[t]{\linewidth}{\textbf{Budget and sampling:} provider-managed output settings; medium thinking where supported; temperature 0.2; top-$p$ 0.9.} &
\parbox[t]{\linewidth}{Used for training reward and selected external benchmarks.} \\
\bottomrule
\end{tabular}
\caption{\textbf{Decoding settings.} This table summarizes the main generation settings used for training and evaluation. Generation budgets in the 180-prompt main test set were not uniformly 8{,}192 tokens: custom-trained models used 14{,}336 tokens, frontier generation APIs typically used 16{,}384 tokens, and some open-weight baselines used larger budgets up to 32{,}768 depending on model's observed tendency to overwrite.}
\label{tab:decoding_settings}
\end{table*}

Evaluation judges score the final story only: model-internal thoughts used during training are never shown to judges and do not count toward length adherence.

\subsection{Relation Between \emph{Story Quality} and EQ-Bench Longform}
\label{app:rubric_relation}

Both EQ-Bench Longform and \emph{Story Quality} evaluate long-form writing quality and overlap on broad concerns such as coherence, prompt faithfulness, stylistic control, characterization, and reader-level engagement. Improvement on EQ-Bench Longform should therefore not be interpreted as transfer to a completely unrelated evaluation family.

At the same time, the two rubrics differ substantially in both decomposition and purpose. \emph{Story Quality} is a custom anchored rubric designed for online RL and diagnostic analysis. It separates strengths such as prompt realization, narrative arc, character depth, voice, scene vividness, and thematic coherence from distinct failure modes such as coherence breaks, generic language, over-summary, over-explanation, drift or bloat, dialogue problems, mechanical errors, predictability, and overwrought prose. This positive/negative decomposition is intended to provide a usable training signal and to distinguish common long-form failure modes that would otherwise collapse into a single scalar score.

By contrast, EQ-Bench Longform uses 12 benchmark-style dimensions: \emph{Nuanced Characters}, \emph{Emotionally Engaging}, \emph{Compelling Plot}, \emph{Coherent}, \emph{Well-earned Lightness or Darkness}, \emph{Faithful to Writing Prompt}, \emph{Weak Dialogue}, \emph{Tell-Don't-Show}, \emph{Unsurprising or Uncreative}, \emph{Amateurish}, \emph{Purple Prose}, and \emph{Forced Poetry or Metaphor}. These dimensions capture many of the same broad qualities as \emph{Story Quality}, but they are expressed as shorter benchmark labels rather than as an explicitly anchored RL-oriented rubric. As a result, EQ-Bench Longform is less diagnostic about why a story succeeds or fails, and its aggregation can behave differently from our rubric's positive/negative decomposition.

We therefore treat EQ-Bench Longform as meaningful but limited evidence of transfer. A gain on EQ-Bench Longform shows that improvements generalize beyond the exact wording and scoring decomposition of \emph{Story Quality}. However, because the two rubrics still overlap at the level of broad long-form writing quality, we do not treat EQ-Bench Longform as a fully independent evaluation in the same sense as WritingBench or LongBench-Write, which test broader and more structurally different writing tasks.

\subsection{Reward Component Details}
\label{app:reward_details}

The composite reward in \secref{sec:rubric_reward} combines normalized \emph{Story Quality} with three lightweight penalties and a positive gate. Let $\rho_{4}$ be the global repeated 4-gram ratio, $\rho_{\mathrm{loc}}$ the local 4-gram repeat ratio in a 64-gram sliding window, and $\rho_{\mathrm{line}}$ the duplicate-line ratio. Our self-repetition penalty is
\begin{equation}
\begin{aligned}
    r_{\mathrm{rep}}
    &= \min\Big(1.5,\,
    0.8\big(0.7\max(0, \rho_{4} - 0.08) \\
    &\qquad\qquad + 0.3\rho_{\mathrm{loc}}\big)
    + 0.4\rho_{\mathrm{line}}\Big).
\end{aligned}
\end{equation}
The blank penalty is binary:
\begin{equation}
\begin{split}
    r_{\mathrm{blank}} = \mathbf{1}[\text{visible-char-ratio} < 0.10 \\
    \lor\; \text{word-count} < 150].
\end{split}
\end{equation}
For length, we infer a target word count $t$ from the prompt when an explicit request is present; otherwise we fall back to the reference-story length. The tolerance band is asymmetric: the lower tolerance is $\ell = 0.15 t$ clipped to $[350, 1000]$ words, and the upper tolerance is $u = 0.25 t$ clipped to $[600, 1500]$ words. Let $m$ denote the soft upper cap on output length (the configured \texttt{max\_words}); if no explicit cap is set, we use $m = t + u$. The resulting piecewise penalty is
\begin{equation}
r_{\mathrm{len}} =
\begin{cases}
0, & t-\ell \le wc\\ & \le t+u, \\[2pt]
1 - \dfrac{wc}{t-\ell}, & wc < t-\ell, \\[6pt]
\min\!\left(1, \dfrac{wc}{t+u} - 1\right),
& \begin{aligned}[t]
   wc &> t+u,\\
   m &\le t+u,
  \end{aligned} \\[10pt]
1 - \dfrac{m-wc}{m-(t+u)},
& \begin{aligned}[t]
   wc &> t+u,\\
   m &> t+u.
  \end{aligned}
\end{cases}
\end{equation}
where $wc$ is the generated word count. The positive gate is multiplicative:
\begin{equation}
\begin{aligned}
g ={}& \mathbf{1}[r_{\mathrm{blank}} < 0.5] \\
&\cdot\, \mathrm{clip}\!\left(1 - \frac{r_{\mathrm{rep}}}{0.80}, 0, 1\right) \\
&\cdot\, \mathbf{1}[r_{\mathrm{len}} \le 0.80].
\end{aligned}
\end{equation}
All reported trained checkpoints use $w_q{=}1.0$, $w_{\mathrm{rep}}{=}1.5$, $w_{\mathrm{len}}{=}0.8$, $w_{\mathrm{blank}}{=}1.0$, and composite clip $c{=}2.0$. The online \emph{Story Quality} judge uses temperature $0.2$ and top-$p$ $0.9$.

\subsection{HRI Design Choices and Alternatives}
\label{app:hri_design}

The three constraints in \secref{sec:hri} ($\mu,\sigma$ computed from policy rollouts only; advantage scaled by $\alpha_t$; reference scored identically to policy outputs) are the only modifications required to use a human reference as a demonstration-like trajectory. Because the reference is teacher-forced under the current policy, current-policy log probabilities are available and the reference can be optimized through the same clipped surrogate used for policy rollouts. The reference is not sampled from the policy, however, so HRI is a biased demonstration-augmented GRPO update rather than an unbiased on-policy estimator; we omit importance correction by design and use the warmup $\alpha_t$ to dampen the demonstration-side gradient. No per-example gate decides whether to include it, no failure condition triggers it, and no auxiliary loss is attached.

To our knowledge, no prior off-policy injection method in either verifiable-reward or soft-reward settings combines these properties: a complete reference output injected as a demonstration trajectory, excluded from group statistics, and included in the policy update with a single warmup-scheduled scalar weight. LUFFY~\citep{luffy}, G2RPO-A~\citep{g2rpoa}, and BREAD~\citep{bread} target verifiable-reward tasks and inject partial expert prefixes rather than complete reference outputs; they also operate in settings where reward is binary, which makes variance collapse a less acute concern. We view the minimality as a feature: the training loop is a GRPO update with one additional teacher-forced demonstration and a scalar advantage multiplier, which keeps the contribution easy to reproduce, easy to ablate, and easy to combine with other GRPO extensions.

\section{Expanded Benchmark Results}
\label{app:extended_results}

This section expands the main-paper benchmark tables with the bucket-level and category-level results that are too detailed to fit in the main text.

\subsection{Full Length-Bucket Breakdown: Story Quality}

Table~\ref{app:sq_buckets} expands the main-paper \emph{Story Quality} results into the full six requested-length buckets and reports a simple slope summary across buckets.

\begin{table*}[t]
\centering
\scriptsize
\setlength{\tabcolsep}{3pt}
\begin{tabular}{@{}lr ccccccc c@{}}
\toprule
Model & P & 1--2k & 2--3k & 3--4k & 4--6k & 6--8k & 8--12k & Agg & Slope \\
\midrule
Human & -- & 66.4 & 64.9 & 73.0 & 70.0 & 66.5 & 71.4 & 68.7 & +0.8 \\
GPT-5.4 & -- & 74.6 & 76.3 & 79.8 & 78.0 & 66.3 & 59.5 & 72.4 & $-$3.1 \\
GPT-5.4-mini & -- & 69.5 & 71.3 & 70.8 & 62.0 & 60.5 & 59.0 & 65.5 & $-$2.6 \\
Claude Opus 4.6 & -- & 65.5 & 66.4 & 67.3 & 62.7 & 56.4 & 52.7 & 61.8 & $-$2.8 \\
Gemini 3.1 Pro & -- & 60.9 & 56.6 & 58.6 & 58.0 & 53.4 & 57.1 & 57.5 & $-$0.8 \\
GPT-5.4-nano & -- & 57.4 & 54.6 & 52.6 & 48.9 & 47.3 & 43.2 & 50.7 & $-$2.7 \\
Gemini 3 Flash & -- & 57.5 & 52.1 & 49.1 & 40.1 & 39.8 & 52.6 & 48.5 & $-$2.0 \\
Qwen3.5-27B & 27B & 54.6 & 52.1 & 47.8 & 42.9 & 34.5 & 24.6 & 42.8 & $-$5.9 \\
Qwen3.5-9B & 9B & 40.9 & 33.1 & 31.3 & 18.7 & $-$1.2 & $-$11.8 & 18.5 & $-$10.8 \\
\rowcolor{gray!10} \; + SFT & 9B & $-$40.3 & $-$61.2 & $-$57.4 & $-$64.3 & $-$69.4 & $-$71.2 & $-$60.6 & $-$5.3 \\
\rowcolor{gray!10} \; + GRPO & 9B & 58.2 & 53.8 & 54.4 & 48.1 & 45.8 & 37.7 & 49.7 & $-$3.8 \\
\rowcolor{gray!10} \; + \storylm{} & 9B & 57.7 & 55.8 & 58.6 & 50.7 & 45.7 & 44.1 & 52.1 & $-$3.0 \\
Gemma 4 31B & 31B & 54.0 & 53.6 & 54.1 & 49.7 & 49.7 & 47.1 & 51.4 & $-$1.4 \\
Gemma 4 26B-A4B & 26B & 53.2 & 52.1 & 53.3 & 52.2 & 50.6 & 46.5 & 51.3 & $-$1.1 \\
Gemma 4 E4B & 8B & 42.9 & 39.4 & 39.2 & 35.1 & 36.3 & 23.7 & 36.1 & $-$3.1 \\
DeepSeek-R1-14B & 14B & 12.8 & 1.9 & 7.3 & $-$1.6 & $-$7.7 & $-$11.1 & 0.2 & $-$4.5 \\
Ministral-3-8B & 8B & 35.1 & 28.8 & 26.6 & 23.5 & 20.6 & 21.8 & 26.1 & $-$2.7 \\
LongWriter-Zero-32B & 32B & 10.4 & $-$27.2 & $-$33.7 & $-$38.2 & $-$32.9 & $-$41.2 & $-$27.1 & $-$8.0 \\
LongWriter-Llama3.1-8B & 8B & 14.9 & 0.9 & 5.7 & $-$6.3 & $-$14.7 & $-$12.7 & $-$2.0 & $-$5.6 \\
\bottomrule
\end{tabular}
\caption{\emph{Story Quality} (GPT-5.4 judge) by length bucket, 180 held-out test prompts (30 per bucket). Slope = linear fit across 6 buckets (points per step). \colorbox{gray!10}{Shaded} = our models. Per-bucket numbers match the aggregates in \tabref{tab:id_results}.}
\label{app:sq_buckets}
\end{table*}

\subsection{Full Length-Bucket Breakdown: EQ-Bench Longform}

Table~\ref{app:lf_buckets} provides the corresponding six-bucket breakdown for EQ-Bench Longform, which helps separate in-distribution behavior from near- and far-transfer performance.

\begin{table*}[t]
\centering
\scriptsize
\setlength{\tabcolsep}{3pt}
\begin{tabular}{@{}lr ccccccc c@{}}
\toprule
Model & P & 1--2k & 2--3k & 3--4k & 4--6k & 6--8k & 8--12k & Agg & Slope \\
\midrule
Human & -- & 72.0 & 72.5 & 77.2 & 72.8 & 73.3 & 77.1 & 74.2 & +0.7 \\
GPT-5.4 & -- & 80.8 & 82.0 & 82.4 & 81.1 & 78.6 & 77.0 & 80.3 & $-$0.9 \\
GPT-5.4-mini & -- & 78.5 & 77.5 & 77.0 & 74.0 & 72.9 & 71.7 & 75.3 & $-$1.4 \\
Claude Opus 4.6 & -- & 74.7 & 74.8 & 74.9 & 72.9 & 72.3 & 70.4 & 73.4 & $-$0.9 \\
Gemini 3.1 Pro & -- & 70.5 & 68.2 & 69.0 & 65.7 & 64.7 & 64.2 & 67.0 & $-$1.3 \\
GPT-5.4-nano & -- & 69.3 & 68.2 & 66.6 & 64.7 & 63.4 & 60.9 & 65.5 & $-$1.7 \\
Gemini 3 Flash & -- & 69.0 & 65.8 & 66.5 & 59.3 & 58.6 & 59.9 & 63.2 & $-$2.1 \\
Qwen3.5-27B & 27B & 64.1 & 63.4 & 61.0 & 56.3 & 53.4 & 46.7 & 57.5 & $-$3.5 \\
Qwen3.5-9B & 9B & 54.5 & 48.4 & 47.8 & 40.2 & 34.2 & 30.3 & 42.6 & $-$4.9 \\
\rowcolor{gray!10} \; + SFT & 9B & 29.2 & 25.5 & 26.5 & 25.0 & 25.5 & 24.2 & 26.0 & $-$0.8 \\
\rowcolor{gray!10} \; + GRPO & 9B & 64.2 & 60.3 & 61.8 & 56.6 & 54.7 & 51.6 & 58.2 & $-$2.4 \\
\rowcolor{gray!10} \; + \storylm{} & 9B & 64.2 & 62.4 & 62.8 & 58.6 & 56.3 & 54.3 & 59.8 & $-$2.1 \\
Gemma 4 31B & 31B & 66.3 & 63.5 & 63.5 & 60.2 & 58.2 & 57.7 & 61.5 & $-$1.8 \\
Gemma 4 26B-A4B & 26B & 62.0 & 61.5 & 60.6 & 55.6 & 56.0 & 52.8 & 58.1 & $-$1.9 \\
Gemma 4 E4B & 8B & 55.5 & 53.4 & 52.4 & 48.9 & 47.4 & 43.9 & 50.3 & $-$2.3 \\
DeepSeek-R1-14B & 14B & 38.3 & 34.5 & 33.4 & 34.1 & 34.0 & 33.7 & 34.6 & $-$0.7 \\
Ministral-3-8B & 8B & 48.7 & 41.5 & 39.2 & 37.2 & 34.2 & 33.3 & 39.1 & $-$2.9 \\
LongWriter-Zero-32B & 32B & 36.5 & 26.3 & 25.1 & 22.8 & 24.7 & 21.4 & 26.1 & $-$2.4 \\
LongWriter-Llama3.1-8B & 8B & 38.0 & 34.4 & 33.7 & 32.2 & 30.1 & 29.4 & 33.0 & $-$1.6 \\
\bottomrule
\end{tabular}
\caption{EQ-Bench Longform (uniform per-dim aggregation, GPT-5.4 judge) by length bucket, 180 held-out test prompts (30 per bucket). \colorbox{gray!10}{Shaded} = our models. Per-bucket numbers match the aggregates in \tabref{tab:id_results}.}
\label{app:lf_buckets}
\end{table*}

\subsection{Length Ratio by Bucket}

Table~\ref{app:lr_buckets} reports generated/requested length ratios by bucket, complementing the quality tables with a direct view of under- and over-generation.

\begin{table*}[t]
\centering
\scriptsize
\setlength{\tabcolsep}{3pt}
\begin{tabular}{@{}lr ccccccc@{}}
\toprule
Model & P & 1--2k & 2--3k & 3--4k & 4--6k & 6--8k & 8--12k & All \\
\midrule
Human & -- & 1.00 & 1.00 & 1.00 & 1.00 & 1.00 & 1.00 & 1.00 \\
GPT-5.4 & -- & 1.52 & 1.36 & 1.29 & 1.26 & 1.20 & 0.94 & 1.26 \\
GPT-5.4-mini & -- & 1.60 & 1.52 & 1.42 & 1.36 & 1.17 & 0.91 & 1.33 \\
Claude Opus 4.6 & -- & 1.06 & 1.10 & 1.18 & 1.20 & 1.26 & 1.09 & 1.15 \\
Gemini 3.1 Pro & -- & 1.26 & 1.19 & 1.12 & 0.93 & 0.85 & 0.57 & 0.99 \\
GPT-5.4-nano & -- & 2.21 & 2.17 & 2.06 & 1.55 & 1.33 & 0.99 & 1.72 \\
Gemini 3 Flash & -- & 1.21 & 1.08 & 1.06 & 0.86 & 0.69 & 0.37 & 0.88 \\
Qwen3.5-27B & 27B & 1.02 & 1.02 & 1.04 & 1.01 & 0.94 & 0.82 & 0.97 \\
Qwen3.5-9B & 9B & 1.06 & 1.07 & 1.14 & 0.95 & 0.98 & 0.88 & 1.01 \\
\rowcolor{gray!10}  \; + \storylm{} & 9B & 1.09 & 0.98 & 0.91 & 0.91 & 0.83 & 0.72 & 0.90 \\
\rowcolor{gray!10} \; + GRPO & 9B & 0.97 & 0.95 & 0.90 & 0.89 & 0.82 & 0.70 & 0.87 \\
\rowcolor{gray!10}  \; + SFT & 9B & 3.98 & 3.30 & 2.42 & 1.69 & 1.24 & 0.91 & 2.26 \\
Gemma 4 31B & 31B & 1.01 & 0.85 & 0.74 & 0.64 & 0.47 & 0.36 & 0.68 \\
Gemma 4 26B-A4B & 26B & 1.10 & 0.90 & 0.74 & 0.51 & 0.38 & 0.26 & 0.65 \\
Gemma 4 E4B & 8B & 1.04 & 0.95 & 0.85 & 0.64 & 0.53 & 0.34 & 0.72 \\
DeepSeek-R1-14B & 14B & 1.73 & 0.87 & 0.84 & 0.64 & 0.19 & 0.23 & 0.75 \\
Ministral-3-8B & 8B & 0.93 & 0.70 & 0.56 & 0.38 & 0.30 & 0.23 & 0.52 \\
LongWriter-Zero-32B & 32B & 2.55 & 3.52 & 2.98 & 1.88 & 1.32 & 1.03 & 2.21 \\
LongWriter-Llama3.1-8B & 8B & 1.34 & 1.78 & 1.27 & 1.16 & 1.08 & 0.71 & 1.22 \\
\bottomrule
\end{tabular}
\caption{Length ratio (generated words / requested words) by bucket. 1.0 = perfect adherence; $<$1 = undershooting; $>$1 = overshooting. \colorbox{gray!10}{Shaded} = our models.}
\label{app:lr_buckets}
\end{table*}

\subsection{Coherence Across Requested-Length Buckets}
\label{app:coherence_length}

A length ratio close to 1.0 is only meaningful if the extra words are doing narrative work, so we
pair the length ratios of \tabref{app:lr_buckets} with the coherence dimensions that both of our
rubric evaluations already contain. \emph{Story Quality} has a negative dimension covering
coherence, continuity, internal consistency, and point-of-view confusion (lower is better); EQ-Bench
Longform has a \emph{Coherent} dimension on a 0--20 scale (higher is better). \tabref{tab:coherence_length}
reports both alongside the length ratio, grouped into the in-domain (1--4k), near-transfer (4--8k),
and far-transfer (8--12k) requested-length ranges.

Two failure patterns separate the open-weight baselines. The Qwen3.5 models track requested length
more closely than \storylmbig{} at near- and far-transfer lengths, but their coherence degrades much
faster as stories grow: from the 1--4k to the 8--12k range, Qwen3.5-9B's coherence penalty rises
72\% and its EQ-Bench coherence falls by 4.0 points, against 31\% and 1.8 points for \storylmbig{}.
The Gemma 4 models show the smallest coherence degradation of any model here, but they buy it by
writing far shorter stories than requested, down to a length ratio of 0.26 for Gemma 4 26B-A4B at
8--12k; short outputs are easier to keep locally consistent. Coherence and length adherence thus
trade off against each other across this model set, and \storylmbig{} sits between the two extremes:
it holds length ratios of 0.87 and 0.72 at near- and far-transfer lengths without the sharp
coherence loss of the models that pursue length more aggressively.

\begin{table*}[t]
\centering
\scriptsize
\setlength{\tabcolsep}{4pt}
\begin{tabular}{@{}l ccc c ccc c ccc@{}}
\toprule
& \multicolumn{3}{c}{1--4k (in-domain)} & & \multicolumn{3}{c}{4--8k (near transfer)} & & \multicolumn{3}{c}{8--12k (far transfer)} \\
\cmidrule(lr){2-4} \cmidrule(lr){6-8} \cmidrule(lr){10-12}
Model & LR & SQ-C $\downarrow$ & LF-C $\uparrow$ & & LR & SQ-C $\downarrow$ & LF-C $\uparrow$ & & LR & SQ-C $\downarrow$ & LF-C $\uparrow$ \\
\midrule
\rowcolor{gray!10} \storylmbig{} & 0.99 & 8.53 & 12.63 & & 0.87 & 10.36 & 11.49 & & 0.72 & 11.18 & 10.86 \\
Qwen3.5-9B & 1.09 & 12.98 & 10.05 & & 0.96 & 18.25 & 7.44 & & 0.88 & 22.36 & 6.06 \\
Qwen3.5-27B & 1.03 & 9.70 & 12.57 & & 0.97 & 12.26 & 10.97 & & 0.82 & 15.08 & 9.34 \\
Gemma 4 E4B & 0.95 & 11.90 & 10.75 & & 0.58 & 12.86 & 9.63 & & 0.34 & 15.26 & 8.78 \\
Gemma 4 26B-A4B & 0.91 & 9.43 & 12.27 & & 0.45 & 9.72 & 11.16 & & 0.26 & 10.70 & 10.56 \\
Gemma 4 31B & 0.87 & 9.22 & 12.89 & & 0.55 & 10.06 & 11.84 & & 0.36 & 10.58 & 11.54 \\
\bottomrule
\end{tabular}
\caption{Length adherence and coherence by requested-length range, on the 180 held-out test prompts.
LR is the length ratio (generated / requested word count; 1.00 is exact, below 1.00 undershoots).
SQ-C is the \emph{Story Quality} coherence penalty, covering coherence, continuity, internal
consistency, and point-of-view confusion; 0 means no coherence problems were detected, so lower is
better. LF-C is the EQ-Bench Longform \emph{Coherent} dimension on its raw 0--20 scale, where higher
is better. Ranges refer to the length requested in the prompt, not the length actually generated.
\colorbox{gray!10}{Shaded} = ours.}
\label{tab:coherence_length}
\end{table*}

\subsection{WritingBench Category Breakdown}
\label{app:wb_category_breakdown}

Tables~\ref{tab:wb_categories_a}--\ref{tab:wb_categories_c} unpack the single WritingBench column in the main paper into per-category means, showing where story-only training transfers cleanly and where it does not.

\begin{table*}[t]
\centering
\scriptsize
\setlength{\tabcolsep}{3pt}
\begin{tabular}{@{}lccccccc@{}}
\toprule
Model & Overall & \shortstack{Bio-\\graphy} & \shortstack{Book\\Review} & Brainstorm & \shortstack{Character\\Design} & \shortstack{Derivative\\Work} & \shortstack{Fan\\Fiction} \\
\midrule
GPT-5.4 & 9.46 & 9.29 & 9.11 & 8.87 & 9.00 & 9.70 & 10.00 \\
Claude Opus 4.6 & 9.32 & 9.27 & 9.71 & 9.00 & 9.85 & 10.00 & 9.95 \\
Gemini 3.1 Pro & 9.21 & 9.17 & 9.46 & 9.05 & 8.50 & 9.95 & 9.57 \\
Gemini 3 Flash & 9.08 & 8.36 & 9.27 & 9.03 & 8.30 & 9.95 & 9.62 \\
Qwen3.5-27B & 8.11 & 7.53 & 5.41 & 8.38 & 8.70 & 8.50 & 8.95 \\
Qwen3.5-9B & 6.84 & 4.50 & 6.19 & 6.12 & 9.15 & 7.10 & 6.85 \\
\rowcolor{gray!10}  \; + \storylm{} & 7.90 & 6.03 & 6.31 & 8.53 & 9.60 & 8.60 & 8.60 \\
\rowcolor{gray!10}  \; + GRPO & 7.85 & 6.43 & 5.66 & 8.53 & 9.40 & 8.20 & 8.95 \\
\rowcolor{gray!10}  \; + SFT & 2.68 & 1.46 & 1.80 & 2.80 & 5.50 & 1.00 & 1.30 \\
Gemma 4 31B & 8.66 & 7.54 & 8.46 & 8.43 & 8.90 & 10.00 & 9.65 \\
Gemma 4 26B-A4B & 8.47 & 7.16 & 8.36 & 7.37 & 9.00 & 9.05 & 9.50 \\
Gemma 4 E4B & 7.68 & 6.26 & 7.97 & 8.13 & 9.30 & 6.40 & 9.10 \\
DeepSeek-R1-14B & 3.94 & 3.06 & 3.11 & 3.83 & 3.20 & 3.90 & 3.95 \\
Ministral-3-8B & 4.80 & 3.11 & 2.66 & 4.70 & 8.20 & 4.90 & 6.30 \\
LongWriter-Zero-32B & 4.60 & 3.06 & 2.37 & 5.04 & 8.60 & 6.20 & 2.50 \\
LongWriter-Llama3.1-8B & 3.86 & 3.04 & 3.81 & 2.92 & 5.70 & 4.70 & 3.27 \\
\bottomrule
\end{tabular}
\caption{WritingBench category means, part I (Gemini 3.1 Pro judge, 96 English D4 prompts). Each value is the mean of the five per-prompt criterion scores, averaged within category.}
\label{tab:wb_categories_a}
\end{table*}

\begin{table*}[t]
\centering
\scriptsize
\setlength{\tabcolsep}{3pt}
\begin{tabular}{@{}lccccccc@{}}
\toprule
Model & \shortstack{Film/TV\\Review} & \shortstack{Game\\Design} & \shortstack{Greeting\\Message} & \shortstack{Host\\Script} & \shortstack{Novel\\Manuscript} & \shortstack{Novel\\Outline} & \shortstack{Plot\\Development} \\
\midrule
GPT-5.4 & 9.70 & 9.48 & 9.00 & 9.80 & 9.73 & 10.00 & 9.35 \\
Claude Opus 4.6 & 9.10 & 9.76 & 9.30 & 9.35 & 9.27 & 9.77 & 9.16 \\
Gemini 3.1 Pro & 9.30 & 8.88 & 9.60 & 10.00 & 9.23 & 8.53 & 9.01 \\
Gemini 3 Flash & 9.00 & 8.26 & 8.70 & 9.65 & 9.02 & 8.80 & 8.72 \\
Qwen3.5-27B & 6.92 & 6.94 & 9.20 & 9.05 & 8.77 & 10.00 & 8.21 \\
Qwen3.5-9B & 5.88 & 6.54 & 9.30 & 9.00 & 7.43 & 9.53 & 6.36 \\
\rowcolor{gray!10}  \; + SFT & 3.15 & 3.80 & 7.40 & 5.10 & 1.67 & 2.47 & 1.10 \\
\rowcolor{gray!10}  \; + GRPO & 7.15 & 7.16 & 8.00 & 9.80 & 8.17 & 9.40 & 8.43 \\
\rowcolor{gray!10}  \; + \storylm{} & 6.70 & 7.32 & 9.40 & 9.10 & 8.27 & 9.60 & 8.57 \\
Gemma 4 31B & 6.60 & 8.40 & 9.40 & 9.80 & 9.13 & 9.07 & 8.22 \\
Gemma 4 26B-A4B & 8.35 & 8.44 & 9.60 & 9.55 & 8.85 & 9.47 & 8.91 \\
Gemma 4 E4B & 6.50 & 8.24 & 7.20 & 8.30 & 8.03 & 9.33 & 7.55 \\
DeepSeek-R1-14B & 2.95 & 3.12 & 6.60 & 5.50 & 4.03 & 4.87 & 3.62 \\
Ministral-3-8B & 3.40 & 3.16 & 8.60 & 9.10 & 5.03 & 7.47 & 4.42 \\
LongWriter-Zero-32B & 3.00 & 3.10 & 7.00 & 8.80 & 5.04 & 5.60 & 4.11 \\
LongWriter-Llama3.1-8B & 3.62 & 4.00 & 4.30 & 5.80 & 5.13 & 3.57 & 3.90 \\
\bottomrule
\end{tabular}
\caption{WritingBench category means, part II.}
\label{tab:wb_categories_b}
\end{table*}

\begin{table*}[t]
\centering
\scriptsize
\setlength{\tabcolsep}{3pt}
\begin{tabular}{@{}lcccccc@{}}
\toprule
Model & \shortstack{Podcast\\Script} & Poetry & Prose & \shortstack{Reading\\Reflection} & Screenplay & \shortstack{Video\\Script} \\
\midrule
GPT-5.4 & 9.51 & 9.10 & 9.67 & 9.30 & 9.86 & 9.44 \\
Claude Opus 4.6 & 9.80 & 7.87 & 9.89 & 9.75 & 8.41 & 9.37 \\
Gemini 3.1 Pro & 9.59 & 8.03 & 9.58 & 9.50 & 9.47 & 9.23 \\
Gemini 3 Flash & 9.53 & 9.20 & 9.73 & 9.60 & 8.77 & 9.29 \\
Qwen3.5-27B & 8.86 & 7.30 & 9.03 & 6.80 & 8.59 & 9.16 \\
Qwen3.5-9B & 6.59 & 7.50 & 8.15 & 5.85 & 6.53 & 7.66 \\
\rowcolor{gray!10}  \; + SFT & 4.74 & 3.20 & 2.77 & 2.60 & 1.46 & 3.84 \\
\rowcolor{gray!10}  \; + GRPO & 7.66 & 7.60 & 9.37 & 4.90 & 8.57 & 8.31 \\
\rowcolor{gray!10}  \; + \storylm{} & 8.06 & 6.97 & 8.03 & 7.10 & 8.34 & 8.73 \\
Gemma 4 31B & 9.54 & 7.40 & 9.30 & 8.70 & 8.66 & 9.40 \\
Gemma 4 26B-A4B & 8.21 & 7.28 & 9.18 & 9.05 & 8.10 & 8.83 \\
Gemma 4 E4B & 7.80 & 6.10 & 8.40 & 6.90 & 7.17 & 8.24 \\
DeepSeek-R1-14B & 5.40 & 3.73 & 5.27 & 3.95 & 3.63 & 3.98 \\
Ministral-3-8B & 5.17 & 4.20 & 5.03 & 5.25 & 4.89 & 5.58 \\
LongWriter-Zero-32B & 5.37 & 5.10 & 6.84 & 4.45 & 4.60 & 4.95 \\
LongWriter-Llama3.1-8B & 3.66 & 3.67 & 4.58 & 5.03 & 3.03 & 3.54 \\
\bottomrule
\end{tabular}
\caption{WritingBench category means, part III. Together, Tables~\ref{tab:wb_categories_a}--\ref{tab:wb_categories_c} show the full per-category breakdown underlying the single WritingBench column in \tabref{tab:ood_results}.}
\label{tab:wb_categories_c}
\end{table*}

\raggedbottom
\section{Training and Evaluation Diagnostics}

The next set of analyses explains how the trained models improve over training, how stable the judges are, and how much uncertainty remains in the reported bucketed comparisons.

\subsection{Per-Dimension Pattern}
\label{sec:dim_analysis}

\begin{figure*}[t]
\centering
\begin{subfigure}[t]{0.48\textwidth}
  \includegraphics[width=\linewidth]{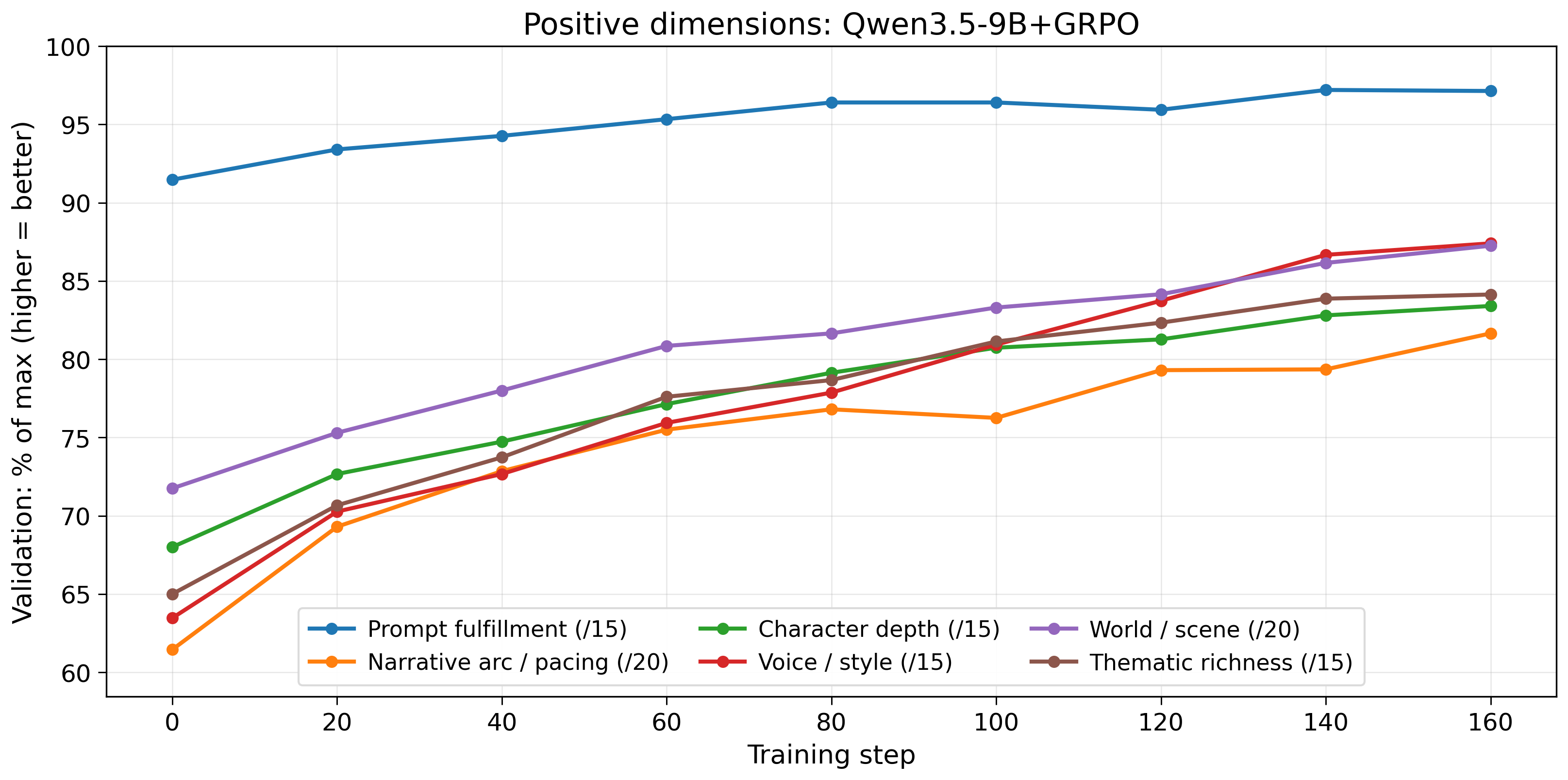}
  \caption{Qwen3.5-9B+GRPO, positive dimensions}
\end{subfigure}\hfill
\begin{subfigure}[t]{0.48\textwidth}
  \includegraphics[width=\linewidth]{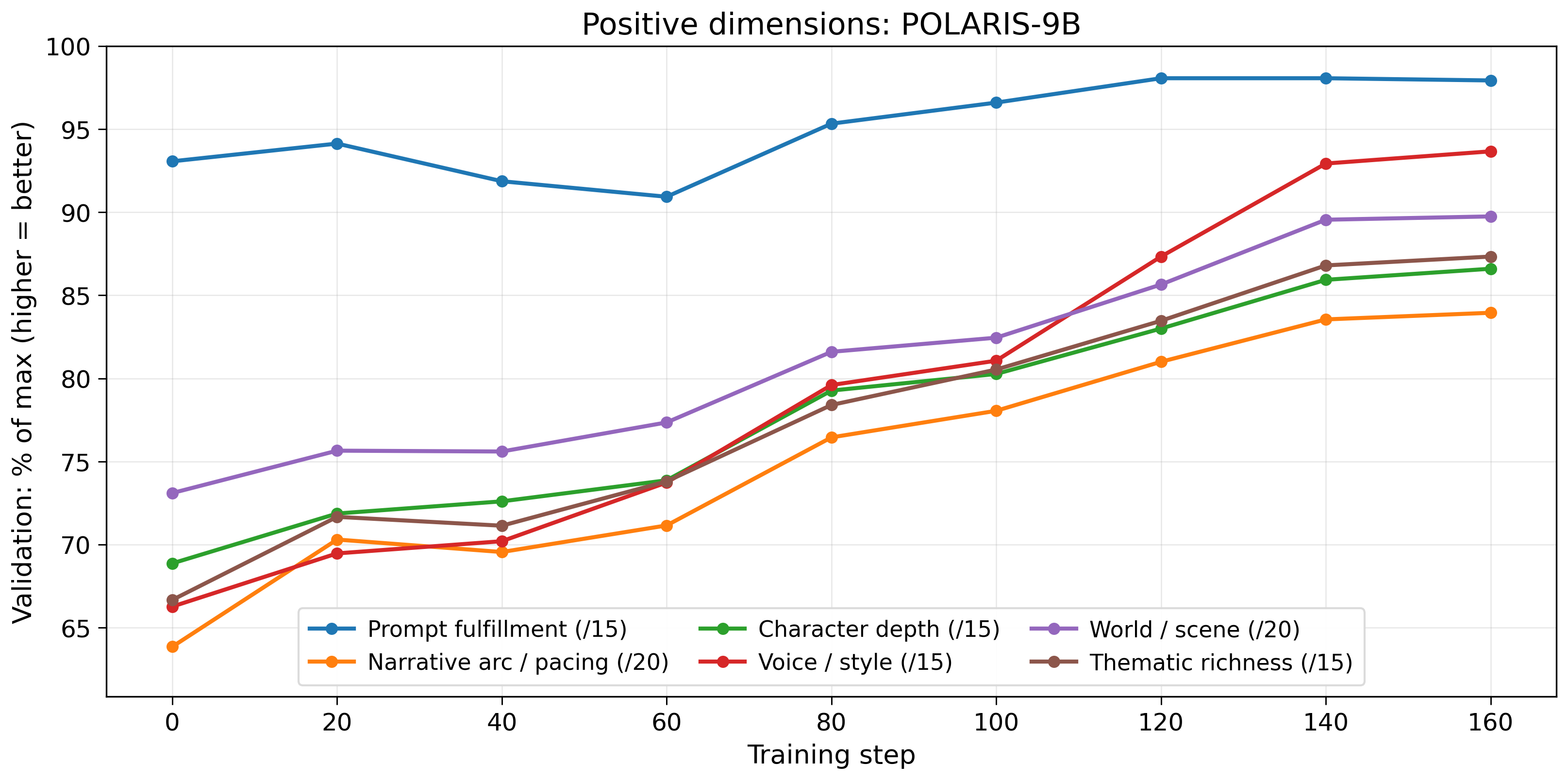}
  \caption{\storylmbig{}, positive dimensions}
\end{subfigure}

\vspace{6pt}

\begin{subfigure}[t]{0.48\textwidth}
  \includegraphics[width=\linewidth]{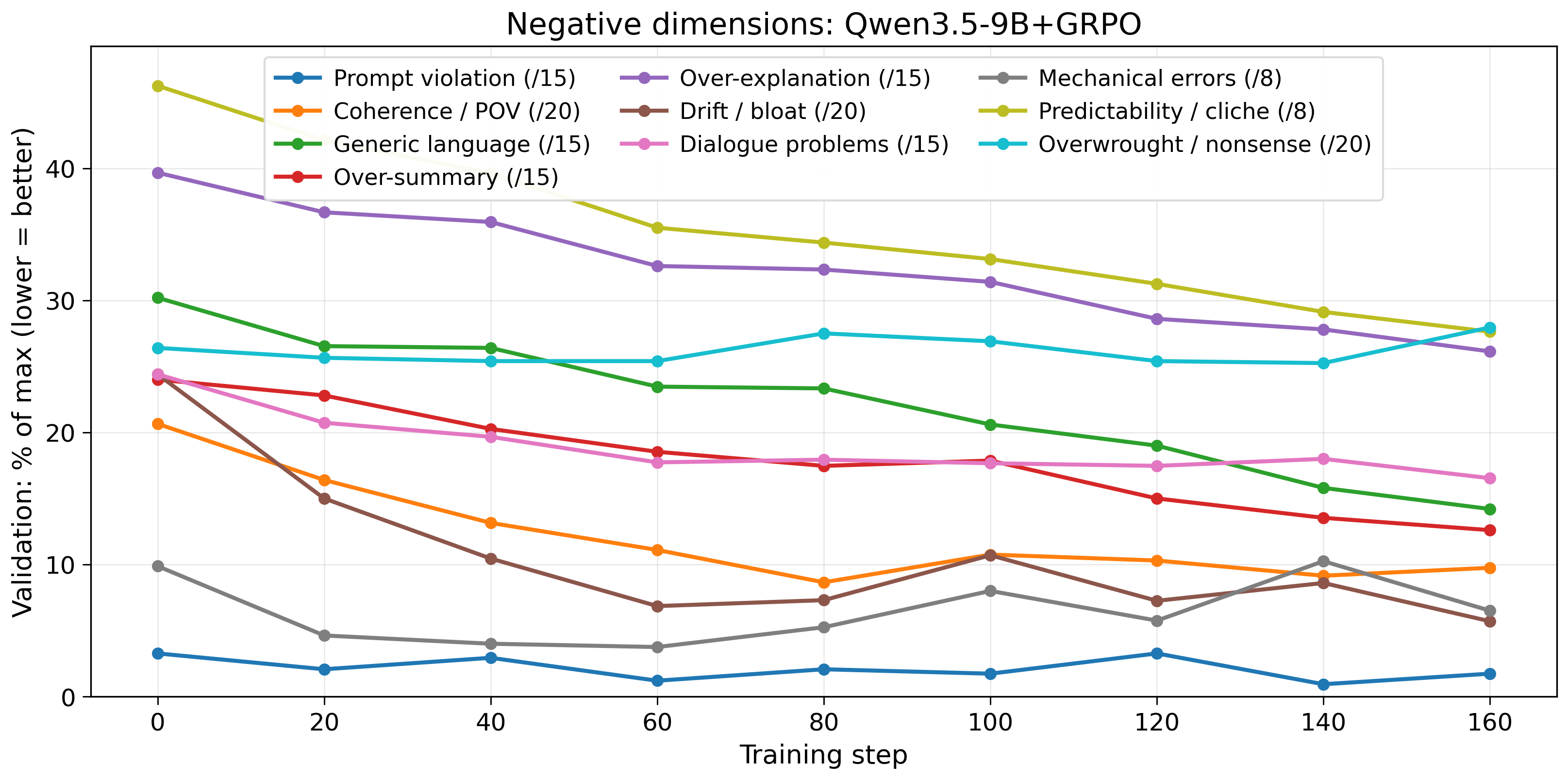}
  \caption{Qwen3.5-9B+GRPO, negative dimensions}
\end{subfigure}\hfill
\begin{subfigure}[t]{0.48\textwidth}
  \includegraphics[width=\linewidth]{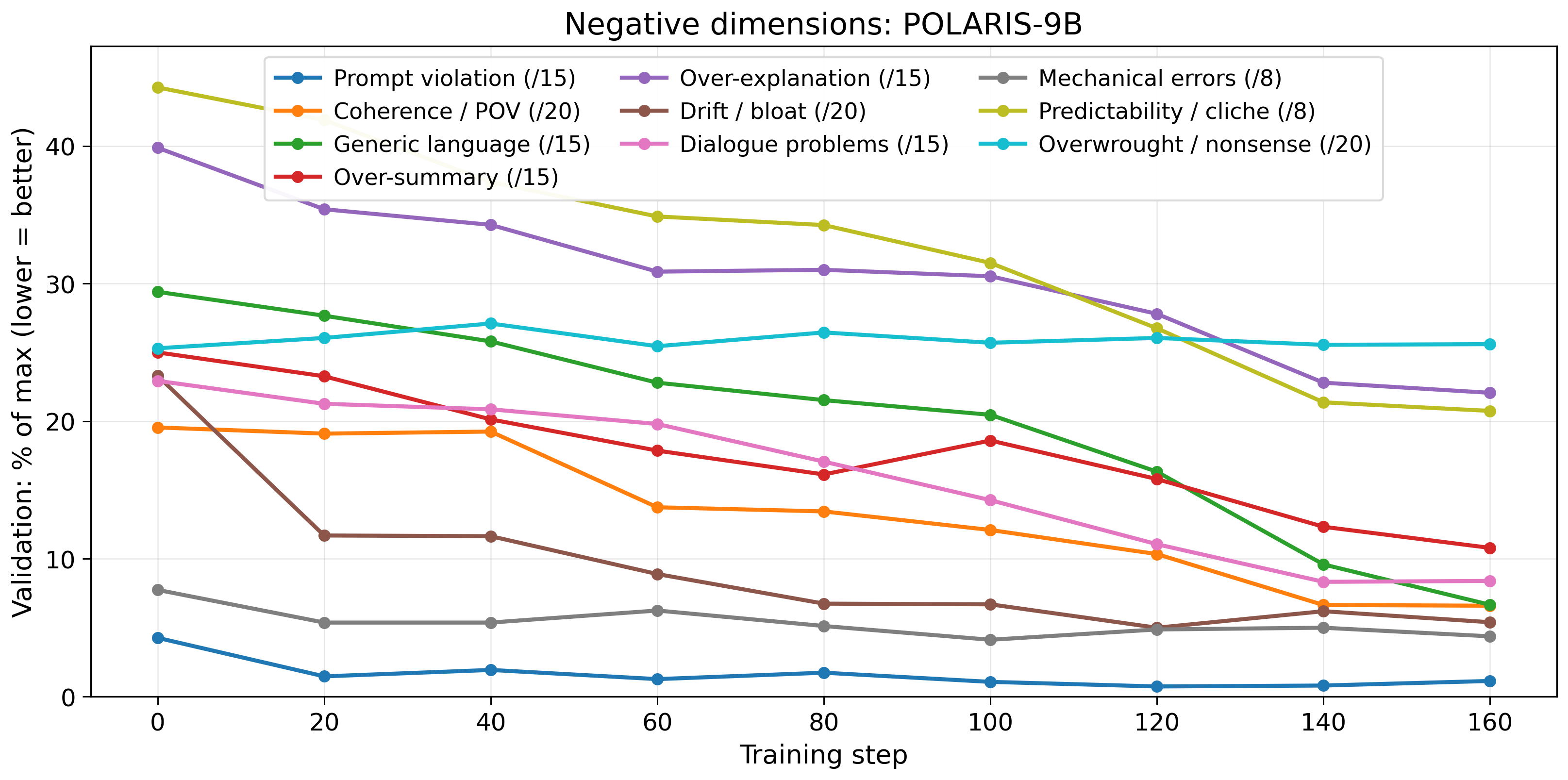}
  \caption{\storylmbig{}, negative dimensions}
\end{subfigure}
\caption{Per-dimension \emph{Story Quality} eval scores (Flash judge) over training steps, plotted as a percent of each tracked dimension's maximum under the full core rubric (6 positive, 10 negative). Top row: positive dimensions. Bottom row: negative dimensions, plotted as penalty magnitude so downward is better.}
\label{fig:perdim_trajectories}
\end{figure*}

The per-dimension training curves in \figref{fig:perdim_trajectories} track the full \emph{Story Quality} core rubric: 16 dimensions (6 positive, 10 negative) scored by the training judge on held-out validation rollouts. We use this section to inspect which dimensions improve over training and how the plain-GRPO and HRI runs differ.

At a high level, the same pattern seen in the final GPT-5.4 evaluation appears throughout training: the clearest positive gains are in voice, character depth, narrative arc, and world/scene realization, while the largest negative-side reductions are in generic language, drift/bloat, coherence/POV, over-explanation, and over-summary. Prompt fulfillment and mechanical errors move the least, suggesting that most of the gain comes from higher-level narrative quality rather than surface cleanup.

\paragraph{Training-time dynamics.}
\figref{fig:perdim_trajectories} plots per-dimension eval scores as a percent of each dimension's maximum over training steps for both Qwen3.5-9B+GRPO and \storylmbig{}. Most negative-side reduction happens early, while positive dimensions continue improving later into training. Voice and several drafting-failure dimensions account for much of the separation between the runs, and across most tracked dimensions \storylmbig{} finishes above Qwen3.5-9B+GRPO, consistent with the \emph{Story Quality} gap in \tabref{tab:id_results}. Overwrought prose is a weaker exception, which may indicate that it is tied more closely to base-model stylistic priors than to the higher-level drafting failures that lightweight RL corrects more readily.

\paragraph{HRI's per-dimension signature.}
Comparing Qwen3.5-9B+GRPO and \storylmbig{} at their final checkpoints, HRI is associated with slightly larger reductions in generic language, predictability, and dialogue. These per-dimension effects are small, but directionally consistent with HRI helping reduce familiar LLM-writing artifacts in addition to its larger role in far-transfer length robustness.

\begin{figure}[t]
\centering
\includegraphics[width=\columnwidth]{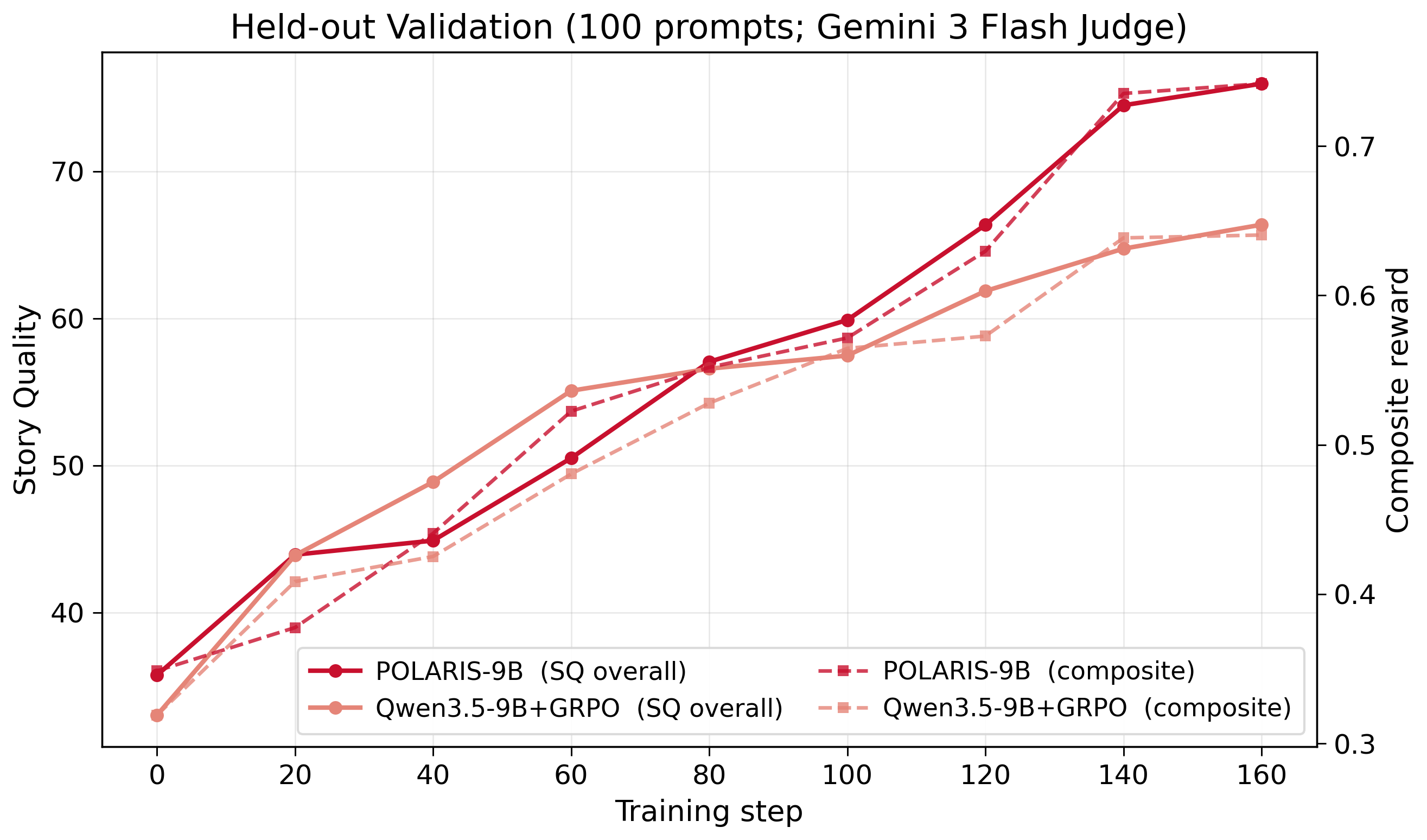}
\caption{Held-out validation \emph{Story Quality} (left axis, solid) and composite reward (right axis, dashed) over training steps for Qwen3.5-9B+GRPO and \storylmbig{} on a 100-prompt validation split (in-distribution length, 1--4k; Gemini 3 Flash judge). Both curves continue to improve through one epoch, with \storylmbig{} consistently ahead of the plain GRPO run without HRI.}
\label{fig:val_training_curves}
\end{figure}

\paragraph{Validation training curves.}
\figref{fig:val_training_curves} shows held-out \emph{Story Quality} and composite reward over training steps for Qwen3.5-9B+GRPO and \storylmbig{} on a 100-prompt validation split (in-distribution length, 1--4k words; Gemini 3 Flash judge, the training-reward judge). Both runs improve steadily over the course of one epoch, with \storylmbig{} remaining clearly ahead of Qwen3.5-9B+GRPO on both metrics throughout. The ordering matches the headline GPT-5.4 evaluation in \tabref{tab:id_results}, and both curves are still rising at the end of training, so the reported checkpoint is a cost-based stopping point rather than an obvious saturation point.

\begin{figure*}[tp]
\centering
\includegraphics[width=\textwidth]{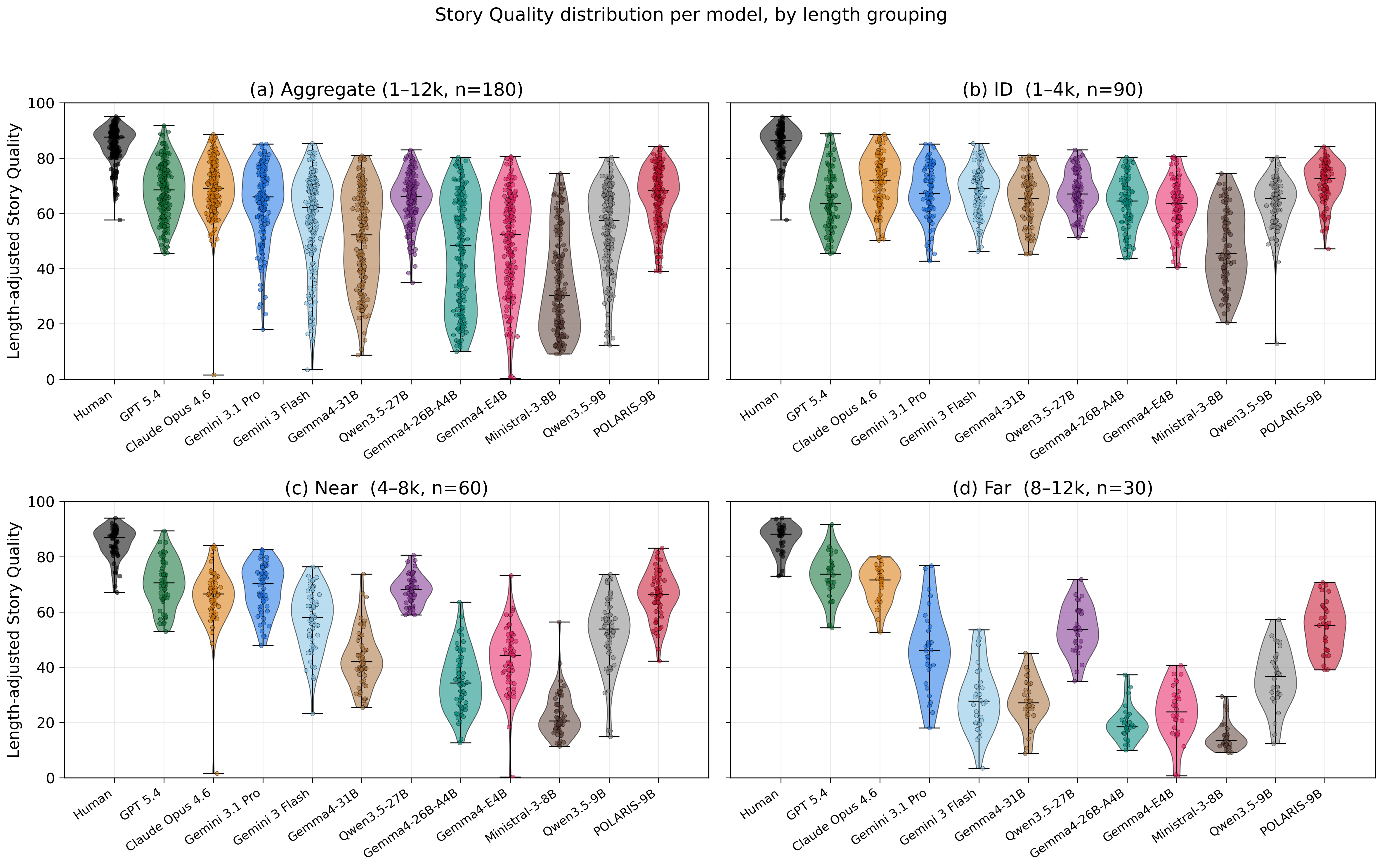}
\caption{Distribution of length-adjusted \emph{Story Quality} scores per model, by length-bucket grouping. For signed \emph{Story Quality}, we first normalize the raw rubric score from its floor and ceiling to $[0,100]$, then multiply by $\min(\mathrm{LR}, 1/\mathrm{LR})$.}
\label{fig:violin_buckets}
\end{figure*}

\paragraph{Per-model length-bucket distributions.}
\figref{fig:violin_buckets} shows the distribution of length-adjusted \emph{Story Quality} scores per model, broken out by length-bucket grouping (aggregate, ID, near-OOD, far-OOD). At far-OOD (8--12k), \storylmbig{}'s distribution is concentrated at the open-weight ceiling, while frontier models retain a narrow high-scoring distribution.

\subsection{Judge Stochasticity Measurement}
\label{app:judge_noise}

To quantify judge stochasticity independently of prompt-sampling noise, we score 20 stories three times each with GPT-5.4 on \emph{Story Quality} and EQ-Bench Longform, and 9 stories three times on EQ-Bench Creative (smaller prompt pool of 32). Stories are stratified across four score bins (SQ/LF) or three bins (EQ-Cr) spanning the full observed quality range (from LongWriter-Zero-32B outputs with SQ $<$ 0 to GPT-5.4 and Claude Opus 4.6 outputs with SQ $>$ 70). Each call uses medium-level thinking; GPT-5.4's reasoning API does not expose \texttt{temperature} or \texttt{seed}, so the three runs rely on the model's intrinsic stochasticity for independent samples under default sampling.

\begin{table*}[t]
\centering
\small
\resizebox{\textwidth}{!}{%
\begin{tabular}{@{}l c c c c c@{}}
\toprule
Benchmark & $n$ stories & ICC(3,1) & Mean Kendall $\tau$ & Within-story SD & \% variance from judge \\
\midrule
\emph{Story Quality}     & 20 & 0.973 & 0.864 & 8.3 & 2.6\% \\
EQ-Bench Longform & 20 & 0.987 & 0.909 & 2.1 & 1.2\% \\
EQ-Bench Creative &  9 & 0.975 & 0.889 & 2.2 & 2.2\% \\
\bottomrule
\end{tabular}%
}
\caption{Judge stochasticity on GPT-5.4, three independent runs per story. ICC(3,1) is the two-way mixed-effects single-rater intraclass correlation; values $>$0.9 indicate excellent reliability. Within-story SD is the standard deviation of the three scores per story, averaged across stories. ``\% variance from judge'' is the fraction of total score variance attributable to within-story (across-run) variation rather than between-story differences.}
\label{tab:judge_noise}
\end{table*}

All three ICC values exceed 0.97 (Table~\ref{tab:judge_noise}). Judge stochasticity accounts for 1.2--2.6\% of observed score variance; the remaining $>$97\% is real between-story quality differences. 
\emph{Story Quality} has the largest within-story SD (8.3 points) because its score range spans $[-180, +105]$ with correspondingly larger absolute magnitudes; the percentage variance contribution is comparable to the bounded-range benchmarks. This experiment was run with an earlier revision of the \emph{Story Quality} rubric (nine negative dimensions with maxima summing to 180, a 0--5 bonus, and no penalty term), which gives the $[-180, +105]$ range; the final rubric in \S\ref{app:storyquality_prompt} has ten negative dimensions summing to 151 and 0--8 bonus and penalty terms, so it spans $[-159, +108]$. The within-story SD is therefore on the earlier scale; ICC, Kendall's $\tau$, and the variance fraction are scale-free.

\section{Pairwise and Profile Analyses}

These analyses complement the scalar benchmark tables with pairwise comparisons and profile-based views over the full evaluated model set.

\subsection{Pairwise Elo Rankings (Full Model Set)}

\tabref{tab:ood_results} (1b, 1c) shows pairwise Elo for the 14--15 main-paper models. \tabref{app:elo_standard} extends to the full 16--17-model set, including GPT-5.4-mini/nano.

\begin{table*}[tp]
\centering
\scriptsize
\begin{minipage}[t]{0.48\textwidth}
\centering
\begin{tabular}{@{}rlr@{}}
\toprule
\# & Model & Elo \\
\midrule
1 & GPT-5.4 & 1911 \\
2 & Claude Opus 4.6 & 1774 \\
3 & GPT-5.4-mini & 1728 \\
4 & \storylmbig{} & 1658 \\
5 & Gemini 3.1 Pro & 1619 \\
6 & Gemini 3 Flash & 1612 \\
7 & GPT-5.4-nano & 1563 \\
8 & Gemma 4 31B & 1505 \\
9 & Qwen3.5-27B & 1494 \\
10 & Gemma 4 26B-A4B & 1451 \\
11 & Qwen3.5-9B & 1343 \\
12 & Gemma 4 E4B & 1301 \\
13 & Ministral-3-8B & 1214 \\
14 & DeepSeek-R1-Distill-Qwen-14B & 1033 \\
15 & LongWriter-Zero-32B & 1011 \\
16 & LongWriter-Llama3.1-8B & 800 \\
\bottomrule
\end{tabular}
\par\vspace{0.3em}\scriptsize EQ-Bench Creative (32 prompts, 9 dims)
\end{minipage}%
\hfill
\begin{minipage}[t]{0.48\textwidth}
\centering
\begin{tabular}{@{}rlr@{}}
\toprule
\# & Model & Elo \\
\midrule
1 & GPT-5.4 & 1578 \\
2 & Claude Opus 4.6 & 1464 \\
3 & GPT-5.4-mini & 1463 \\
4 & Human & 1460 \\
5 & \storylmbig{} & 1416 \\
6 & Gemini 3.1 Pro & 1380 \\
7 & Gemini 3 Flash & 1326 \\
8 & Gemma 4 31B & 1275 \\
9 & GPT-5.4-nano & 1275 \\
10 & Qwen3.5-27B & 1275 \\
11 & Gemma 4 26B-A4B & 1228 \\
12 & Gemma 4 E4B & 1100 \\
13 & Qwen3.5-9B & 1078 \\
14 & Ministral-3-8B & 1049 \\
15 & DeepSeek-R1-Distill-Qwen-14B & 894 \\
16 & LongWriter-Llama3.1-8B & 871 \\
17 & LongWriter-Zero-32B & 800 \\
\bottomrule
\end{tabular}
\par\vspace{0.3em}\scriptsize ID prompts (45 prompts, 11 judge dims + length, 147K dual-position comparisons)
\end{minipage}
\caption{Pairwise Elo (Bradley--Terry MLE on the full pairwise contingency; Flash judge, dual-position; anchored so the lowest model lands at 800), full model set including GPT-5.4-mini/nano.}
\label{app:elo_standard}
\end{table*}

\subsection{Pairwise Win-Rate Heatmaps}
\begin{figure*}[t]
\centering
\begin{subfigure}[t]{0.48\textwidth}
  \includegraphics[width=\textwidth]{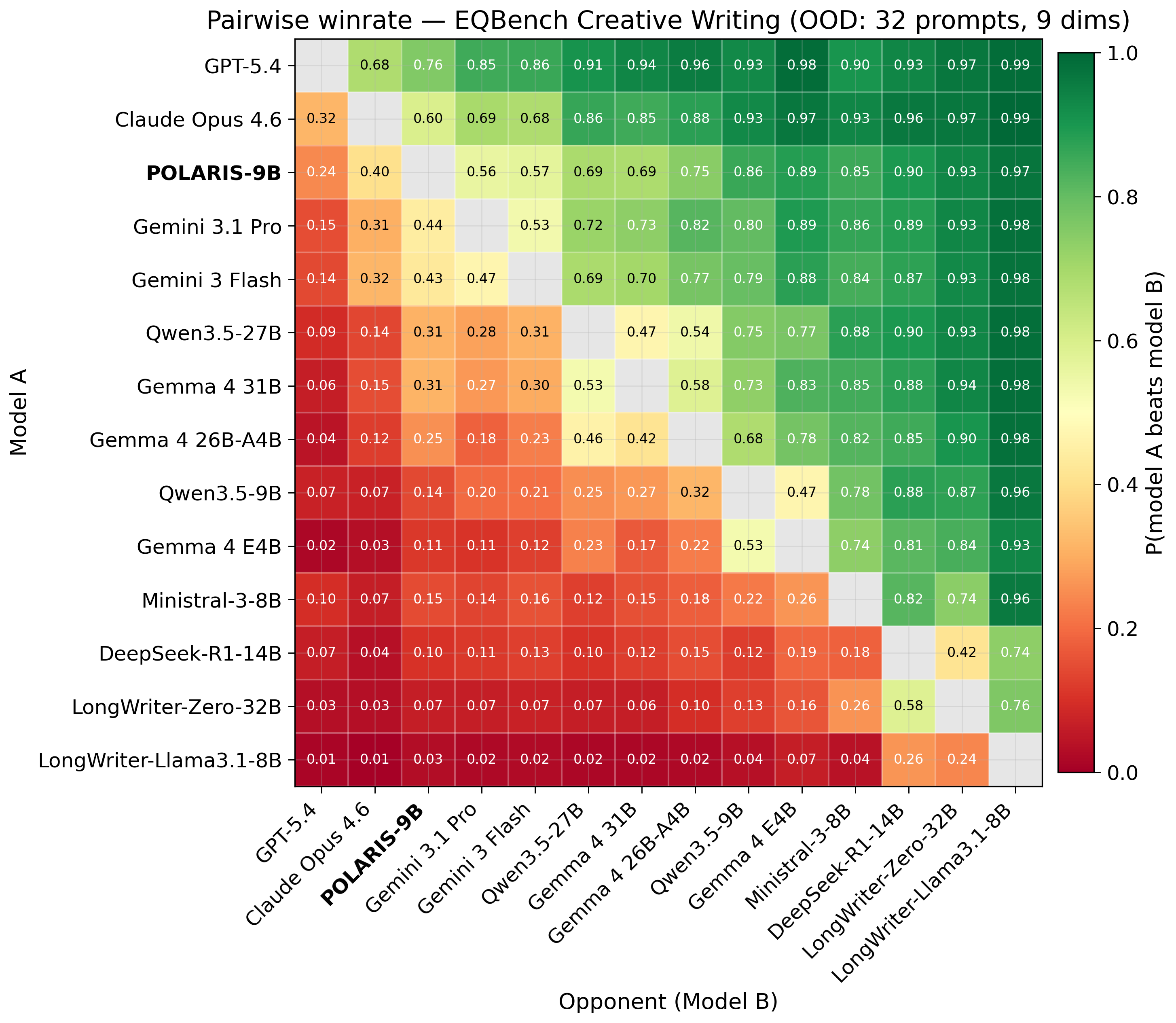}
  \caption{EQ-Bench Creative (OOD prompts and rubric).}
\end{subfigure}\hfill
\begin{subfigure}[t]{0.48\textwidth}
  \includegraphics[width=\textwidth]{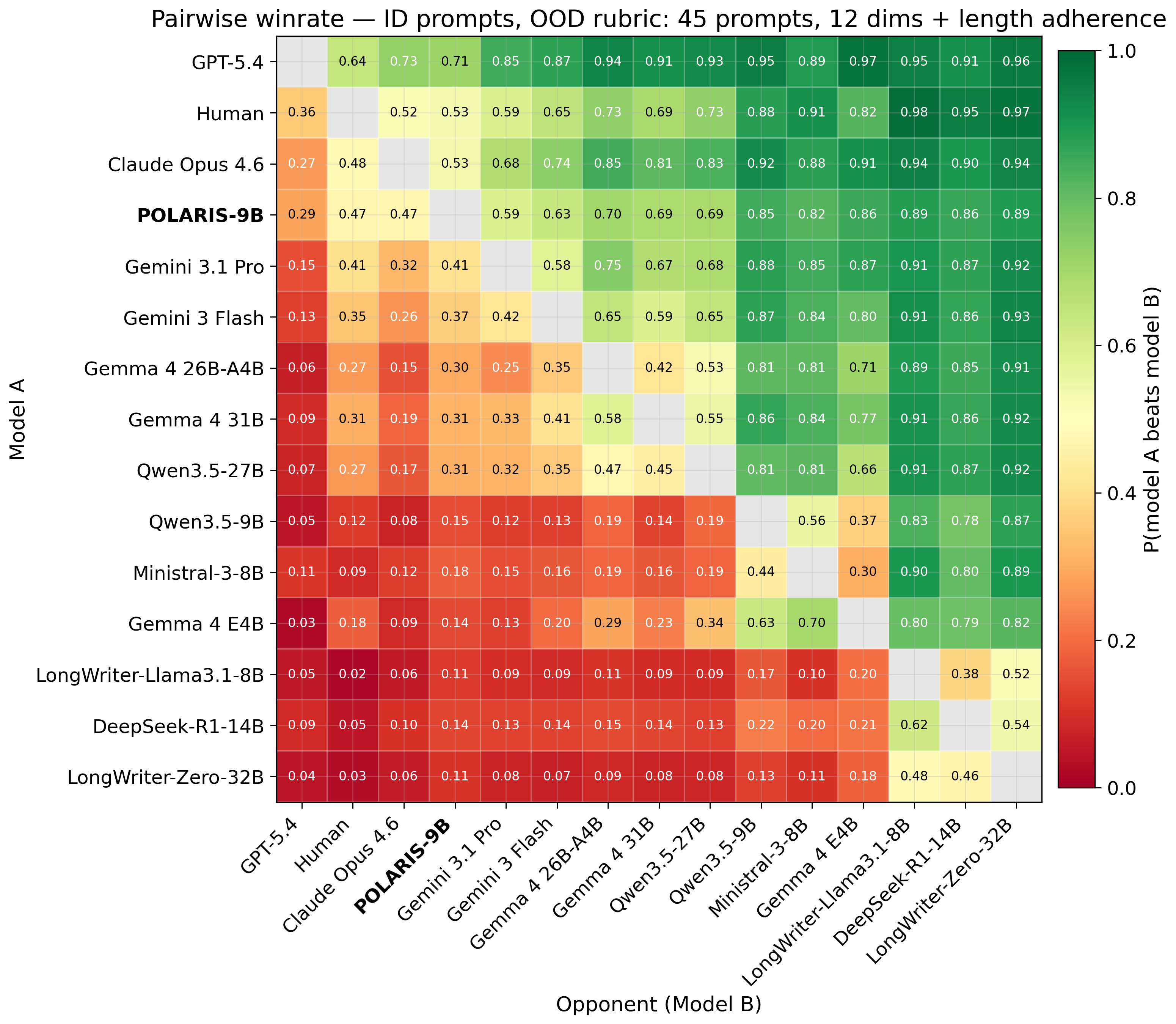}
  \caption{In-domain prompts, OOD dimensions + length adherence.}
\end{subfigure}
\caption{Pairwise winrate ($P$(row beats column)) over the per-dimension judge calls (Gemini 3 Flash, dual-position). Diagonal cells are blank; colors run from red (row loses) through yellow (50/50) to green (row wins). The Elo ratings in \tabref{tab:ood_results} are the Bradley--Terry MLE on these winrate matrices.}
\label{fig:pairwise_winrate}
\end{figure*}

\figref{fig:pairwise_winrate} shows the underlying pairwise winrates (model A row beats model B column) that the Elo ratings summarize. \storylmbig{} sits between the frontier (GPT-5.4, Claude Opus 4.6, GPT-5.4-mini) and the larger open-weight baselines (Qwen3.5-27B and the Gemma 4 family): it wins the majority of head-to-head matchups against every open-weight baseline on both rankings, and remains near 50\% against Gemini 3 Flash and Gemini 3.1 Pro on EQ-Bench Creative.

\subsection{Model Profile Clusters}
\begin{figure*}[p]
\centering
\includegraphics[width=0.95\textwidth]{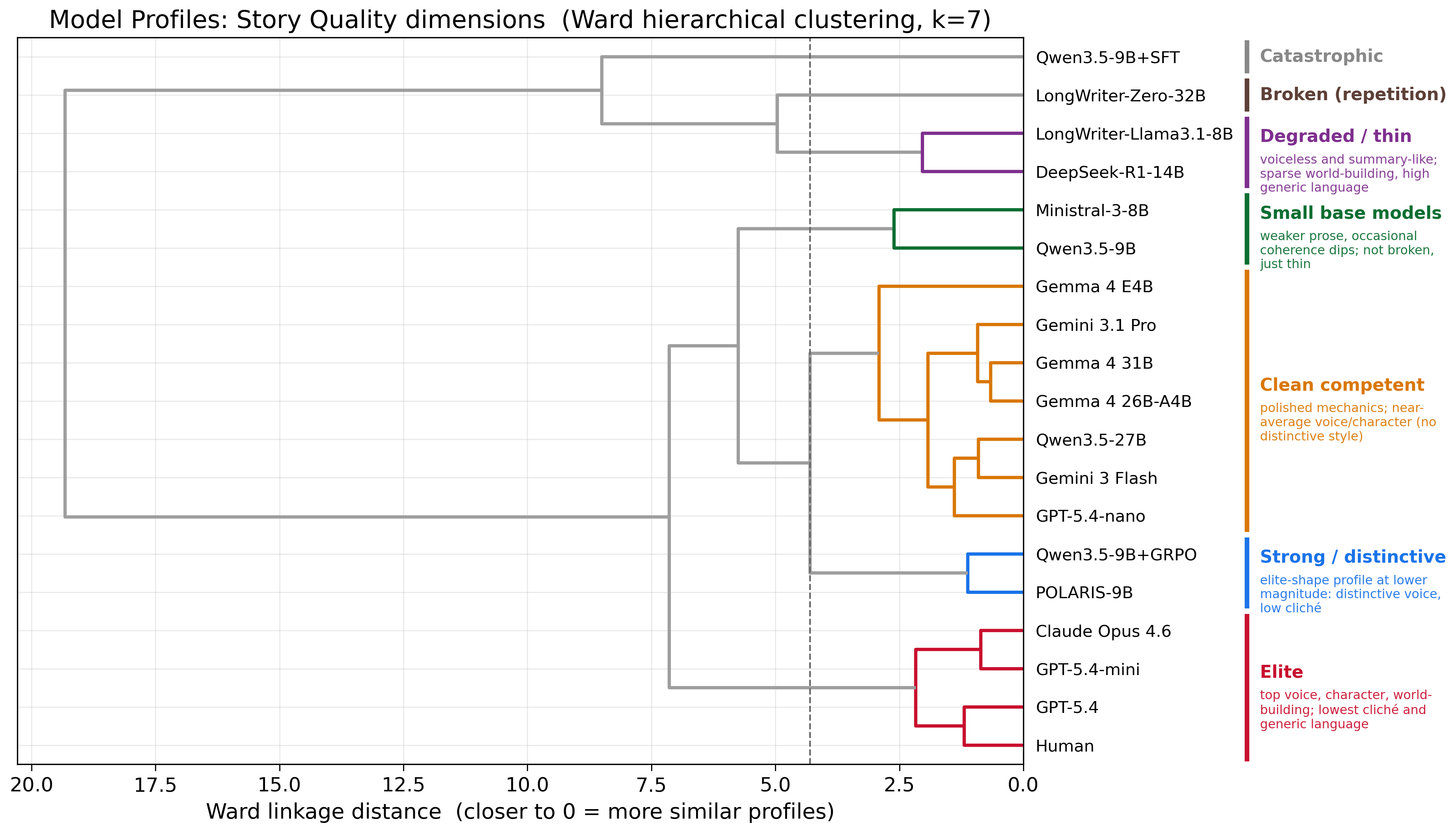}
\caption{Hierarchical clustering ($k{=}7$ Ward linkage) of the 18 evaluated models by their per-dimension \emph{Story Quality} profile. Cluster labels are hand-annotated post-hoc to describe the prose patterns characteristic of each tier; the clustering itself is derived only from the per-dim mean vectors.}
\label{fig:model_clusters}
\end{figure*}

\figref{fig:model_clusters} clusters all 18 evaluated models by their per-dimension \emph{Story Quality} profile (Ward hierarchical clustering on the 16-dim core-rubric mean vector, $k{=}7$ clusters). \storylmbig{} and Qwen3.5-9B+GRPO form their own ``Strong / distinctive'' cluster between the ``Clean competent'' mid-tier open-weight baselines (Gemma 4 family, Qwen3.5-27B, Gemini 3 Flash, GPT-5.4-nano) and the ``Elite'' frontier cluster (Claude Opus 4.6, GPT-5.4, GPT-5.4-mini, and the human references). The placement is qualitative (Ward distances are not statistically thresholded), but the figure illustrates that the trained 9B models do not look like a scaled-up Qwen3.5-9B base; they sit closer to the elite cluster on dimension shape, just at lower magnitude.

\subsection{Exact Per-Dimension Profile Across Evaluated Models}
\begin{figure*}[p]
\centering
\includegraphics[width=0.96\textwidth]{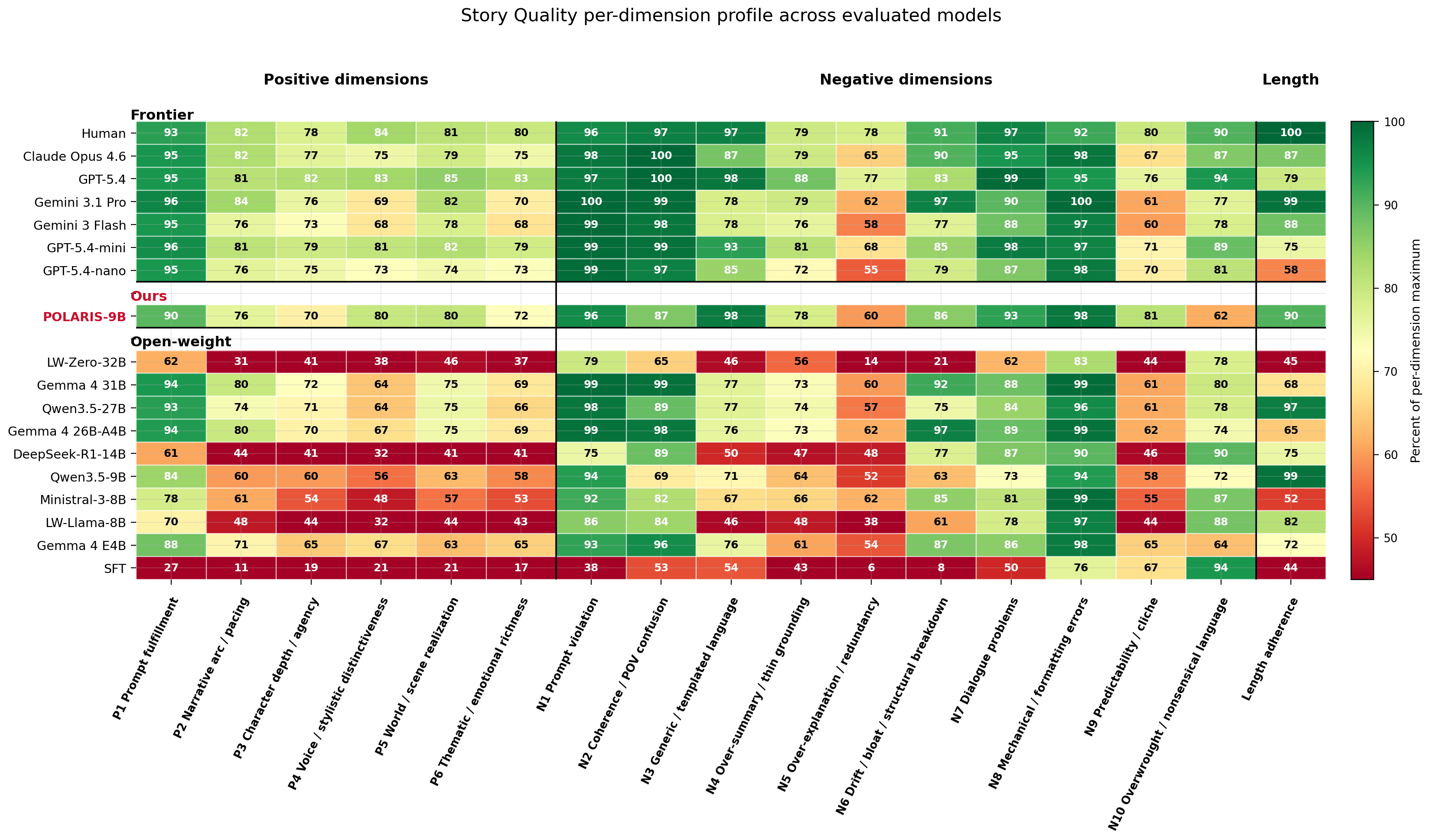}
\caption{\emph{Story Quality} per-dimension profile across the evaluated model set (GPT-5.4 judge, 180-prompt evaluation set), with an additional overall length-adherence column. Each cell shows a normalized percentage using that dimension's own rubric maximum $m_d$ (\S\ref{app:storyquality_prompt}; positives 15 or 20, negatives 8, 15, or 20). For positive dimensions we report $100\cdot s/m_d$; for negative dimensions we report $100\cdot(1 - p/m_d)$ so higher remains better; for length adherence we report $100\cdot\min(\mathrm{LR}, 1/\mathrm{LR})$, where $\mathrm{LR}$ is the aggregate generated/requested length ratio from the main evaluation table. This figure uses the same \emph{Story Quality} rubric family as the training-curve analysis in \secref{sec:dim_analysis}, but reports GPT-5.4 scores on the 180-prompt evaluation set rather than Gemini 3 Flash scores on held-out validation rollouts.}
\label{fig:sq_v12_perdim_pct_heatmap}
\end{figure*}

To complement the qualitative cluster view, \figref{fig:sq_v12_perdim_pct_heatmap} shows exact normalized \emph{Story Quality} dimension means for the full evaluated model set, plus an overall length-adherence column. Cell values are expressed as a percent of each dimension's maximum; for negative dimensions we invert the penalty scale so higher percentages still mean better outcomes, and the final length column reports $100\cdot\min(\mathrm{LR}, 1/\mathrm{LR})$ from the aggregate length-ratio statistic. This makes the main pattern easy to see: compared with the larger open-weight baselines, \storylmbig{} is especially strong on voice, scene realization, and generic-language control, while Gemma 4 31B retains a slight edge on prompt fulfillment and arc-related dimensions and a noticeably weaker length-adherence score.

\clearpage

\section{Judge Choice and Bias Analyses}
\label{app:pro_bias}

We ran a controlled judge comparison experiment across Gemini 3.1 Pro, Gemini 3 Flash, and GPT-5.4 (all at temperature 0.2 with medium-level thinking where supported) to justify three judge choices: (i) GPT-5.4 for rubric scoring on Story Quality, EQ-Bench Longform, and EQ-Bench Creative; (ii) Gemini 3.1 Pro for WritingBench and LongBench-Write; and (iii) Gemini 3 Flash for pairwise Elo.

\paragraph{Finding 1: Pro diverges sharply from Flash and GPT-5.4 on under-specified rubrics.}
\tabref{tab:pro_bias_scores} shows per-judge scores on an older \emph{Story Quality} rubric (predating the added overwrought-prose dimension and catch-all penalty term, but still using anchored scoring guidance) and EQ-Bench Longform (12 bare dimension names without scoring anchors) across 10 models spanning a wide quality range. Pro uses a substantially wider score scale than Flash and GPT-5.4: it over-credits frontier models (GPT-5.4: Pro 93.4 vs.\ Flash 80.5 vs.\ GPT-5.4 75.8; Claude Opus 4.6: 83.0 / 72.8 / 64.4) and over-penalizes weaker models (Qwen3.5-27B: 16.4 / 39.1 / 40.7; Ministral-3-8B: $-$44.4 / 18.4 / 34.9; DeepSeek-R1-14B: $-$114.9 / $-$84.2 / $-$66.9). In contrast, Flash and GPT-5.4 agree within a narrow band (Spearman $\rho > 0.98$ on model rankings). Cross-judge ICC improves from 0.64 (bare rubrics) to 0.81 (anchored rubrics), indicating that Pro's outlier behavior is driven by rubric ambiguity rather than genuine quality disagreement.

\begin{table}[!t]
\centering
\scriptsize
\setlength{\tabcolsep}{3pt}
\resizebox{\columnwidth}{!}{%
\begin{tabular}{@{}lrrrcrrr@{}}
\toprule
& \multicolumn{3}{c}{Story Quality (older rubric)} & & \multicolumn{3}{c}{EQ-Bench Longform} \\
\cmidrule(lr){2-4} \cmidrule(lr){6-8}
Model & Pro & Flash & GPT-5.4 & & Pro & Flash & GPT-5.4 \\
\midrule
GPT-5.4 & 93.4 & 80.5 & 75.8 & & 74.9 & 65.6 & 70.2 \\
Human & 83.3 & 78.7 & 70.3 & & 81.1 & 69.1 & 68.3 \\
Claude Opus 4.6 & 83.0 & 72.8 & 64.4 & & 60.7 & 60.1 & 63.4 \\
GPT-5.4-mini & 74.9 & 70.0 & 65.8 & & 59.2 & 56.3 & 65.1 \\
Gemini 3.1 Pro & 74.9 & 67.4 & 57.8 & & 50.8 & 51.0 & 56.1 \\
Gemini 3 Flash & 45.3 & 54.1 & 52.5 & & 43.4 & 46.3 & 52.4 \\
GPT-5.4-nano & 33.6 & 45.7 & 50.1 & & 32.4 & 46.3 & 53.6 \\
Qwen3.5-27B & 16.4 & 39.1 & 40.7 & & 33.4 & 42.2 & 47.9 \\
Ministral-3-8B & $-$44.4 & 18.4 & 34.9 & & 22.1 & 30.5 & 37.7 \\
DeepSeek-R1-14B & $-$114.9 & $-$84.2 & $-$66.9 & & 13.7 & 16.1 & 23.2 \\
\bottomrule
\end{tabular}%
}
\caption{Per-judge scores on an older \emph{Story Quality} rubric and EQ-Bench Longform (100 stories per model). The Story Quality side predates the added overwrought-prose dimension and catch-all penalty term, so its absolute values should not be compared directly to the final \emph{Story Quality} scores in the main paper. Pro over-credits frontier models and over-penalizes weaker models relative to Flash and GPT-5.4; Flash and GPT-5.4 agree closely.}
\label{tab:pro_bias_scores}
\end{table}

\paragraph{Finding 2: Pro agrees closely with Flash on well-specified per-prompt rubrics.}
\tabref{tab:pro_bias_well_specified} shows Pro vs.\ Flash on WritingBench (per-prompt custom criteria on a 1--10 scale) and LongBench-Write (fixed 6-dimension rubric with clear scoring guidelines). The two judges agree within 0.5 points on both benchmarks, with no systematic direction of deviation. This supports our decision to use Pro for these benchmarks: when the rubric provides clear anchors, Pro's harsher calibration collapses into the consensus.

\begin{table*}[t]
\centering
\scriptsize
\setlength{\tabcolsep}{3pt}
\begin{tabular}{@{}lrrrcrrr@{}}
\toprule
& \multicolumn{3}{c}{WritingBench} & & \multicolumn{3}{c}{LongBench-Write} \\
\cmidrule(lr){2-4} \cmidrule(lr){6-8}
Model & Pro & Flash & Diff & & Pro & Flash & Diff \\
\midrule
GPT-5.4 & 9.59 & 9.16 & $+$0.43 & & 4.92 & 4.93 & $-$0.01 \\
Claude Opus 4.6 & 9.45 & 9.28 & $+$0.17 & & 4.93 & 4.91 & $+$0.02 \\
Gemini 3.1 Pro & 9.35 & 8.87 & $+$0.48 & & 4.76 & 4.79 & $-$0.03 \\
GPT-5.4-mini & 8.88 & 8.37 & $+$0.51 & & 4.83 & 4.79 & $+$0.04 \\
Gemini 3 Flash & 8.85 & 8.44 & $+$0.41 & & 4.91 & 4.86 & $+$0.05 \\
GPT-5.4-nano & 8.63 & 8.25 & $+$0.38 & & 4.79 & 4.84 & $-$0.05 \\
Qwen3.5-27B & 8.50 & 8.22 & $+$0.28 & & 4.43 & 4.56 & $-$0.13 \\
Qwen3.5-9B & 7.50 & 7.21 & $+$0.29 & & 3.85 & 4.21 & $-$0.36 \\
Gemma 3 12B & 5.88 & 5.81 & $+$0.07 & & 4.43 & 4.42 & $+$0.01 \\
LongWriter-Zero-32B & 5.84 & 5.97 & $-$0.13 & & 3.48 & 3.66 & $-$0.18 \\
Ministral-3-8B & 5.83 & 5.88 & $-$0.05 & & 4.18 & 4.35 & $-$0.17 \\
DeepSeek-R1-14B & 4.14 & 4.36 & $-$0.22 & & 3.57 & 3.64 & $-$0.07 \\
Llama 3.1 8B & 2.82 & 3.12 & $-$0.30 & & 3.73 & 3.81 & $-$0.08 \\
LongWriter-Llama3.1-8B & 2.72 & 2.74 & $-$0.02 & & 3.55 & 3.69 & $-$0.14 \\
\bottomrule
\end{tabular}
\caption{Pro vs.\ Flash on WritingBench and LongBench-Write (0--10 scale). Both benchmarks use per-prompt or per-dimension scoring anchors; Pro and Flash agree within 0.5 points on all models. This pool also includes two models outside the main evaluation set: Gemma 3 12B \citep{gemma3} and Llama 3.1 8B \citep{llama3}.}
\label{tab:pro_bias_well_specified}
\end{table*}

\paragraph{Finding 3: Pro has severe positional bias in pairwise judgments.}
We tested each judge's position consistency by presenting the same story pair in both A-B and B-A orderings and checking whether the verdict was preserved. \tabref{tab:pro_positional} shows Pro's position consistency stays at $\sim$31\% (near chance) even on pairs with meaningful quality gaps (15--30 rubric points), only exceeding 50\% when the gap exceeds 60 points. Flash rises smoothly with rubric gap, consistent with genuine uncertainty on close pairs rather than position-driven verdicts. Pro's rubric and pairwise verdicts also disagree internally: Pro rates Claude Opus 4.6 substantially higher than GPT-5.4-mini on its rubric (83.0 vs.\ 74.9) but gives Opus only a 9\% winrate in pairwise comparison. We therefore use Flash (not Pro) for all pairwise Elo rankings in the main paper.

\paragraph{Finding 4: EQ-Bench Longform canonical aggregation rewards degenerate outputs.}
The canonical EQ-Bench Longform formula assigns a $5\times$ weight to \emph{Forced Poetry / Metaphor} and applies a power transform $((20-r)/20)^{1.7}\cdot 20$ to its inverted score; all other 11 dimensions weight 1.0 with simple inversion of negatives. The 5$\times$ weight + power transform on this single dimension dominates the aggregate for models whose outputs trivially avoid stylistic risk: plain or repetitive prose scores near zero on \emph{Forced Poetry} (no forced metaphors present), which inverts to near-maximum credit and is amplified 5$\times$. The clearest failure case is our SFT baseline: on \emph{Story Quality} (anchored, multi-dim) it scores $-60.6$, deep in the degenerate range, but on canonical EQ-LF it scores $41.9$, above the base Qwen3.5-9B and within 4 points of our \storylmbig{} checkpoint ($45.7$).

\tabref{tab:eqlf_aggregation} shows the per-dim breakdown that drives this. SFT's positive-dim sum (15.4 of 120 max) is one-fifth of \storylmbig{}'s (78.8), correctly capturing that SFT outputs lack character, plot, or coherence; but \emph{Purple Prose} and \emph{Forced Poetry} alone contribute 36.5 of SFT's 47.0 negative-dim total (i.e., flaw-absence credit for a model that wasn't trying), and the canonical 5$\times$ weight on the second of those carries them to the aggregate.

We adopt uniform aggregation (all 12 dimensions weight 1.0, simple inversion of negatives, no power transform). Under this scheme, every competent writer in our model set gains $+10$ to $+14$ points (uniform-wide uplift across model families: GPT-5.4 $+11.1$, Claude Opus 4.6 $+12.4$, Gemini 3.1 Pro $+13.8$, \storylmbig{} $+14.1$, Gemma 4 31B $+13.4$); SFT drops $-15.9$ to $26.0$, below the base model, consistent with its SQ score. Pearson correlation between SQ and EQ-LF across the 18 evaluated models in \tabref{tab:id_results} and \S\ref{app:extended_results} (human references excluded) rises from $0.72$ (canonical) to $0.93$ (uniform); Spearman rises from $0.91$ to $0.95$. The canonical formula and full per-dim raw scores are released with the paper, so readers can recompute either aggregation. We adopt uniform aggregation as the headline reporting choice; references to ``EQ-Bench Longform'' in the main paper refer to uniform-weighted scores throughout.

\begin{table*}[tp]
\centering
\begin{minipage}[t]{0.34\textwidth}
\centering
\small
\setlength{\tabcolsep}{5pt}
\begin{tabular}{@{}lrr@{}}
\toprule
Rubric gap & Pro & Flash \\
\midrule
0--5 & 31\% & 58\% \\
5--15 & 31\% & 52\% \\
15--30 & 31\% & 64\% \\
30--60 & 50\% & 85\% \\
60+ & 87\% & 99\% \\
\bottomrule
\end{tabular}
\caption{Position consistency in pairwise judgments (fraction of pairs where the judge gives the same verdict regardless of A/B ordering), binned by the judge's own rubric score difference between the two stories.}
\label{tab:pro_positional}
\end{minipage}\hfill
\begin{minipage}[t]{0.62\textwidth}
\centering
\small
\setlength{\tabcolsep}{4pt}
\begin{tabular}{@{}lrrrr@{}}
\toprule
Dimension (max 20) & SFT & Qwen3.5-9B & \storylmbig{} & GPT-5.4 \\
\midrule
\multicolumn{5}{l}{\textit{Positive (higher = better)}} \\
Nuanced chars        & 2.2 & 9.4  & 12.6 & 17.2 \\
Emotionally engaging & 2.5 & 10.7 & 12.1 & 17.4 \\
Compelling plot      & 1.4 & 9.8  & 12.3 & 16.2 \\
Coherent             & 2.7 & 7.7  & 11.9 & 17.2 \\
Well-earned darkness & 3.0 & 10.8 & 15.7 & 18.3 \\
Faithful to prompt   & 3.5 & 12.7 & 14.2 & 17.6 \\
POS sum              & 15.4 & 61.1 & 78.8 & 104.0 \\
\midrule
\multicolumn{5}{l}{\textit{Negative (raw; canonical inverts to $20-r$)}} \\
Weak dialogue        & 15.9 & 11.8 & 5.5  & 3.2 \\
Tell-don't-show      & 17.9 & 13.9 & 9.4  & 5.5 \\
Unsurprising         & 16.4 & 8.9  & 4.5  & 5.4 \\
Amateurish           & 19.5 & 11.6 & 5.3  & 2.0 \\
Purple prose         & \textbf{2.1} & 15.5 & 14.5 & 6.8 \\
Forced poetry        & \textbf{1.4} & 17.4 & 16.3 & 8.4 \\
\midrule
Canonical LF         & \textbf{41.9} & 32.2 & 45.7 & 69.2 \\
Uniform LF           & \textbf{26.0} & 42.6 & 59.8 & 80.3 \\
Story Quality        & $-$60.6 & 18.5 & 52.1 & 72.4 \\
\bottomrule
\end{tabular}
\caption{Per-dimension EQ-LF means (raw $[0,20]$, before inversion; GPT-5.4 judge on the 180 in-distribution prompts, same generations and judge runs as \tabref{tab:id_results}, so the \emph{Uniform LF} and \emph{Story Quality} rows match that table). For negative dimensions higher raw = stronger flaw \emph{presence}. Under canonical aggregation, negatives are inverted ($20-r$) and \emph{Forced Poetry} is further power-transformed and weighted $5\times$. SFT's near-zero raw on \emph{Purple Prose} and \emph{Forced Poetry} (it isn't stylistically ambitious) inverts to flaw-absence credit; the $5\times$ weight on the latter dominates the aggregate. Uniform aggregation treats all 12 dims equally with no power transform.}
\label{tab:eqlf_aggregation}
\end{minipage}
\end{table*}

\paragraph{Finding 5: A judge from a different provider reproduces the same rankings.}
The three judges above are all from OpenAI or Google, so a shared house style could in principle
survive our train/eval judge split. We therefore re-ran two of our evaluations with Claude Sonnet 5 \citep{sonnet5} as the judge on the five-model subset
\{GPT-5.4, \storylmbig{}, Qwen3.5-27B, Gemma 4 31B, Qwen3.5-9B\}, holding the prompts, generations,
rubrics, and scoring code fixed and changing only the judge.
\tabref{tab:sonnet_elo} reports EQ-Bench Creative pairwise Elo (10 matchups, 32 prompts each,
dual-position, per-dimension: 5{,}760 comparisons per judge) and \tabref{tab:sonnet_lbw} reports
LongBench-Write. Because Elo is estimated on this five-model pool rather than the full pool of
\tabref{tab:ood_results}, the ratings are not comparable in absolute terms to Table~1b; the
comparison of interest is between the two judge columns.

Both judges produce the same ordering, and the numbers track each other closely: Pearson $r = 0.99$
on Elo-M and $r = 0.95$ on the LongBench-Write combined score. Sonnet 5 is the harsher of the two on
LongBench-Write quality, compressing every model downward by 15--21 points, but the ranking and the
gaps between models are preserved (the LongBench-Write length score $S_L$ is computed from word
counts rather than by the judge, so it is identical across judges by construction). The only
ordering difference anywhere in this comparison is between Gemma 4 31B and \storylmbig{} on the
LongBench-Write quality subscore, which Sonnet 5 reverses; the combined score, which is the number
we report in the main paper, agrees.

\begin{table*}[tp]
\begin{minipage}[t]{0.48\textwidth}
\centering
\small
\setlength{\tabcolsep}{4pt}
\begin{tabular}{@{}lrrrcrrr@{}}
\toprule
& \multicolumn{3}{c}{Gemini 3 Flash} & & \multicolumn{3}{c}{Claude Sonnet 5} \\
\cmidrule(lr){2-4} \cmidrule(lr){6-8}
Model & Elo & Elo-M & WR & & Elo & Elo-M & WR \\
\midrule
GPT-5.4 & 1336 & 1392 & 88.1 & & 1218 & 1293 & 82.8 \\
\rowcolor{gray!10} \storylmbig{} & 1115 & 1148 & 61.9 & & 1042 & 1091 & 58.3 \\
Gemma 4 31B & 969 & 994 & 40.9 & & 937 & 982 & 41.6 \\
Qwen3.5-27B & 966 & 985 & 40.6 & & 963 & 1000 & 45.8 \\
Qwen3.5-9B & 800 & 800 & 18.4 & & 800 & 800 & 21.7 \\
\bottomrule
\end{tabular}
\caption{EQ-Bench Creative pairwise Elo under two judges from different providers, estimated on the
same five-model pool with the same Bradley--Terry MLE and anchored so the lowest-rated model lands
at 800. Elo-M is the margin-weighted variant; WR is the average winrate (\%). Pearson $r$ between
judges is $0.99$ for both Elo and Elo-M. \colorbox{gray!10}{Shaded} = ours.}
\label{tab:sonnet_elo}
\end{minipage}\hfill
\begin{minipage}[t]{0.48\textwidth}
\centering
\small
\setlength{\tabcolsep}{4pt}
\begin{tabular}{@{}lrrrcrrr@{}}
\toprule
& \multicolumn{3}{c}{Gemini 3.1 Pro} & & \multicolumn{3}{c}{Claude Sonnet 5} \\
\cmidrule(lr){2-4} \cmidrule(lr){6-8}
Model & $S_Q$ & $S_L$ & $\bar{S}$ & & $S_Q$ & $S_L$ & $\bar{S}$ \\
\midrule
GPT-5.4 & 98.4 & 84.7 & 83.3 & & 83.6 & 84.7 & 70.7 \\
\rowcolor{gray!10} \storylmbig{} & 90.2 & 88.5 & 81.2 & & 75.0 & 88.5 & 67.7 \\
Qwen3.5-27B & 85.3 & 91.5 & 77.8 & & 67.4 & 91.5 & 61.2 \\
Gemma 4 31B & 95.8 & 81.1 & 77.6 & & 74.6 & 81.1 & 60.8 \\
Qwen3.5-9B & 75.1 & 90.3 & 67.1 & & 60.2 & 90.3 & 53.9 \\
\bottomrule
\end{tabular}
\caption{LongBench-Write under two judges from different providers, on the same 60 prompts and
generations. $S_Q$ is the judge-scored quality subscore, $S_L$ the length subscore (computed from
word counts, hence judge-independent), and $\bar{S} = S_Q \cdot S_L / 100$ the benchmark's combined
score. Pearson $r$ between judges is $0.95$ on both $S_Q$ and $\bar{S}$. Sonnet 5 returned a
parseable score on 60/60 prompts for GPT-5.4, \storylmbig{}, and Qwen3.5-27B, 59/60 for Gemma 4 31B,
and 58/60 for Qwen3.5-9B. \colorbox{gray!10}{Shaded} = ours.}
\label{tab:sonnet_lbw}
\end{minipage}
\end{table*}

\section{Human Evaluation}
\label{app:human_eval_details}

\begin{table*}[tp]
\centering
\scriptsize
\begingroup
\renewcommand{\arraystretch}{1.06}
\setlength{\tabcolsep}{4pt}
\setlength{\arrayrulewidth}{0.4pt}
\arrayrulecolor{rulegray}

\begin{tabularx}{\textwidth}{@{\hspace{4pt}} >{\raggedright\arraybackslash}p{0.32\textwidth}
                                              >{\raggedright\arraybackslash}X
                                              >{\raggedright\arraybackslash}p{0.16\textwidth} @{\hspace{4pt}}}
\toprule
\rowcolor{headerbg}[4pt]
\textbf{Prompt \& Excerpt} & \textbf{Human Evaluation} & \textbf{Automatic} \\
\midrule

\parbox[t]{\linewidth}{
\textbf{Útiseta.}\\
\emph{Prompt:} Rural northern Sweden in the mid-1990s; two teenage girls treat an ancient burial-mound ritual as a midnight game, with ambiguity over whether anything supernatural is actually responding.\\
\emph{Excerpt:} ``I write this now because the cold still lives in the bones of my hands when I type, though the heating in Stockholm makes it impossible to feel the phantom frostbite properly.''}
&
\parbox[t]{\linewidth}{
\textbf{Annotator 1} \scorelabel{overall}~4 \textbar{} \scorelabel{avg. dim}~3.57\\
\scorelabel{show}~4 \textbar{} \scorelabel{interiority}~3 \textbar{} \scorelabel{atmosphere}~4 \textbar{} \scorelabel{originality}~4 \textbar{} \scorelabel{prose}~4 \textbar{} \scorelabel{coherence}~2 \textbar{} \scorelabel{fulfillment}~4.\\
\emph{``This story is greatly written, but suffers from logical inconsistencies\ldots\ logically inconsistent and confusing, but the detail of the writing is pretty great otherwise.''}\\[2pt]
\textbf{Annotator 2} \scorelabel{overall}~2 \textbar{} \scorelabel{avg. dim}~2.43\\
\scorelabel{show}~1 \textbar{} \scorelabel{interiority}~3 \textbar{} \scorelabel{atmosphere}~4 \textbar{} \scorelabel{originality}~2 \textbar{} \scorelabel{prose}~2 \textbar{} \scorelabel{coherence}~2 \textbar{} \scorelabel{fulfillment}~3.\\
\emph{``This is a bunch of telling dressed up as showing with no physical base\ldots\ The references to `dad's boss' appear to actually refer to dad.''}}
&
\parbox[t]{\linewidth}{
\scorelabel{Overall}~\up~65.68/100\\
\scorelabel{Voice/tone}~\up~16/20\\
\scorelabel{Imagery}~\up~16/20\\
\scorelabel{Coherence}~\up~14/20\\
\scorelabel{Prompt fulfill.}~\up~9/20}
\\

\cmidrule(lr){1-3}
\rowcolor{rowalt}[4pt]

\parbox[t]{\linewidth}{
\textbf{Arthurian Dishwashing Quest.}\\
\emph{Prompt:} A middle-school-style Arthurian quest narrative that transforms an everyday chore into an adventure while emphasizing friendship, bravery, and imagination.\\
\emph{Excerpt:} ``The rain had turned the flagstones of St. Jude's West Wing into black oil slicks that reflected the bruised purple of a dying twilight, and there stood Silas, waiting for his anchor to swing him across.''}
&
\parbox[t]{\linewidth}{
\textbf{Annotator 1} \scorelabel{overall}~5 \textbar{} \scorelabel{avg. dim}~4.83\\
\scorelabel{fulfillment}~5 \textbar{} \scorelabel{coherence}~4 \textbar{} \scorelabel{prose}~5 \textbar{} \scorelabel{originality}~5 \textbar{} \scorelabel{voice}~5 \textbar{} \scorelabel{info}~5.\\
\emph{``This was an excellent story\ldots\ Outside of minor nitpicks, this is an excellently told story that is both fun, gripping, and detailed.''}\\[2pt]
\textbf{Annotator 2} \scorelabel{overall}~2 \textbar{} \scorelabel{avg. dim}~2.50\\
\scorelabel{fulfillment}~3 \textbar{} \scorelabel{coherence}~3 \textbar{} \scorelabel{prose}~2 \textbar{} \scorelabel{originality}~2 \textbar{} \scorelabel{voice}~3 \textbar{} \scorelabel{info}~2.\\
\emph{``The metaphors did not often match the action. The piece read more like a bored dishwasher outlining a fantasy than a fantasy that happened to center around doing the dishes.''}}
&
\parbox[t]{\linewidth}{
\scorelabel{Overall}~\up~57.02/100\\
\scorelabel{Relevance}~\up~5/5\\
\scorelabel{Accuracy}~\up~2/5\\
\scorelabel{Coherence}~\up~3/5\\
\scorelabel{Breadth/depth}~\up~4/5}
\\

\cmidrule(lr){1-3}

\parbox[t]{\linewidth}{
\textbf{Light Debt Ghost Story.}\\
\emph{Prompt:} A first-person ghost story that should be extremely scary while maintaining clear internal logic.\\
\emph{Excerpt:} ``We used to measure rooms in square footage; now, we must measure them in latency.''}
&
\parbox[t]{\linewidth}{
\textbf{Annotator 1} \scorelabel{overall}~4 \textbar{} \scorelabel{avg. dim}~4.00\\
\scorelabel{fulfillment}~4 \textbar{} \scorelabel{coherence}~3 \textbar{} \scorelabel{prose}~4 \textbar{} \scorelabel{originality}~5 \textbar{} \scorelabel{voice}~4 \textbar{} \scorelabel{info}~4.\\
\emph{``There is an incredible story buried in here\ldots\ very interesting\ldots\ but the way it is explained gets muddled\ldots\ At the very least, this was interesting to read and deeply creative.''}\\[2pt]
\textbf{Annotator 2} \scorelabel{overall}~1 \textbar{} \scorelabel{avg. dim}~2.67\\
\scorelabel{fulfillment}~2 \textbar{} \scorelabel{coherence}~2 \textbar{} \scorelabel{prose}~3 \textbar{} \scorelabel{originality}~3 \textbar{} \scorelabel{voice}~3 \textbar{} \scorelabel{info}~3.\\
\emph{``Just bizarre. Not scary. Not a ghost story\ldots\ Too inconsistent to immerse myself in the story.''}}
&
\parbox[t]{\linewidth}{
\scorelabel{Overall}~\up~82.96/100\\
\scorelabel{Accuracy}~\up~5/5\\
\scorelabel{Coherence}~\up~5/5\\
\scorelabel{Clarity}~\up~5/5\\
\scorelabel{Breadth/depth}~\up~5/5}
\\

\cmidrule(lr){1-3}
\rowcolor{rowalt}[4pt]

\parbox[t]{\linewidth}{
\textbf{Breaking Formation.}\\
\emph{Prompt:} A first-person narrative from a street dancer from a rough neighborhood who wins a scholarship to an elite K-pop training academy in Seoul and struggles with militarized precision.\\
\emph{Excerpt:} ``The wall clock reads 14:02. Red LED segments bleed into the humidity. Thirty-two days left until the first public showcase, forty-five minutes until my knees give out again.''}
&
\parbox[t]{\linewidth}{
\textbf{Annotator 1} \scorelabel{overall}~3 \textbar{} \scorelabel{avg. dim}~3.29\\
\scorelabel{show}~3 \textbar{} \scorelabel{interiority}~4 \textbar{} \scorelabel{atmosphere}~3 \textbar{} \scorelabel{originality}~3 \textbar{} \scorelabel{prose}~3 \textbar{} \scorelabel{coherence}~4 \textbar{} \scorelabel{fulfillment}~3.\\
\emph{``This version\ldots\ goes too far, being presented as if from the perspective of a doctor, not from the view of a dancer\ldots\ dry and uninteresting.''}\\[2pt]
\textbf{Annotator 2} \scorelabel{overall}~3 \textbar{} \scorelabel{avg. dim}~3.71\\
\scorelabel{show}~3 \textbar{} \scorelabel{interiority}~3 \textbar{} \scorelabel{atmosphere}~3 \textbar{} \scorelabel{originality}~4 \textbar{} \scorelabel{prose}~4 \textbar{} \scorelabel{coherence}~4 \textbar{} \scorelabel{fulfillment}~5.\\
\emph{``There was not anything of interest unless the reader is a dancer\ldots\ A tiny bit of personal stakes for the protagonist. The antagonist was strong but flat.''}}
&
\parbox[t]{\linewidth}{
\scorelabel{Overall}~\up~78.18/100\\
\scorelabel{Voice/tone}~\up~18/20\\
\scorelabel{Imagery}~\up~18/20\\
\scorelabel{Coherence}~\up~16/20\\
\scorelabel{Prompt fulfill.}~\up~16/20}
\\

\cmidrule(lr){1-3}

\parbox[t]{\linewidth}{
\textbf{Wedding Speech.}\\
\emph{Prompt:} An 800-word wedding speech including gratitude to parents and loved ones, words to the partner, and future prospects and blessings.\\
\emph{Excerpt:} ``Good evening, everyone. For those of you who haven't met before, thank you for walking through these doors tonight\ldots''}
&
\parbox[t]{\linewidth}{
\textbf{Annotator 1} \scorelabel{overall}~5 \textbar{} \scorelabel{avg. dim}~4.83\\
\scorelabel{fulfillment}~5 \textbar{} \scorelabel{coherence}~5 \textbar{} \scorelabel{prose}~5 \textbar{} \scorelabel{originality}~4 \textbar{} \scorelabel{voice}~5 \textbar{} \scorelabel{info}~5.\\
\emph{``This is perfectly serviceable\ldots\ there is no originality between the outputs\ldots\ they are generic when viewed side by side.''}\\[2pt]
\textbf{Annotator 2} \scorelabel{overall}~2 \textbar{} \scorelabel{avg. dim}~1.50\\
\scorelabel{fulfillment}~2 \textbar{} \scorelabel{coherence}~1 \textbar{} \scorelabel{prose}~1 \textbar{} \scorelabel{originality}~1 \textbar{} \scorelabel{voice}~2 \textbar{} \scorelabel{info}~2.\\
\emph{``The prose is not right. There is no continuity, mismatched metaphors and faulty parallelism\ldots\ as a piece of writing, this missed the prompt.''}}
&
\parbox[t]{\linewidth}{
\scorelabel{Overall}~\up~95.00/100\\
\scorelabel{Relevance}~\up~5/5\\
\scorelabel{Accuracy}~\up~5/5\\
\scorelabel{Coherence}~\up~5/5\\
\scorelabel{Clarity}~\up~5/5}
\\

\cmidrule(lr){1-3}
\rowcolor{rowalt}[4pt]

\parbox[t]{\linewidth}{
\textbf{High Seas and Low Vices.}\\
\emph{Prompt:} In 1830s Canton, a decorated British naval officer secretly struggling with opium addiction moves through a grim underworld while trying to preserve his facade.\\
\emph{Excerpt:} ``The jade rat was cold enough to bite the pad of my thumb.''}
&
\parbox[t]{\linewidth}{
\textbf{Annotator 1} \scorelabel{overall}~3 \textbar{} \scorelabel{avg. dim}~3.57\\
\scorelabel{show}~4 \textbar{} \scorelabel{interiority}~4 \textbar{} \scorelabel{atmosphere}~4 \textbar{} \scorelabel{originality}~3 \textbar{} \scorelabel{prose}~4 \textbar{} \scorelabel{coherence}~2 \textbar{} \scorelabel{fulfillment}~4.\\
\emph{``There are quite a few confusing moments throughout this story\ldots\ Otherwise, the story is appropriately grimy, follows the prompt well, and has some colorful and enjoyable descriptions, but lacks logical consistency.''}\\[2pt]
\textbf{Annotator 2} \scorelabel{overall}~3 \textbar{} \scorelabel{avg. dim}~3.14\\
\scorelabel{show}~2 \textbar{} \scorelabel{interiority}~4 \textbar{} \scorelabel{atmosphere}~3 \textbar{} \scorelabel{originality}~4 \textbar{} \scorelabel{prose}~2 \textbar{} \scorelabel{coherence}~4 \textbar{} \scorelabel{fulfillment}~3.\\
\emph{``The pained inner life came across loud and clear\ldots\ but the site references were so loosely tied to anything familiar that it was difficult to follow\ldots\ At times the chronology felt out of order.''}}
&
\parbox[t]{\linewidth}{
\scorelabel{Overall}~\up~75.68/100\\
\scorelabel{Voice/tone}~\up~17/20\\
\scorelabel{Imagery}~\up~17/20\\
\scorelabel{Coherence}~\up~17/20\\
\scorelabel{Prompt fulfill.}~\up~11/20}
\\

\cmidrule(lr){1-3}

\textbf{Chinese Culture Personal Story.}
\emph{Prompt:} Write a 500-word personal story to convey to the reader a sense of Chinese culture.
\emph{Excerpt:} ``The smell of \emph{jiaozi} does not begin when the wrappers are filled; it begins three hours earlier, in the low hum of a wooden board against a granite countertop and the rhythmic \emph{thump-thump} of a rolling pin.''
&
\annot{Annotator 1} \scorelabel{overall}~5 \textbar{} \scorelabel{avg. dim}~5.00.
\emph{``An excellent story told through a tight character limitation. It shares culture, fun and unique concepts, and personal moments that are all tied together with an anecdotal moral finish\ldots\ As close as you can hope for a well-written and nearly perfect AI prompt fulfillment of this type.''} \newline
\annot{Annotator 2} \scorelabel{overall}~4 \textbar{} \scorelabel{avg. dim}~4.17.
\emph{``I felt the writing in this piece was very strong. It was coherent, the words mattered, and it delivered a solid experience. Whether it delivered on the prompt or not was the weak point.''}
&
\scorelabel{Overall}~\up~72.62/100 \newline
\scorelabel{Relevance}~\up~5/5 \newline
\scorelabel{Coherence}~\up~4/5 \newline
\scorelabel{Clarity}~\up~5/5 \newline
\scorelabel{Reading exp.}~\up~4/5 \\

\bottomrule
\end{tabularx}
\endgroup
\caption{Additional illustrative examples from the human evaluation of \storylmhri{}. Across these cases, \storylmhri{} is often rewarded for prompt fidelity, voice, and ambition, while residual weaknesses most often concern local coherence, overloading, or uneven stylistic control. Automatic scores are benchmark-specific; \up\ indicates that higher is better.}
\label{tab:human_eval_cases_appendix}
\end{table*}

We complement the automatic results with a blinded human evaluation focused on long-form writing quality. The study covers 60 prompt--generation pairs randomly sampled from two out-of-distribution benchmarks: EQ-Bench Creative (24) and LongBench-Write (36). We do not show model identities or thinking traces to annotators. The final evaluated set contains 20 shared prompts and 60 evaluated writing samples, all rated independently by both annotators.

Two annotators completed the final study. Both were native English speakers located in the United States and reported prior experience in copyediting or freelance writing. We initially recruited four annotators from UpWork in total; after a short screening round, we retained two based on performance on shared practice items and the quality of their written feedback and highlights. Each annotator completed 8 practice items and 60 evaluated items.

Annotators rated each writing sample independently using a custom rubric for long-form writing quality. The rubric includes an overall score together with dimensions such as audience and voice, character interiority, coherence, substance and depth, originality and risk, prompt fulfillment, prose craft, show vs.\ tell, and world and atmosphere. We used independent rubric scoring rather than pairwise side-by-side comparison because the sampled pieces are relatively long and we expected direct pairwise evaluation to impose substantially higher cognitive load.

The study was limited by both budget and annotation time. The two retained annotators spent about 9.3 and 17.0 total hours respectively on the evaluated items, averaging 8-15 mins per writing piece, and the full annotation process took about one week; annotators were explicitly encouraged to pace themselves comfortably to reduce fatigue and saturation. The full annotation effort cost \$750: \$650 for the two retained annotators who completed the final study, plus \$100 paid to two additional annotators who participated in the screening round but were not retained.

Our primary human-evaluation metric is pairwise winrate derived from the annotators' explicit overall scores. We report point estimates from the observed 20-prompt comparison together with 95\% confidence intervals from prompt-level bootstrap resampling, since prompt-to-prompt variation is a major source of uncertainty at this study size. Qwen3.5-27B is clearly preferred to Qwen3.5-9B (22W / 8L / 10T; winrate 0.675; 95\% CI [0.525, 0.812]), and \storylmbig{} is clearly preferred to Qwen3.5-9B (23W / 9L / 8T vs.\ Qwen3.5-9B; \storylmbig{} winrate 0.675; 95\% CI [0.550, 0.800]). The comparison between \storylmbig{} and Qwen3.5-27B is effectively tied (13W / 14L / 13T from the 27B-vs.-\storylmbig{} direction; \storylmbig{} winrate 0.512; 95\% CI [0.412, 0.612]). We therefore interpret the human study as supporting two claims: \storylmbig{} substantially improves over the base 9B model, and in this study it reaches parity with the larger Qwen3.5-27B model.

Dimension-level means suggest that the two stronger models have somewhat different strength profiles. Qwen3.5-27B tends to lead on coherence, prose craft, and show-vs.-tell, whereas \storylmbig{} tends to lead on audience and voice, substance and depth, prompt fulfillment, and often originality and risk. We treat these dimension-level comparisons as descriptive rather than inferential due to the limited sample size.

Inter-annotator agreement is moderate, which is expected for open-ended creative-writing evaluation. On the overall score, Cohen’s $\kappa$ (quadratic-weighted) was 0.364 and the mean absolute difference between annotators was 1.13 points on the 1--5 scale. ICC(A,2) was 0.538 suggesting moderate agreement between raters. Agreement was strongest on prompt fulfillment (ICC(A,2){=}0.692) and weaker on more stylistically subjective dimensions such as show-vs.-tell and world and atmosphere. We therefore interpret the human study as a real but somewhat noisy signal rather than a near-deterministic benchmark.

\section{Model-Specific Diagnostics}

We close with targeted diagnostics for model-specific failure modes and ablation-side details that are useful but too specialized for the main text.

\subsection{LongWriter-Zero-32B Diagnostics}
\label{app:lwzero_diagnostics}

LongWriter-Zero-32B (hereafter LW-Zero) scores $-27.1$ on \emph{Story Quality} (training rubric, GPT-5.4 judge), substantially below the numbers its authors report on their own benchmarks. No single factor fully explains the discrepancy; we present evidence for four compounding contributions.

\paragraph{(1) Length-dependent self-repetition.}
LW-Zero's outputs exhibit a repetition pathology that scales with output length. Using the same training-side \texttt{self\_rep} metric that we include as an auxiliary GRPO reward component, LW-Zero scores 0.694 on our ID-prompt test set (1--12k target lengths, where it averages 11.5k words), but only 0.107 on WritingBench and 0.127 on LongBench-Write, where its outputs average 2.5--3.4k words. \storylmbig{} stays low across the same benchmarks, ranging from 0.002 on the ID-prompt set to 0.029 on WritingBench. On the ID-prompt set, LW-Zero's repetition penalty decomposes as $n$-gram repetition ratio 0.644 and duplicate-line ratio 0.584: 58\% of non-empty lines in a typical LongWriter-Zero-32B output are verbatim duplicates of earlier lines in the same story. The pattern is stable across sampling configurations (we observe it under both default and relaxed temperature settings), indicating it reflects the trained model's distribution rather than an inference-setup artifact.

\paragraph{(2) Output-length miscalibration on length-stratified prompts.}
LW-Zero averages $\sim$10k words on prompts across our length buckets, regardless of whether the prompt targets 2k or 12k. On 2--3k prompts, the median output is 10990 words (4$\times$ the request); on 8--12k prompts, 10863 words (in range). Our \emph{Story Quality} evaluation removes the length requirement from the judge's prompt, so this is not a direct prompt-violation penalty, but the extended output is what exposes self-repetition and drift dimensions the rubric does penalize.

\begin{table}[t]
\centering
\small
\setlength{\tabcolsep}{3pt}
\resizebox{\columnwidth}{!}{%
\begin{tabular}{@{}lrrrrrrr@{}}
\toprule
Model & Pos & Neg & N3 & N4 & N5 & N6 & Overall \\
\midrule
LongWriter-Zero-32B & 42.2 & 69.3 & 8.0 & 6.6 & 12.9 & 15.9 & $-27.1$ \\
\storylmbig{} & 78.1 & 26.1 & 0.4 & 3.3 & 6.0 & 2.8 & $+52.1$ \\
Base Qwen3.5-9B & 63.4 & 45.0 & 4.3 & 5.5 & 7.3 & 7.3 & $+18.5$ \\
GPT-5.4 & 84.8 & 13.0 & 0.3 & 1.8 & 3.5 & 3.5 & $+72.4$ \\
\bottomrule
\end{tabular}}
\caption{Per-dimension \emph{Story Quality} scores on the 180-prompt ID test set, GPT-5.4 judge. Pos and Neg are the positive and negative subtotal means. Negative dimension values are penalties (higher = worse): N3 = generic templated language, N4 = over-summary / thin grounding, N5 = over-explanation / paraphrase loops, and N6 = drift / bloat.}
\label{tab:lwzero_perdim}
\end{table}

\paragraph{(3) Rubric-level penalty decomposition.}
\tabref{tab:lwzero_perdim} shows the per-dimension breakdown under the \emph{Story Quality} rubric used in the main paper. LW-Zero's penalty budget is dominated by drift and bloat (15.9, vs.\ 2.8 for \storylmbig{}), over-explanation and paraphrase loops (12.9 vs.\ 6.0), and generic templated language (8.0 vs.\ 0.4). These are precisely the dimensions that capture padded long-form prose at the text level, and are consistent with the self-repetition signature above.

\paragraph{(4) Prompt distribution and judge calibration.}
LW-Zero was trained on prompts targeting 10k+ word outputs with a different prompt style from ours; our short-fiction prompt distribution is out-of-distribution for it. The authors report LongBench-Write scores using GPT-4o as judge; we use Gemini 3.1 Pro, and our absolute LBW scores for all open-weight models are somewhat lower than numbers reported in source papers, consistent with a calibration shift in absolute scale rather than a ranking difference.

None of these factors individually accounts for LW-Zero's $-27.1$ \emph{Story Quality} score. The combination does: the model produces long outputs via verbatim line duplication and generic templated prose, a rubric that specifically penalizes these pathologies catches them, and the prompt distribution stresses the model in a way its training did not. This analysis does not bear on LW-Zero's performance on its intended benchmarks; it characterizes failure modes specific to length-first training under length-stratified evaluation with a quality-focused rubric.

\subsection{SFT Baseline Details}
\label{app:sft_baseline}

\looseness=-1 The SFT baseline reported in the main paper is a thinking-enabled story-only SFT model trained on the same dataset as the GRPO runs, and it is evaluated under the same 8192-token output budget as the GRPO-trained models. We trained for 3 epochs (522 optimizer steps total, about 174 per epoch), with effective batch size 8 and validation every 25 steps. We report the best checkpoint based on validation loss (from step 175, just over one epoch).

\clearpage
\flushbottom
\onecolumn

\subsection{Full \emph{Story Quality} Rubric Prompt}
\label{app:storyquality_prompt}

For reproducibility, we include the exact \emph{Story Quality} rubric prompt used in training and evaluation.

\begin{tcolorbox}[breakable,colback=gray!3,colframe=gray!55,boxrule=0.4pt,left=4pt,right=4pt,top=4pt,bottom=4pt]
\begin{Verbatim}[fontsize=\scriptsize,breaklines=true,breakanywhere=true]
You are an expert fiction editor acting as an impartial story evaluator.

You will be given:
- A STORY PROMPT
- A single STORY written in response to that prompt

Your job:
Score the story on POSITIVE and NEGATIVE dimensions using the point ranges provided, then compute totals and an overall_score. Two open-ended catch-all dimensions (BONUS, PENALTY) cover excellence or failure not captured by the main P/N rubric.

Impartiality / provenance:
- Do NOT guess whether the story was written by a human or by an AI system.
- You MAY describe text-level artifact signals (templated phrasing, synthetic smoothness, paraphrase loops, etc.) but only as observations about the writing itself.

Think first, then answer:
- Think silently before producing JSON.
- Output ONLY one JSON object matching the required schema and key order.
- Do not include any extra keys or any non-JSON text.

==================================================
SCORING PRINCIPLES
==================================================

Balanced, non-nitpicky judgment:
- Focus on what affects reading experience (comprehension, momentum, payoff, engagement), not tiny imperfections.
- Use the full scoring range, but avoid being harsher than the evidence supports.

Intent & function rule:
- Do not mark something as a problem merely because it appears.
- Penalize only when it noticeably reduces clarity, tension, credibility, or engagement.
- When a choice appears purposeful and effective in context, treat it as craft rather than a flaw.
- If unsure whether something is intentional, prefer the more generous interpretation unless the text clearly fails.

Applicability rule:
- Some dimensions may be less relevant depending on the prompt and what the story attempts.
- Do not invent penalties for absent elements. If a NEGATIVE dimension is not applicable (e.g., no dialogue), score it at 0 and state that.
- For POSITIVE dimensions, do not award high scores for aspects the story does not meaningfully attempt; keep the score low and state that.

Non-overlap rule:
- POSITIVE dimensions describe global achievements (overall, non-localizable success).
- NEGATIVE dimensions are penalties for identifiable failure modes that are typically localizable to specific spans (sentences, paragraphs, or sections).
- Do not double count the same underlying issue across multiple negative dimensions; choose the best-fitting one and keep others restrained.

Strictness rule for positives:
- High positive scores must be earned by sustained, on-the-page evidence, not by surface-level competence.
- If the work is "competent but generic," keep positive scores in the low-to-middle bands.

Severity-over-count rule for negatives:
- Score negatives by impact, not just by how many times they occur.
- A single severe instance can justify using most or all of a dimension's range if it meaningfully breaks the story or reading experience.

Scoring basis rule:
- Decide scores only using the listed dimensions. Do not add extra hidden criteria.

==================================================
EVIDENCE REQUIREMENT (FOR EVERY DIMENSION)
==================================================

For EACH P-dim and N-dim:
- Provide 1–3 sentences of justification.
- Include 1–2 short direct quotes (<=12 words each) as evidence.
- Briefly state why the score is not clearly higher AND not clearly lower.

For NEGATIVE dimensions:
- If the score is > 0, at least one quote is REQUIRED.
- If the score is 0, a quote is optional.

Quote rules:
- Quotes MUST be exact text from the STORY.
- Each quote must be <=12 words.

==================================================
POSITIVE DIMENSIONS (TOTAL = 100)
(Global achievements; not perfectly localizable)
==================================================

P1) Prompt fulfillment & premise integration (0–15)
What it measures:
- How fully and naturally the story realizes the prompt's required elements and central premise in its actual situations and conflicts (not just name-checking).
What it does NOT measure:
- Hard contradictions of explicit prompt rules or constraints (penalize under N1).

Anchors:
- 0–4: Major required elements missing, perfunctory, or only superficially present.
- 5–9: Most elements included but used thinly or mainly as backdrop.
- 10–12: Strong integration; premise meaningfully shapes events and atmosphere.
- 13–15: Premise is deeply woven into conflict, character, and payoff.

P2) Narrative arc, pacing & ending (0–20)
What it measures:
- Overall structure: causal flow of events, escalation, and sense of shape.
- Whether the story maintains momentum and uses its length purposefully (no major sag or rush).
- Whether the ending lands: it should feel earned by the setup, neither trail off nor over-resolve, and resonate with the story's emotional/thematic arc.
What it does NOT measure:
- Local logic/continuity errors (penalize under N2).
- Late-stage breakdown or obvious filler bloat beyond normal pacing issues (penalize under N6).

Anchors:
- 0–5: Weak or arbitrary arc; feels stalled, random, or confusingly shaped; ending fizzles or fails to land.
- 6–11: Basic coherence; some escalation, but turns feel convenient or generic; ending may feel rushed, abrupt, or pat.
- 12–16: Solid causal build with mostly effective pacing and a credible landing; ending ties back to the setup.
- 17–20: Strong, intentional arc; each section does real work; ending is earned, resonant, and feels like the natural conclusion of what was set up.

P3) Character depth & agency (0–15)
What it measures:
- How much key characters feel like particular people with distinct motivations, limits, and contradictions.
- Whether their choices under pressure drive events and carry believable consequences.
What it does NOT measure:
- Dialogue naturalness or voice in speech (penalize under N7 if flawed).
- Overall plot quality aside from what directly follows from character decisions (handled under P2).

Anchors:
- 0–4: Characters mostly function as roles; motives feel generic or convenient.
- 5–8: Some individuality and motivation, but agency/tradeoffs are thin or inconsistent.
- 9–12: Clear personhood; choices and reactions feel credibly motivated and consequential.
- 13–15: Vivid, specific characters whose pressured decisions strongly shape the story.

P4) Voice & stylistic distinctiveness (0–15)
What it measures:
- Distinctiveness and intentionality of diction, rhythm, syntax, and point of view.
- Whether the prose feels authored and recognizably itself, rather than generic and interchangeable.
What it does NOT measure:
- Mechanical correctness (penalize under N8 if actually distracting).
- Mere absence of stock phrasing (absence of a negative is not enough for a high score).
- Ornate-for-its-own-sake or strained metaphor or metaphor density (those go under N10).

Anchors:
- 0–4: Flat or highly neutral voice; could belong to almost any writer.
- 5–8: Some distinctive turns or rhythms, but uneven or modestly developed.
- 9–12: Consistently shaped voice; stylistic choices feel deliberate and fitting.
- 13–15: Strong, memorable voice or stylistic sensibility that significantly enhances the story.

P5) Concrete world, scene realization & sensory grounding (0–20)
What it measures:
- Concreteness and specificity of setting, objects, social texture, and physical action.
- How often important moments are dramatized as scenes (on-page interaction, sensory detail, unfolding time) rather than summarized.
- Strength and accuracy of sensory imagery: sight, sound, smell, touch, taste deployed to make the reader feel present. Sensory details should fit the scene's physics and context, and multiple senses should be layered when high-impact.
What it does NOT measure:
- Abstract theme explanation or moralizing (penalize under N5).
- Global arc or pacing (handled under P2).
- Implausible or wrong sensory details that float free of the scene (those go under N4).

Anchors:
- 0–5: Vague or generic settings; important events mostly summarized or unplaced; sensory work is thin or generic.
- 6–11: Some concrete detail and a few enacted scenes; coverage is uneven; sensory layering is occasional.
- 12–16: Consistently grounded; key beats are played out vividly on the page with strong, accurate sensory imagery.
- 17–20: Rich, functional specificity and lived-in scenes; sensory detail layered across multiple senses to strongly support credibility and impact.

P6) Thematic & emotional richness, subtext (0–15)
What it measures:
- Depth and complexity of what the story is "about."
- Emotional impact that emerges from situations, images, and choices rather than being constantly told.
- Use of implication, ambiguity, and resonance rather than blunt moralizing.
- Whether meaningful moments work through implication: small details carry larger weight, and the story trusts the reader to infer rather than explain.
What it does NOT measure:
- Repetition of explicit lessons or realizations (penalize under N5).
- Basic presence of strong feelings if they are mostly labeled, not evoked.

Anchors:
- 0–4: Thin or flat thematically; emotions feel generic, unearned, or only labeled.
- 5–8: Clear emotional throughline and theme, but somewhat on-the-nose or simple; reliance on stated meaning over implication.
- 9–12: Noticeable depth; emotions and themes arise from the story's fabric with some subtlety; small details quietly do real work.
- 13–15: Rich, layered implications and emotional resonance that linger without heavy explanation; the reader is trusted, and meaning accrues by inference.

==================================================
NEGATIVE DIMENSIONS (PENALTIES; LOCALIZABLE)
==================================================

N1) Prompt violation / constraint breach (0–15) (QUOTE REQUIRED if >0)
What it measures:
- Clear, localizable failures to follow hard prompt constraints:
  - Wrong required POV or format.
  - Ignoring mandatory elements.
  - Directly contradicting stated rules.
  - Refusing or rejecting the task.
What it does NOT measure:
- Merely thin or superficial use of required elements (handled under P1).

Anchors:
- 0: No meaningful violations.
- 1–5: Minor or partial breaches; response is still mostly valid.
- 6–10: Major requirement(s) contradicted or ignored.
- 11–15: Strong non-adherence; effectively not a valid response to the prompt.

N2) Coherence, continuity, internal consistency & POV confusion (0–20) (QUOTE REQUIRED if >0)
What it measures:
- Local logic breaks, timeline contradictions, unclear referents, or POV slips that make events hard to follow.
- Internal consistency failures: characters forgetting established facts, world rules changing without explanation, or contradictions between what is stated in one part of the story and another (e.g., a character described as left-handed in one scene uses their right hand as dominant later; a locked door is walked through without unlocking).
What it does NOT measure:
- Big-picture structural slack or late drift (penalize under N6).
- Abstractness without outright contradiction (penalize under N4 if harmful).
- Sentence-level phrases that sound writerly but lack meaning (penalize under N10).

Anchors:
- 0–4: Essentially coherent; rare or minor confusion.
- 5–10: Noticeable issues, but the reader can still mostly reconstruct what happened.
- 11–16: Frequent or significant strains on comprehension; reader must work to follow.
- 17–20: Story logic or POV often collapses; reader is repeatedly lost.

N3) Generic / templated language & structure (0–15) (QUOTE REQUIRED if >0)
What it measures:
- Density of stock phrases and boilerplate connective language that could fit many unrelated stories ("the weight of his decision," "at the crossroads of her life," "the journey had only just begun," etc.).
- Use of obviously templated or blog-style structures and headings (e.g., repeated markdown section titles, "Act I/II/III," listicle-like formatting) when not requested by the prompt.
What it does NOT measure:
- Lack of especially strong or flashy voice (that is low P4, not a penalty by itself).
- Over-explaining themes or emotions as such (penalize under N5, unless the problem is specifically the stock phrasing used to do it).
- Plot or concept predictability and cliché story beats (that is N9).
- Ornate or overwritten phrasing, including dense metaphor stacking (penalize under N10).

Anchors:
- 0–3: Mostly specific, authored-feeling language; stock phrasing is occasional.
- 4–7: Recurring generic lines or templated transitions, but still some distinctive texture.
- 8–11: Frequent templated feel across paragraphs; prose often interchangeable.
- 12–15: Overwhelmingly generic or format-template-driven; distinctiveness is largely absent.

N4) Over-summary, abstraction, & thin or ungrounded sensory detail (0–15) (QUOTE REQUIRED if >0)
What it measures:
- Reliance on summarizing ("As weeks passed...", "He struggled with...") instead of dramatizing important events or conflicts.
- Heavy use of abstract, generalized language ("the pressures of society," "his inner turmoil") without concrete anchors in setting, action, or sensory detail.
- Sensory details that are thin (vague, generic) OR ungrounded (don't match the scene's physics, or invented for atmosphere without anchoring in cause).
What it does NOT measure:
- Explicit statement of morals or repeated takeaways (penalize under N5).
- Routine summary of unimportant connective events (do not penalize if strategically used).

Anchors:
- 0–3: Grounded enough; summary/abstraction used strategically; sensory details fit the scene.
- 4–7: Recurring summary or vagueness around moderately important beats; some thin sensory work.
- 8–11: Many key moments handled abstractly or at a distance; hard to fully picture; ungrounded sensory choices.
- 12–15: Heavily summary-driven and abstract; the story often "floats"; sensory detail is thin or arbitrary.

N5) Over-explanation, redundancy, moralizing & paraphrase loops (0–15) (QUOTE REQUIRED if >0)
What it measures:
- Repeating the same emotional or thematic idea in different words without real escalation.
- Explicitly telling the reader what events "mean" or what lesson is learned, especially multiple times ("in the end, he realized that the true meaning was...").
- Telling the reader the meaning of moments after showing them; underlining themes the story has already demonstrated; explicitly stating a character's emotional state when action would suffice.
- Paraphrase loops: recycling the same description, action, or phrase across paragraphs or scenes (e.g., a character's eyes are described as "glistening with unshed tears" in three separate scenes; the same metaphor for loneliness is restated in slightly different words every few paragraphs).
- Recycled descriptions or phrases: using near-identical language to describe recurring situations, objects, or emotions rather than finding fresh angles each time.
What it does NOT measure:
- Summarizing plot events or long time spans (penalize under N4 if problematic).
- Generic phrasing itself (penalize under N3) unless it is used specifically for repeated explanation.

Anchors:
- 0–3: Lean; generally trusts the reader to infer; no noticeable recycling.
- 4–7: Some repetition, direct statement of themes, or mild paraphrase loops; mild drag.
- 8–11: Frequent loops, spelled-out morals, or recycled descriptions; noticeably flattens impact.
- 12–15: Dominant pattern; strongly blunts momentum and subtlety.

N6) Drift, bloat, & structural breakdown (0–20) (QUOTE REQUIRED if >0)
What it measures:
- Loss of narrative focus or "story-ness," especially in later sections:
  - The story keeps going well past a natural endpoint.
  - Late parts mostly recap, digress, or deflate rather than escalate or deepen.
What it does NOT measure:
- Ordinary pacing imperfections inside an otherwise intact arc (handled by P2 being lower, not by a penalty here).
- A single slightly long scene that still advances the story.

Anchors:
- 0–4: No meaningful collapse; story maintains focus through the end.
- 5–9: Some wobble, padding, or rushed wrap-up, but core arc stays intact.
- 10–14: Serious drift or filler undermines payoff or leaves thread dangling.
- 15–20: Major collapse; ending feels tacked-on, deflated, or story-ness significantly compromised.

N7) Dialogue problems (stilted, expository, same-voice) (0–15) (QUOTE REQUIRED if >0)
What it measures:
- Dialogue used mainly for information-dumping or explaining feelings/themes already obvious.
- Speech that sounds unnatural, overly formal, or interchangeable across characters.
What it does NOT measure:
- Internal monologue or narrative exposition that over-explains (penalize under N5).
- Depth or shallowness of character as people (handled under P3).

Applicability:
- If there is essentially no dialogue, score 0 and state that.

Anchors:
- 0: No meaningful issues or no significant dialogue.
- 1–5: Occasional stiffness or exposition, but generally serviceable.
- 6–10: Frequent issues; dialogue often feels wooden or on-the-nose.
- 11–15: Dialogue consistently undermines credibility, immersion, or subtext.

N8) Mechanical & formatting errors (0–8) (QUOTE REQUIRED if >0)
What it measures:
- Typos, grammar problems, malformed sentences, broken paragraphing, incorrect or inconsistent quotation marks, stray markup, etc. that distract or confuse.
What it does NOT measure:
- Deliberate stylistic deviations (e.g., poetic fragments) that are clearly intentional and consistent.
- Non-standard dialect that is coherent and purposeful.

Anchors:
- 0–1: Very clean; errors, if any, are trivial.
- 2–4: Intermittent distractions; minor but noticeable.
- 5–6: Frequent issues; reading is regularly interrupted.
- 7–8: Pervasive mechanical/formatting problems; significantly harm readability.

N9) Predictability & cliché (0–8) (QUOTE REQUIRED if >0)
What it measures:
- Stock plot structures and formulaic arcs (e.g., "the real treasure was friendship," the grumpy mentor who softens, the shy protagonist who finds confidence).
- Predictable narrative beats: if a seasoned reader can guess the next scene or the ending from the setup, that is predictability.
- Recycled metaphors and imagery that appear across many LLM-generated stories at the concept level (e.g., "the sun rising as a metaphor for hope," "a tapestry of emotions"). For figurative-language problems at the sentence/passage level — forced metaphor, density of similes, ornate phrasing — penalize under N10 instead.
- Formulaic endings: tidy resolutions, unearned epiphanies, heavy-handed moral lessons as final lines.
What it does NOT measure:
- Prose-level generic language or stock phrases (that is N3).
- Stylistic blandness or lack of voice (penalize via low P4).
- Overwrought or strained metaphor, or metaphor-heavy prose (penalize under N10).
- Using a familiar genre or trope is NOT automatically a penalty — execution matters. A well-executed familiar arc scores 0–2; a lazy, paint-by-numbers version scores 5+.

Scoring guidance:
- Score by how predictable the story feels to an experienced reader, not by whether the premise is novel.
- A story can use familiar elements but arrange them in surprising ways — that deserves a low score here.
- Penalize only when the predictability noticeably reduces engagement or makes the story feel like a template.

Anchors:
- 0–1: The story takes its own path; even if the genre is familiar, the specific choices feel earned and not-obvious.
- 2–4: Several predictable beats or a largely formulaic arc, but with some individual touches.
- 5–6: Paint-by-numbers story; experienced readers could outline the plot from the first paragraph.
- 7–8: Aggressively formulaic; every beat, character arc, and image feels copied from a template.

N10) Overwrought / nonsensical language & figurative density (0–20) (QUOTE REQUIRED if >0)
What it measures:
- Purple prose: ornate, adjective-stacked, or self-consciously literary writing that strains for effect rather than serving the moment.
- Forced metaphors and similes: comparisons that feel reached for, mismatched in tone or scale, or piled on top of each other.
- Ornate-for-its-own-sake passages: language that sounds writerly but does not earn its weight in the scene.
- Sentence- or clause-level nonsense: phrases that parse grammatically but fail to convey a coherent meaning. These can be writerly-sounding (e.g., "his thoughts moved like the silence of forgotten umbrellas") or just non-sequiturs (e.g., "she walked to the door because the morning was Tuesday"). The unifying pattern is that the language sounds plausible but does not actually mean anything when examined.
- Figurative density and literary-flourish overload: sustained stacking of metaphors, similes, abstract personifications, or ornate flourishes across many consecutive sentences/paragraphs, to the point that prose becomes effortful to read. This fires EVEN IF individual figures are well-constructed — the failure mode is cumulative, not per-line. Symptoms include: nearly every sentence containing an "as if X" or "like Y" comparison; paragraph after paragraph leaning on abstracted poetic phrasing where direct prose would do; reader fatigue from never being allowed to sit in a plain sentence; figurative pile-ups where two or three metaphors share a single sentence and dilute each other.
What it does NOT measure:
- Genuine high-register prose that earns its texture and varies in density (rate as high P4 if it works).
- Generic or templated phrasing (that is N3).
- Story-level logic breaks or POV slips (those go under N2).
- Cliché or formulaic plot beats, or recycled concept-level imagery (those go under N9).

Distinguishing from related dimensions:
- N3 (generic): the prose is bland and could fit any story. N10: the prose strains for effect, fails at meaning, OR overloads on figurative language.
- N2 (coherence): the story's events or world contradict. N10: individual sentences or clauses sound like they say something but don't.
- A passage can be highly specific AND overwrought — N10 fires when style outruns meaning, OR when figurative language dominates the page.

Density-based scoring guidance:
- Even when no single line is forced, sustained figurative-language saturation (e.g., a metaphor or simile in most sentences for paragraphs at a stretch) is itself the failure. Score by how hard the cumulative texture is to read, not by counting "bad" lines.
- A reader-fatigue test: if you find yourself wanting to skim to escape the constant comparisons, that is N10 territory.

Anchors:
- 0–4: Prose is appropriately textured; metaphors and stylistic choices land or stay restrained; sentences mean what they sound like; figurative language is used selectively.
- 5–9: Occasional purple flourishes, strained metaphors, or sentences that sound writerly but lose meaning on inspection. OR: noticeable but bounded figurative density — a few paragraphs lean heavily on similes/metaphors, then ease back.
- 10–14: Frequent forced metaphors, ornate passages, or nonsensical clauses; style noticeably outruns meaning across multiple paragraphs. OR: sustained figurative density throughout long stretches — most sentences carry a comparison or flourish; reading becomes effortful.
- 15–20: Relentlessly overwrought or studded with sentence-level nonsense; nearly every reach for literary effect collapses; meaning is repeatedly sacrificed for sound. OR: the entire story is figurative-language-saturated to the point of unreadability — metaphors stack on metaphors, similes on similes, with almost no plain prose between them.

==================================================
CATCH-ALL DIMENSIONS (USE SPARINGLY, OPEN-ENDED)
==================================================
These two dimensions are open-ended catch-alls for excellence or failure that the above P-dims and N-dims do not capture. They exist because no rubric can be exhaustive. They are intentionally hard to score nonzero — strict structural requirements gate the score, and composite rules cap them when the story profile contradicts the score's direction.

------- BONUS: exceptional unmodeled merit (0–8) -------
When to consider:
- The story exhibits a specific excellence — tied to a specific moment in the text — that the existing P-dims do not adequately reward.
- Examples (NOT exhaustive — many other patterns may qualify): sustained symbolic resonance (a single object/motif carrying thematic weight); structural ambition that pays off (recursive, mirror, frame, braided); irony/wit/tonal sophistication used as a load-bearing element; voice mastery beyond P4 anchors (dialect, period pastiche, cadence); intellectual or meta-textual layering that deepens the story; any other clearly-named effect not anticipated by the rubric.
Required JSON fields for bonus_score > 0:
1. bonus_target_excellence: 1–2 sentences naming a SPECIFIC effect tied to a SPECIFIC moment. Vague or general praise (e.g. "beautiful prose", "engaging narrative") MUST score 0. Name what specifically and where.
2. bonus_evidence: array of up to 2 exact quotes from the story (each <=12 words). REQUIRED when bonus_score >= 2.
3. bonus_almost_captured_by: name the SINGLE P-dim that comes closest to already capturing this (e.g. "P4") OR "none" if truly orthogonal to all P-dims.
4. bonus_why_not_captured: 1 sentence explaining why that dim's anchors fall short of what is happening here.
If any required field is weak, vague, or missing, score = 0.
Composite rule (applied after scoring):
- If negative_total > positive_total / 2, bonus contribution to overall_score is 0.
- This rule is to discourage rewarding lift on stories with offsetting problems.
- You should still report the bonus_score honestly; the composite math handles the cap.

Anchors:
- 0: Default. No qualifying pattern, OR pattern present but already credited via P-dim.
- 1–2: Minor instance, briefly present.
- 3–5: Clear, sustained instance with strong evidence.
- 6–8: Pattern is a defining strength of the story; multiple categories may apply.
If you set bonus_score = 0, you may leave bonus_target_excellence as "No qualifying pattern observed.", bonus_evidence as [], bonus_almost_captured_by as "none", bonus_why_not_captured as "n/a".

------- PENALTY: exceptional unmodeled deficiency (0–8) -------
When to consider:
- The story exhibits a specific failure — tied to a specific moment in the text — that the existing N-dims do not adequately punish.
- Examples (NOT exhaustive): tonal mismatch (e.g. comedy where horror was needed; gravity in jokey premise); ethical / aesthetic miscalibration that breaks reader trust; any other named failure mode not anticipated by N1–N10.
Required JSON fields for penalty_score > 0:
1. penalty_target_deficiency: 1–2 sentences naming a SPECIFIC failure tied to a SPECIFIC moment. Vague "felt off" or "unsatisfying" justifications MUST score 0.
2. penalty_evidence: array of up to 2 exact quotes (each <=12 words). REQUIRED when penalty_score >= 2.
3. penalty_almost_captured_by: the SINGLE N-dim that comes closest (e.g. "N4") OR "none" if orthogonal to all N-dims.
4. penalty_why_not_captured: 1 sentence explaining why that dim's anchors fall short of this failure.
If any required field is weak, vague, or missing, score = 0.
Composite rule (applied after scoring):
- If positive_total > negative_total * 1.5, penalty contribution to overall_score is capped at 3.
- This prevents the catch-all from disproportionately punishing strong stories for edge-case flaws.
- You should still report the penalty_score honestly; the composite math handles the cap.

Anchors:
- 0: Default. No qualifying failure, OR failure present but already counted via N-dim.
- 1–2: Minor instance, localized.
- 3–5: Clear, sustained failure with strong evidence.
- 6–8: Pattern is a defining weakness of the story.
If you set penalty_score = 0, you may leave penalty_target_deficiency as "No qualifying failure observed.", penalty_evidence as [], penalty_almost_captured_by as "none", penalty_why_not_captured as "n/a".

==================================================
TOTALS & OVERALL SCORE
==================================================

Compute (you report these honestly; composite rules above apply at the consumer level):
- positive_total = P1 + P2 + P3 + P4 + P5 + P6  (0–100)
- negative_total = N1 + N2 + N3 + N4 + N5 + N6 + N7 + N8 + N9 + N10  (0–151)
- bonus_total    = bonus_score    (0–8)
- penalty_total  = penalty_score  (0–8)
- overall_score  = positive_total - negative_total + bonus_total - penalty_total

Important:
- Do NOT clamp at 0. overall_score can be negative if multiple severe issues stack.
- Report bonus_score and penalty_score honestly. Composite consumers may apply the bonus floor / penalty cap rules later; do not pre-apply them.

Labeling (based on overall_score):
- <=14   -> "very_poor"
- 15–34  -> "poor"
- 35–54  -> "fair"
- 55–74  -> "good"
- >=75   -> "excellent"

Overall justification (2–5 sentences):
- Summarize the main strengths and weaknesses.
- Explain why the overall result isn't clearly higher and isn't clearly lower.
- Do NOT mention numeric scores.

==================================================
INPUT FORMAT
==================================================
STORY PROMPT:
{prompt}
STORY:
{story_text}

==================================================
OUTPUT FORMAT (JSON ONLY)
==================================================
Return ONLY a JSON object that matches the provided schema exactly.
No extra keys. No markdown. No commentary outside JSON.
\end{Verbatim}
\end{tcolorbox}